\documentclass[smallextended,twocolumn]{svjour3}
\usepackage{amsmath,amssymb}
\usepackage{float}
\usepackage{caption}
\usepackage{subcaption}
\usepackage{mathtools}
\usepackage{colortbl}
\usepackage{lmodern}
\definecolor{Black}{rgb}{0,0,0}
\usepackage{hyperref}
\usepackage{multicol}
\usepackage{scalerel}
\usepackage{tikz}
\usepackage{wasysym}
\usetikzlibrary{svg.path}
\usepackage{lineno}

\usepackage{pifont}
\newcommand{\cmark}{\ding{51}}

\definecolor{orcidlogocol}{HTML}{A6CE39}
\tikzset{
	orcidlogo/.pic={
		\fill[orcidlogocol] svg{M256,128c0,70.7-57.3,128-128,128C57.3,256,0,198.7,0,128C0,57.3,57.3,0,128,0C198.7,0,256,57.3,256,128z};
		\fill[white] svg{M86.3,186.2H70.9V79.1h15.4v48.4V186.2z}
		svg{M108.9,79.1h41.6c39.6,0,57,28.3,57,53.6c0,27.5-21.5,53.6-56.8,53.6h-41.8V79.1z M124.3,172.4h24.5c34.9,0,42.9-26.5,42.9-39.7c0-21.5-13.7-39.7-43.7-39.7h-23.7V172.4z}
		svg{M88.7,56.8c0,5.5-4.5,10.1-10.1,10.1c-5.6,0-10.1-4.6-10.1-10.1c0-5.6,4.5-10.1,10.1-10.1C84.2,46.7,88.7,51.3,88.7,56.8z};
	}
}
\newcommand\orcidicon[1]{\href{https://orcid.org/#1}{\mbox{\scalerel*{
				\begin{tikzpicture}[yscale=-1,transform shape]
				\pic{orcidlogo};
				\end{tikzpicture}
			}{|}}}}

\begin{document}
	
	\title{A Neurodynamic model of Saliency prediction in V1}
	
	\author{David~Berga*         \and Xavier~Otazu }
	\institute{Computer Vision Center, Universitat Autonoma de Barcelona \\
		Edifici O, Campus UAB, 08193, Bellaterra \\
		Tel.: +34 935 81 18 28 \\
		\email{dberga@cvc.uab.es}\\
		Correspondence: David Berga
	}
	
	

	\date{Received: date / Accepted: date}

	\maketitle
	
	
	\begin{abstract}
		\textit{Objectives:} Lateral connections in the primary visual cortex (V1) have long been hypothesized to be responsible of several visual processing mechanisms such as brightness induction, chromatic induction, visual discomfort and bottom-up visual attention (also named saliency). Many computational models have been developed to independently predict these and other visual processes, but no computational model has been able to reproduce all of them simultaneously. In this work we show that a biologically plausible computational model of lateral interactions of V1 is able to simultaneously predict saliency and all the aforementioned visual processes.\\
		\textit{Methods:} Our model's (NSWAM) architecture is based on Pennachio's neurodynamic model of lateral connections of V1. It is defined as a network of firing rate neurons, sensitive to visual features such as brightness, color, orientation and scale. We tested NSWAM saliency predictions using images from several eye tracking datasets. \\
		\textit{Results:} We show that accuracy of predictions, using shuffled metrics, obtained by our architecture is similar to other state-of-the-art computational methods, particularly with synthetic images (CAT2000-Pattern \& SID4VAM) which mainly contain low level features. Moreover, we outperform other biologically-inspired saliency models that are specifically designed to exclusively reproduce saliency. 
		\textit{Conclusions:} Hence, we show that our biologically plausible model of lateral connections can simultaneously explain different visual proceses present in V1 (without applying any type of training or optimization and keeping the same parametrization for all the visual processes). This can be useful for the definition of a unified architecture of the primary visual cortex.


	\end{abstract}
	
	

	\section{Introduction}
	
	Visual salience can be defined as ``the distinct subjective perceptual quality which makes some items in the world stand out from their neighbors and immediately grab our attention" \cite{Itti2007}. Hence, saliency could be defined as one of the properties of the visual scene that attracts our attention toward a particular set of visual features. Although not being the best option for the study of visual saliency, several studies of eye movements using different approaches have been performed. Eye movements are controlled by many different factors, e.g. low/high-level information, task, endogenous factors, etc. Hence, prediction of eye movement cannot be performed only by one of these factors. Koch and Ullman \cite{Koch1987} propose a computational framework in which visual features are integrated to generate a saliency map. These visual features are projected to V1 and later processed distinctively on the ventral (``what") and dorsal (``where") streams. These connections are projected to the superior colliculus (SC), which would generate either top-down (relevance) or bottom-up (saliency) control of eye movements by combining neuronal activity from distinct brain areas to a unique map (priority map) \cite{Egeth1997}\cite{WhiteMunoz2011}. 
	
	\subsection*{Related Work}
	
	Given these distinct levels of processing from the human visual system (HVS), a set of computational models are proposed in order to reproduce eye movement behavior. Itti et al. introduce a biologically-inspired model \cite{Itti1998} in which low-level features are extracted using linear DoG filters, their conspicuity is calculated using center-surround differences (inspired by V1's simple cell computations) and integrated (pooled to the SC as a master saliency map) using winner-take-all mechanisms. Although computations of existing saliency models seem to mimic HVS mechanisms, complexity of scenes make eye-movement behavior hard to predict because of the aforementioned additional factors. Bruce \& Tsotsos model \cite{Bruce2005} offered a semi-supervised mechanism to account for relevant information of the scenes in combination with the bottom-up computations of V1, predicting eye movement behavior at distinct scene contexts. Given the basis of these models, a myriad of computational models, both with artificial and biological inspiration \cite{Judd2012}\cite{Borji2013c}\cite{Zhang2013}\cite{Riche2016a}, have implemented distinct ways to predict human eye movements obtaining better performance on its predictions \cite{Riche2013a}\cite{Borji2013b}\cite{Borji2013d}\cite{Bylinskii2015}. Thus, although proposed computational eye movement prediction models could precisely resemble eye-tracking data, it is questionable to consider that these predictions accurately and specifically represent saliency \cite{Bruce2015}\cite{Berga2018a}\cite{Berga2019c}. We have added a table describing each model with its inspiration  (Cognitive/Biological, Information-based, Probabilistic and Deep Learning, as described in \cite{Judd2012}\cite{Borji2013c}\cite{Berga2019c}) and type of feature processing (with global or local features).
	Saliency corresponds to bottom-up attention, which derives from generating conspicuity from low-level image features [34]. The relation of eye movements in relation to saliency is mainly based on the fact that eye movements are driven by both bottom-up attention (saliency in this case) and top-down attention (also coined with the name ``relevance" [23]). The problem is that ``saliency models" are told to predict saliency while, in fact, they are predicting fixations (which experimentally is dissimilar, as it includes both bottom-up and top-down effects).
	
	\begin{table}[h!]
		\centering
		\scriptsize
		\setlength{\tabcolsep}{5pt}
		\caption{Description of saliency models}
		\hspace*{-1.25cm}\begin{tabular}{ |c|c|c|cccc|cc| } 
			\hline
			Model & Authors & Year & \multicolumn{4}{ c| }{Inspiration} & \multicolumn{2}{ c| }{Type}\\ 
			& & & C & I & P & D & G & L \\
			\hline
			IKN & Itti et al.\cite{Itti1998} & 1998 & \cmark & & & & \cmark & \cmark \\
			AIM & Bruce \& Tsotsos \cite{Bruce2005} & 2005 & \cmark & \cmark & & & & \cmark \\ 
			GBVS & Harel et al.\cite{harel2006} & 2006 & & & \cmark  & & \cmark & \cmark \\
			SUN & Zhang et al. \cite{Zhang2008} & 2008 & &  & \cmark & &  & \cmark \\
			SDSR & Seo \& Milanfar \cite{Seo2009} & 2009 & \cmark & & \cmark & & \cmark & \cmark \\
			SIM & Murray et al.\cite{Murray2011} & 2011 & \cmark & & & & \cmark & \cmark \\
			AWS & Garcia-Diaz et al.\cite{GarciaDiaz2012} & 2012 & \cmark & & & & \cmark & \cmark \\
			OpenSALICON & Jiang et al.\cite{christopherleethomas2016} & 2015 &  & & & \cmark &  & \cmark \\
			ML-Net & Cornia et al.\cite{mlnet2016} & 2016 &  & & & \cmark &  & \cmark \\
			DeepGazeII & K\"ummerer et al.\cite{Kummerer2016} & 2016 &  & & & \cmark &  & \cmark \\
			SalGAN & Pan et al.\cite{Pan_2017_SalGAN} & 2017 & & & & \cmark &  & \cmark \\
			SAM & Cornia et al.\cite{Cornia2016} & 2018 &  & & & \cmark &  & \cmark \\ 
			\hline
		\end{tabular}
		\\
		Inspiration: \{C: Cognitive/Biological, I: Information-Theoretic, P: Probabilistic, D: Machine/Deep Learning\}
			Type: \{G: Global, L: Local\}
		\label{table:models}
	\end{table}

	\subsection*{Motivation}
	
	Li's work \cite{Li1998}\cite{Li1999}\cite{Li2002}\cite{zhaoping2014understanding} proposes that V1's computations, particularly lateral connections, are the ones responsible of the representation of the aforementioned saliency map. Following her work, the role for the early processing of the visual features relies in V1, mainly driven for this case by uniquely processing low-level visual features. 
	These connections are later projected to the SC in order to generate bottom-up saccadic eye movements  \cite{Schiller1974}\cite[Chapter~9]{SchillerTehovnik2001}\cite{White2017a}. Since Li's architecture only worked on lattices of oriented bars, in order to process greyscale images \cite{Penacchio2013} enhanced Li's architecture by adding receptive fields of different orientations and spatial scales.
	
	These authors show that this architecture reproduces the brightness induction visual process for both still images and dynamic visual stimulus (i.e. videos). By enhancing the architecture adding two channels (opponent red-green and blue-yellow) to the luminance channel, the same authors showed that this architecture also reproduces chromatic induction \cite{Cerda2016}. Evenmore, by studying several statistical properties of the spatial and temporal dynamics of the firing-rate activity of the architecture, they showed that it also predicts the visual phenomena of visual discomfort \cite{Penacchio2016} (which is one of the the main triggers of migraines).
	
	Considering that these works have shown that the neuronal mechanism of lateral connections is partially responsible for these effects, we aim to use the same model in order to address another process present in the primary visual cortex: visual saliency. Using this Penacchio's computational architecture, we aim to compute feature conspicuity (distinctiveness between feature maps), which will alternatively represent the function of the aforementioned saliency map.

	\subsection*{Objectives}
	
	In this study we want to test if the computations of our model are able to reproduce eye movement behavior being consistent with eye-tracking psychophysical experimentation. Current evaluation of saliency predictions is unfair and do not consider many biases, specific to saliency. Using metrics not affected by these biases, we want to show that our architecture can obtain similar results in comparison to other state-of-the-art models (or outpeform them in some specific datasets). Concretely, we want to study whether our architecture can obtain results similar or better than other stateof-the-art models when using datasets with less top-down eye movement biases (that is, datasets which mainly reflects bottom-up saliency related eye movements). Additionally, we want to show that we can obtain these results without applying any type of training or optimization and keeping the same parametrization for all the visual processes (e.g. brightness induction, chromatic induction and visual discomfort). Hence, we want to show that a computational model of lateral connections can offer a unified architecture reproducing distinct V1 functionality, leading to a unification of several visual processes. 
	
	\subsection*{Unifying an architecture of several visual processes} \label{sec:multitask}
	Li's architecture is a model of a neuronal mechanism present in the primary visual cortex (and other areas). All the perceptual processes that rely on this mechanism could be computationally reproduced, at least partially, with the same architecture. As explained in previous section we showed in previous studies \cite{Penacchio2013}\cite{Cerda2016}\cite{Penacchio2016} that the proposed firing-rate neurodynamic model of V1's intra-cortical interactions extended in Penacchio's model \cite{Penacchio2013} is able to simultaneously reproduce several visual processes such as brightness and color induction effects as well as visual discomfort mechanisms.
	
\begin{figure}[h!] 
		\centering
		\begin{subfigure}{0.25\linewidth}
			\includegraphics[width=\linewidth]{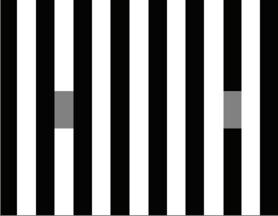}
			\caption*{\centering \textbf{A}}
		\end{subfigure}
		\hspace{0.5mm}
		\begin{subfigure}{0.4\linewidth}
			\includegraphics[width=\linewidth]{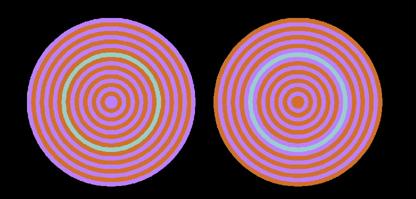} 
			\caption*{\centering \textbf{B}}
		\end{subfigure}
		\begin{subfigure}{0.25\linewidth}
			\includegraphics[width=\linewidth,trim={2cm 0 0 7.80cm},clip]{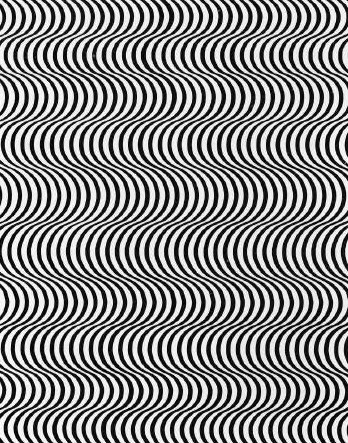} 
			\caption*{\centering \textbf{C}}
		\end{subfigure}
		\caption{\textbf{(A)} Example of Brightness induction present at the White effect \cite{White1979}. The two grey squares are the same luminance, but the the left square is perceived as darker than the right square. \textbf{(B)} Chromatic induction from Monnier \& Shevell's concentric ring stimuli \cite{Monnier2004}. On the left disk we perceive a greenish ring and on the right disk we perceive a bluish, but these two rings are the exactly the same color. \textbf{(C)} Discomfortable image (credit by Nicholas Wade). If we fixate our gaze at this image, after some tens of seconds it could become uncomfortable to look at. \cite{wade1982the}.}
		\label{fig:multitask}
	\end{figure}

	Brightness induction refers to the changes in perceived brightness of a visual target due to the luminance of its surrounding area. From this statement, the HVS can either perceive the visual target and the surrounding area with similar/equal brightness (assimilation) or to perceive brightness differences (contrast). We can observe in \hyperref[fig:multitask]{Fig. \ref*{fig:multitask}\textbf{A}} how two grey patches are perceived distinctively whilst being with same brightness. Similarly, the HVS perceives the chromatic properties of a visual target distinctively depending on the chromaticities of its surrounding area. This phenomena is named chromatic induction. It appears in both ``l" and ``s" opponent channels (``l" for red-green and ``s" for blue-yellow). This effect is observable on \hyperref[fig:multitask]{Fig. \ref*{fig:multitask}\textbf{B}}, where the central ring from the reference stimulus (left) appears to be ``greener" (being perceived with lower ``l" chromatic properties) than the central ring from the test (right), which appears to be ``bluer" instead (being perceived with higher ``s" chromatic properties).
	
	These effects were reproduced previously in a multiresolution wavelet framework with BiWaM \cite{Otazu2008} and CiWaM \cite{Otazu2010} computational models. These models' aim was to mimic V1's simple cell mechanisms by computing center-surround differences at distinct color and luminance opponencies. Being inspired by the aforementioned Li's model, Penacchio et al. \cite{Penacchio2013} modeled an excitatory and inhibitory model of V1 as a more biologically plausible approach to reproduce these visual effects. Considering physiological and neurodynamic properties of V1 cells \cite{Li1998} at different spatial frequencies and orientations, Penacchio et al. \cite{Penacchio2013} show it is possible to simultaneously reproduce psychophysical experiments of brightness \cite{Penacchio2013} and chromatic \cite{Cerda2016} induction effects using a unified computational architecture.
	
	Latest experiments showed that this computational architecture is also able to predict visual discomfort \cite{Penacchio2016}. Specific visual patterns (\hyperref[fig:multitask]{Fig. \ref*{fig:multitask}\textbf{C}}) are shown to cause discomfort, malaise, nausea or even migraine \cite{Penacchio2015}\cite{Le2017}. Taking into account the relative contrast energy from stimulus regions (due to its orientation, luminance, chromatic and spatial frequency distributions), we can predict whether a stimulus can cause hyperexcitability in V1, a possible cause of visual discomfort for certain images.
	
	\subsection*{Hypothesis}
	
	The Hypothesis of the present work is that a computational architecture implementing a biologically plausible model of lateral connections in the primary visual cortex is able to predict low-level saliency while simultaneously reproducing all the previously commented visual processes (e.g. brightness and chromatic induction, and visual discomfort).
	
	\section{Model Description}
	
	The model is extended from previous implementation by Pennacchio et al. \cite{Penacchio2013} in Matlab and C++ \footnote{Code can be downloaded from \url{https://github.com/dberga/NSWAM}}. Here we describe the main steps in relation to the computations done to the images: \hyperref[sec:feat1]{\ref*{sec:feat1}. Feature Extraction}, \hyperref[sec:feat2]{\ref*{sec:feat2}. Feature Conspicuity} and \hyperref[sec:feat3]{\ref*{sec:feat3}. Feature Integration}. In this section, computations in the early visual pathways will be represented in line with a stimulus example. Overall model architecture was inspired by previous work from Murray et al.'s Saliency Induction Model (SIM) \cite{Murray2011}, defining a biologically-inspired and unsupervised low-level model for saliency prediction. Although it provided a promising approach for predicting saliency maps, we want to stress the novelty of  computations of firing rate dynamics proposed in our architecture are in accordance with physiological properties of V1 cells.

	\subsection{From images to Sensory Signals: Feature Extraction} \label{sec:feat1}
	
	\subsubsection{Color representation}
	
	Human retinal cone photoreceptors are sensitive to distinct wavelengths of the visual spectrum, corresponding to long, medium and short wavelengths. Similarly, traditional digital cameras capture light as values in the RGB color space (corresponding to Red, Green and Blue components).
	Retinal ganglion cells (RGC) encode luminance and chromatic signals as an opponent representation. This opponent representation separates channels of ``Red vs Green" and ``Blue vs Yellow" from cone cell responses, and luminance (``Bright vs Dark") from both cones and rod responses. Activity from these channels (R-G, B-Y and L) is then projected respectively to the lateral geniculate nucleus (LGN) and through parvo-cellular (P-), konio-cellular (K-) and magno-cellular (M-) pathways towards V1.
	
	In order to represent this opponent colour information, we use the widely used opponent colour representation:
	
	\begin{align} 
	\begin{split} \label{eq:lab1}
	L=R+G+B,
	\end{split} \\
	\begin{split} \label{eq:lab2}
	rg=\dfrac{R-G}{L},
	\end{split} \\
	\begin{split} \label{eq:lab3}
	by=\dfrac{R+G-2B}{L},
	\end{split}
	\end{align}
	
\begin{figure}[h!]
		\centering
		\begin{subfigure}{0.1\textwidth}
			\includegraphics[width=\textwidth]{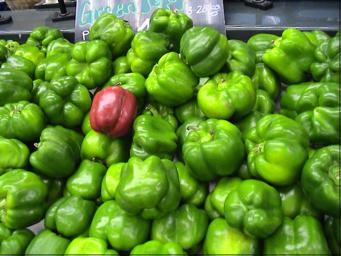} 
			\caption*{}
		\end{subfigure}
		$\Rightarrow$
		\begin{subfigure}{0.1\textwidth}
			\includegraphics[width=\textwidth]{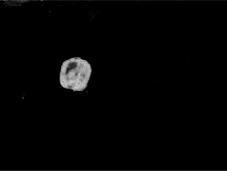} 
			\caption*{\centering \textbf{A}}
		\end{subfigure}
		\begin{subfigure}{0.1\textwidth}
			\includegraphics[width=\textwidth]{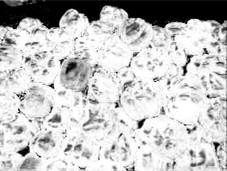} 
			\caption*{\centering \textbf{B}}
		\end{subfigure}
		\begin{subfigure}{0.1\textwidth}
			\includegraphics[width=\textwidth]{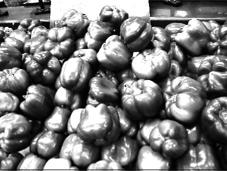} 
			\caption*{\centering \textbf{C}}
		\end{subfigure}
		\caption{Example of RGB image (left image) and its corresponding opponent color representation: \textbf{(A)} ``red vs green" ($rg$) , \textbf{(B)} ``blue vs yellow" ($by$) and \textbf{(C)} ``luminance" ($L$) channels.}
		\label{fig:opp}
	\end{figure}

	We can interpret $L$, $rg$ and $by$ components defined in \hyperref[eq:lab1]{Eqs. \ref*{eq:lab1},\ref*{eq:lab2},\ref*{eq:lab3}} as means of luminance opponency and chrominance oponencies R-G and B-Y, respectively.
	In \hyperref[fig:opp]{Fig.  \ref*{fig:opp}} we illustrate an example of an image and its conversion to this representation, with higher activation on the ``Red vs Green" opponent cells than the case of ``Blue vs Yellow" and ``Bright vs Dark" opponencies. It has been shown that this representation is related to some perceptual properties of colour perception \cite{Parraga98}.
	
	All RGB pixel values of processed images are previously corrected with $\gamma=1/2.2$.
	
	\subsubsection{Multiscale and orientation representation}
	
\begin{figure}[h!] 
		\raggedright \hspace{0.03\linewidth} \hspace{0.025\linewidth} \tiny{$\theta=h$} \hspace{0.025\linewidth} \tiny{$\theta=v$} \hspace{0.025\linewidth} \tiny{$\theta=d$}
		\\ \vspace{2mm}
		\begin{subfigure}{0.025\linewidth}
			\raggedright
			$s$=$1$ \vspace{1.5\linewidth} \\
			$s$=$2$ \vspace{1.5\linewidth} \\
			$s$=$3$ \vspace{1.5\linewidth} \\
			$s$=$4$ \vspace{1.5\linewidth} \\
			$s$=$5$ \vspace{1.5\linewidth} 
		\end{subfigure}
		\hspace{3mm}
		\centering
		\begin{subfigure}{0.3\linewidth}
			\includegraphics[width=\textwidth,height=4cm,clip, trim=50px 27px 50px 27px]{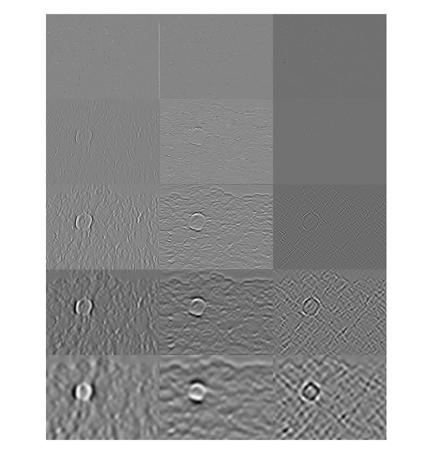}
			\caption*{\centering \textbf{A}}
		\end{subfigure}
		\begin{subfigure}{0.3\linewidth}
			\includegraphics[width=\textwidth,height=4cm,clip, trim=50px 27px 50px 27px]{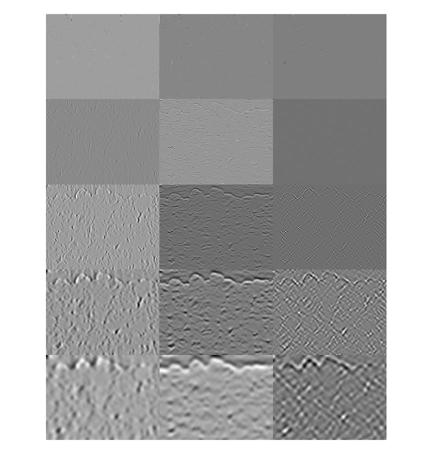}
			\caption*{\centering \textbf{B}}
		\end{subfigure}
		\begin{subfigure}{0.3\linewidth}
			\includegraphics[width=\textwidth,height=4cm,clip, trim=50px 27px 50px 27px]{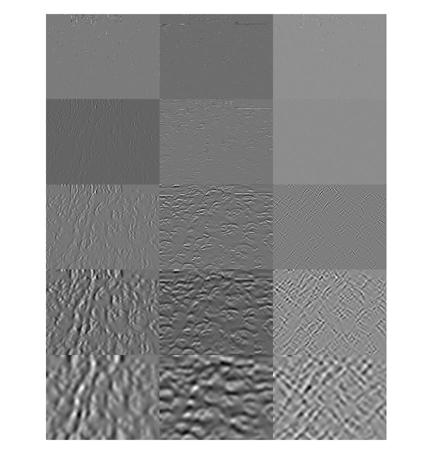}
			\caption*{\centering \textbf{C}}
		\end{subfigure}
		\caption{Output from \textit{\`a-trous DWT} of the signals shown on \hyperref[fig:opp]{Fig. \ref*{fig:opp}}. We show values for rescaled wavelet filters, with scales $s=1..5$ and orientations $\theta=h,v,d$ corresponding to distinct channel opponencies \textbf{(A)} $\omega_{o=rg}$, \textbf{(B)} $\omega_{o=by}$ and \textbf{(C)} $\omega_{o=L}$.}
		\label{fig:atrous}
	\end{figure}

	V1 cell sensitivities to distinct orientations \cite{Hubel1968} and spatial frequencies \cite{Maffei1973} are usually modeled as Gabor filters. Since Gabor transforms cannot be inverted to obtain the original image, we used the \textit{\`a trous} algorithm, which is an undecimated discrete wavelet transform (DWT) \cite{GonzlezAudcana2005}\cite[Chapter~6]{stathaki2008image}. This decomposition allows to perform an inverse, where the basis functions remain similar to Gabor filters. We propose biologically plausible computations for extracting multiple orientations and multiscale feature representations of from V1's receptive field (RF) hypercolumnar organization (\hyperref[fig:atrous]{Fig. \ref*{fig:atrous}}). The wavelet approximation planes $c_{s,\theta}$ ($s$ for scale and $\theta$ for orientation) are computed by convolving the image with the filter $h_s$.
	
	\begin{align}
	\begin{split} 
	c_{s,h}= c_{s-1} \otimes h_s, \label{eq:wavelets1}\\
	c_{s,v}= c_{s-1} \otimes h_s'.
	\end{split}
	\end{align}
	
	\noindent The filter $h_s$ is obtained from $h_{s-1}$ by doubling its size, i.e. $h_s$ = $\uparrow$ $h_{s-1}$, where $\uparrow$ means upsampling by introducing zeros between the coefficients. The filter ($h_s$) for the first scale is
	
	\[h_1=\frac{1}{16}\begin{bmatrix}
	1&4&6&4&1
	\end{bmatrix}\]
	
	\noindent This filter can be also transposed ($h_s'$) to obtain distinct approximation orientation planes $c_{s,h}$ and $c_{s,v}$. From these approximation planes, we can obtain the wavelet coefficients $\omega_{s,\theta}$ at distinct scales and orientations:
	
	\begin{align}
	\begin{split}
	\omega_{s,h}=c_{s-1}-c_{s,h}, \label{eq:wavelets2} \\
	\omega_{s,v}=c_{s-1}-c_{s,v}, \\
	\omega_{s,d}=c_{s-1}-(c_{s,h}\otimes h_s' + \omega_{s,h}+\omega_{s,v}), \\
	c_s = c_{s-1} - (\omega_{s,h}+\omega_{s,v}+\omega_{s,d}). 
	\end{split}
	\end{align}
	
	\noindent Here, $\omega_h$, $\omega_v$ and $\omega_d$ correspond to the coefficients with ``horizontal", ``vertical" and ``diagonal" orientations. Initial $c_0=I_o$ (e.g. $s=0$) is obtained from the opponent components ($o=L,rg,by$) and $c_n$ corresponds to the residual plane of the last wavelet component (e.g. $s=n$). The inverse transform is obtained by integrating wavelet coefficients and residual planes:
	
	\begin{align}
	\begin{split}
	I_o'=\sum_{s=1, \theta=h,v,d}^{n} \omega_{s,\theta} + c_n. \label{eq:wavelets3}
	\end{split}
	\end{align}
	
	Considering that for every image, $M \times N$ is the size of the feature map (resized to $N \leq 128$), the set of spatial scales is ($s={1..S}$), where $S=\lfloor log_2(N/8) \rfloor+2$.

	\subsection{Computing V1 Dynamics: Feature Conspicuity} \label{sec:feat2}
	
\begin{figure}[h!]
		\centering
		\begin{subfigure}{.45\linewidth}
			\includegraphics[width=\textwidth]{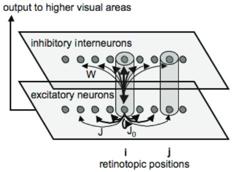}
			\caption*{\centering \textbf{A}}
		\end{subfigure}
		\begin{subfigure}{.45\linewidth}
			\includegraphics[width=\textwidth]{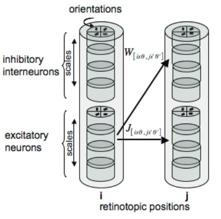}
			\caption*{\centering \textbf{B}}
		\end{subfigure}
		\begin{subfigure}{.9\linewidth}
			\includegraphics[width=\textwidth]{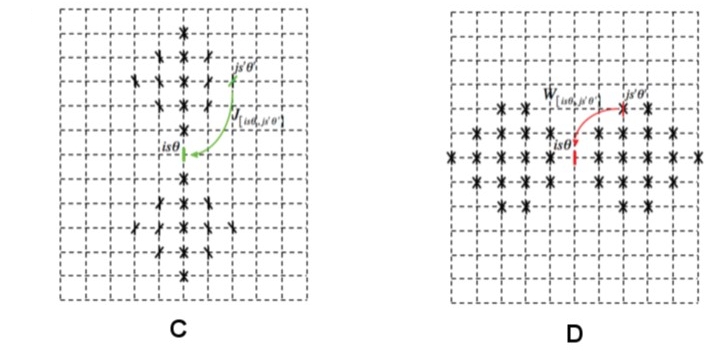}
		\end{subfigure}
		\caption{Illustration of the different elements and their connections that define our computational network. \textbf{(A)} Two populations of excitatory and inhibitory nodes are defined in a 2-dimensional regular discrete lattice (here reduced to a single dimension for the sake of clarity). Nodes of these two populations are connected between them. The output of te layer of excitatory  nodes is considered the output of the network. \textbf{(B)} At each retinotopic position we consider we have an hypercolumn composed by a set excitatory and inhibitory nodes tuned to different spatial orientations and scales. \textbf{(C,D)} Sketches of the weights of both the excitatory $J$ and inhibitory $W$ between retinotopic positions $i$ and $j$. These connections are traslation invariant.}. \scriptsize{Reprinted with permission from ``A Neurodynamical Model of Brightness Induction in V1", 2013, by O. Penacchio, \textit{PLoS ONE, 8(5):e64086}, p.5. Copyright 2013 by the Public Library of Science \cite{Penacchio2013}.}
		\label{fig:model}
	\end{figure}

	Feature conspicuity from previous Murray's SIM model is computed using center-surround feature computations (CS) while applying a contrast sensitivity function (eCSF). Similarly, we extract low-level feature-dependent computations corresponding to the orientation sensitivities ($\theta=0,90,45/135^{\circ}$) of the retinotopic positions ($i$) at distinct spatial frequencies ($s$) for ON and OFF-center cells. These ON and OFF cells activities (before the computation of lateral connections) responses are computed by taking the positive and negative values of the wavelet planes, respectively. Feature distinctiveness is computed with the Penacchio et al. network of excitatory-inhibitory firing rate neurons, simulating V1's lateral interactions (\hyperref[fig:model]{Fig. \ref*{fig:model}}). Contrast enhancement or suppression emerges from lateral connections as an induction mechanism. Lateral interactions are implemented to have self-directed ($J_0$) and monosynaptic connections ($J$) between excitatory neurons. Inhibitory interactions have disynaptic connections ($W$) through all inhibitory interneurons, defined by:
	\begin{align} 
	\begin{split}\label{eq:model4}
	J_{[is\theta,js'\theta']}= 
	\lambda(\Delta_s)  0.126  e^{(-\beta / d_s )^2 - 2(\beta / d_s)^7 - d_s^2/90},
	\end{split}
	\\
	\begin{split}\label{eq:model5}
	W_{[is\theta,js'\theta']}= 
	\lambda(\Delta_s)  0.14  (1-e^{-0.4(\beta / d_s)^{1.5}})e^{-(\Delta_\theta/(\pi/4))^{1.5}},
	\end{split}
	\end{align}
	
	Equation \ref*{eq:model4} is applied if $(0<d\leq 10$ and $\beta<\pi/2.69)$ or $[(0<d\leq 10$ and $\beta<\pi/2.69$) and $|\theta_1|<\pi/5.9$ and $|\theta_2|<\pi/5.9]$, otherwise $J_{[is\theta,js'\theta']}=0$. We take $W_{[is\theta,js'\theta']}=0$ if $d=0$ or $d\geq 10$ or $\beta<\pi/1.1$ or $|\Delta\theta|\geq\pi/3$ or $|\theta_1|<\pi/1.99$, otherwise we use the expression in Equation \ref*{eq:model4}. In these equations, $d=d(i,j)$ is the distance between the nodes at position $i$ and $j$, and $\theta_1, \theta_2$ are the angles between the nodes and the line defined by $i-j$, with $|\theta_1|\leq|\theta_2|\leq\pi/2$. The sign of the angles is determined by the condition $|\theta_i|\leq\pi/2$. Parameter $\beta=2\theta_1+2\sin{|\theta_1+\theta_2|}$ and $\Delta\theta=\theta-\theta'$ (with $|\theta-\theta'\leq\pi/2$). Term $\lambda(\Delta_s)$ is related to the difference between the spatial scales $(\Delta_s=|s-s'|)$ of the two connected nodes. Details of this term can be found on \cite[\href{https://s3-eu-west-1.amazonaws.com/pstorage-plos-3567654/1066250/Text_S1.pdf}{Supporting Information S1}]{Penacchio2013}.
	
	In Fig.\ref{fig:model} C and D we have shown a graphical representation of these connections. Considering these in a simulated retinotopic space (corresponding to a the visual space but at distinct RF sizes) with a radius $\Delta_s=15\times 2^{s-1}$ and radial distance $\Delta_\theta$ (respectively accounting for the distance between RF neurons from different spatial frequencies as $d_s$ and radial distance as $\beta$). We can see that excitatory connections $J$ are defined between nodes with similar orientation that are relatively aligned. In contrast, inhibitory connections $W$ are defined between nodes with similar orientation but non-aligned.

	
	
	Excitatory and inhibitory membrane potentials (their derivatives) are described by 
	
	\begin{align}
	\begin{split}  \label{eq:model2}
	\dot{x}_{is\theta} & = -\alpha_x x_{is\theta}-g_y(y_{is\theta})\\
	&-\sum_{\Delta_s,\Delta_\theta \neq 0}\Psi(\Delta_s,\Delta_\theta)g_y(y_{is}+\Delta_{s\theta}+\Delta_\theta)+J_0 g(x_{is\theta}) \\ 
	&+ \sum_{j\neq i,s',\theta'} J_{[is\theta,js'\theta']}g_x(x_{js'\theta'})+I_{is\theta}+I_0,
	\end{split} 
	\\
	\begin{split} \label{eq:model3}
	\dot{y}_{is\theta} & = -\alpha_y y_{is\theta}-g_x(x_{is\theta}) \\
	&+\sum_{j\neq i,s',\theta'} W_{[is\theta,js'\theta']}g_x(x_{js'\theta'}) + I_c \quad.
	\end{split} 
	\end{align}
	
	Functions $g_x$ and $g_y$ correspond to the activation function (implemented as piece-wise linear functions) for transforming the membrane potentials to firing rate values. The spread of the inhibitory activity within a hypercolumn is represented as $\Psi$. Terms $\alpha_x=1/\tau_x$, $\alpha_y=1/\tau_y$ are the decay constants that define the decay of excitatory and inhibitory potentials to their resting potential values, respectively. Terms $\tau_x$ and $\tau_y$ are the mean time that excitatory and inhibitory membrane potentials, respectively, take to decay to its mean value. We have used $\alpha_x=\alpha_y=1$ values. The variable $I_{is\theta}$ corresponds to the external input values of the image, which in our case are the wavelet coefficients that simulate the response of the classical receptive field of every node ($I_{is\theta}\equiv\omega_{is\theta}$). Inhibitory top-down activity can be introduced to the model through $I_{c}$, including a noise signal to stabilize the nonlinear equilibrium. We suggest to read further details of the model and its parameters are specified in \cite[\href{https://s3-eu-west-1.amazonaws.com/pstorage-plos-3567654/1066250/Text_S1.pdf}{Supporting Information S1}]{Penacchio2013}. We compute the temporal average of ON and OFF-center cells $M(\omega^{t+}_{is})$ and $M(\omega^{t-}_{is})$ as the model output over several oscillation cycles (being the mean of $g_x$ for a specific range of $t$, where $t$ is the membrane time, which corresponds to $10$ ms) from distinct color opponencies ($o=L,rg,by$). Distinctively from the induction cases described in \nameref{sec:multitask}, we do not combine the model output $M(\omega^{t}_{iso})$ to the coefficients $\omega^{t}_{iso}$, instead, we consider the firing rate from the model output as our predictor of feature distinctiveness, which will define our main function for our saliency map (\hyperref[eq:model6]{Eq. \ref*{eq:model6}}). The model output can provide detail of single neuron dynamics of firing rate, which its dynamical properties may vary across stimulus properties such as color opponency, scale and orientation.

		\begin{equation} \label{eq:model6}
		\hat{S}^{t}_{iso\theta}=M(\omega^{t+}_{iso\theta})+M(\omega^{t-}_{iso\theta})+c_i,
		\end{equation}
	
	\subsection{Generating the saliency map: Feature Integration} \label{sec:feat3}
	
	After computing feature distinctiveness for the low-level feature maps, we need to integrate these conspicuity or distinctiveness maps in order to pool the neuronal activity to the projections of the SC as means of acquiring a unique map, which will represent our saliency map. First, we have computed the inverse transform from the DWT (IDWT) \hyperref[eq:wavelets3]{Eq. \ref*{eq:wavelets3}} for integrating the sensitivities for orientation ($\theta$) and spatial frequencies ($s$). In this case, instead of the $\omega_{s,\theta}$, we use $\hat{S}$ as the sum of ON and OFF cells after processing the dynamical model (\hyperref[eq:model6]{Eq. \ref*{eq:model6}}) summated for each channel:

		\begin{align}
		\begin{split}
		\hat{S}_{io}(inverse/sum)=\sum_{s=1..S, \theta=h,v,d}^{n} \hat{S}_{iso\theta} +c_n. \label{eq:sc1} 
		\end{split}
		\end{align}
	
	\noindent 
	Second, we have computed the euclidean norm  ($\hat{S}$) for integrating the firing rate of the distinct color opponencies (\hyperref[eq:saliency2]{Eq. \ref*{eq:saliency2}}). 
	
	\begin{equation}
	\hat{S}_i=\sqrt[]{\hat{S}_{i;rg} + \hat{S}_{i;by} + \hat{S}_{i;L}},
	\label{eq:saliency2}
	\end{equation}
	
	Third, we have normalized the resulting map ($z(\hat{S})$) by the variance of the firing rate (\hyperref[eq:saliency3]{Eq. \ref*{eq:saliency3}}), as stated by Li \cite[Chapter~5]{zhaoping2014understanding}. Finally, we convolved the saliency map with a Gaussian filter in order to simulate a smoothing caused by the deviations of $\sigma=1$ deg given from eye tracking experimentation, recommended by LeMeur \& Baccino \cite{LeMeur2012}.
	
	\begin{equation} 
	z_{i}(\hat{S})=\frac{\hat{S}_i-\mu_{\hat{S}}}{\sigma_{\hat{S}}},
	\label{eq:saliency3}
	\end{equation}
	\hfill\newline
	
	\noindent
	where $\mu_{\hat{S}}$ and $\sigma_{\hat{S}}$ are the mean value and the standard deviation of $\hat{S}_i$ over all $i$ pixels, respectively.

	\section{Experiments}
	
\begin{figure}[h!] 
		\begin{subfigure}{\linewidth}
			\centering
			A\includegraphics[width=.2\linewidth]{examples_72.jpg}
			B \includegraphics[width=.2\linewidth]{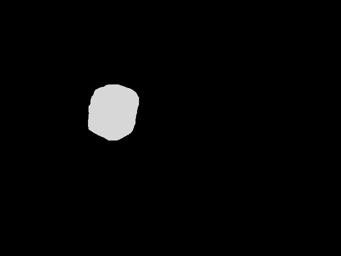}
			C\includegraphics[width=.2\linewidth]{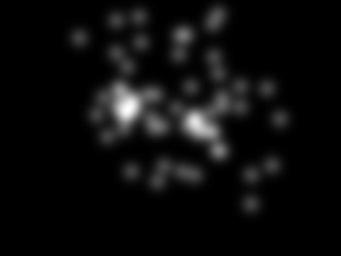}
			\\ \vspace{2mm}
			D\includegraphics[width=.15\linewidth]{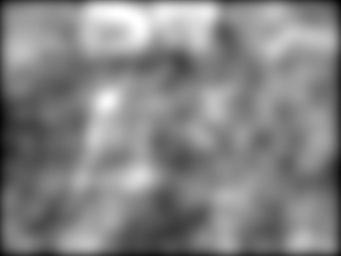}
			E\includegraphics[width=.15\linewidth]{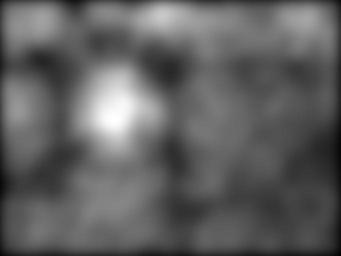}
			F\includegraphics[width=.15\linewidth]{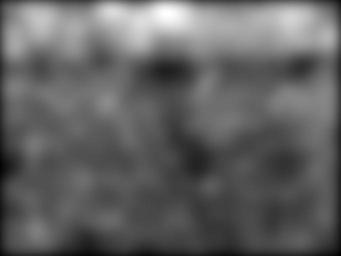}
			G\includegraphics[width=.15\linewidth]{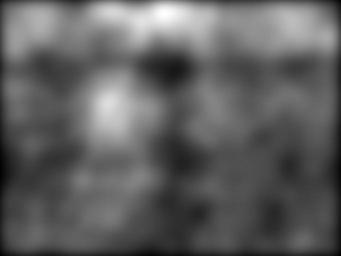}
		\end{subfigure}
		\begin{subfigure}{\linewidth}
			\centering
			\vspace{2mm}
			\includegraphics[width=1\linewidth]{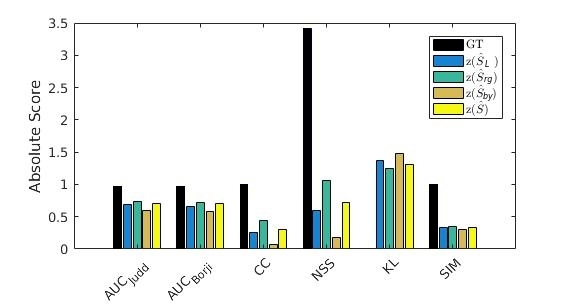}
		\end{subfigure}
		\caption{\textbf{(A)} Example Image. \textbf{(B)} Mask of the salient region (manually defined). \textbf{(C)} Fixation density map (GT, i.e. ground truth) obtained by psychophysical experimentation with observational subjects. \textbf{(D,E,F,G)} Predicted saliency maps of the different color opponent channels $z(\hat{S}_{L})$, $z(\hat{S}_{rg})$, $z(\hat{S}_{by})$ and the final saliency map $z(\hat{S})$ respectively. \textbf{(E)} Comparing the mask and the fixations (both by GT and the computationally predicted) we calculate different metrics from these saliency maps. Results for $z(\hat{S})$ corresponds to our model's saliency prediction (NSWAM). We can see that NSWAM obtain results very similar to other methods.}
		\label{fig:example}
	\end{figure}

	In order to test the validity of our hypothesis, we tested the accuracy of NSWAM for prediction of visual saliency using fixations from eye-tracking experiments. Eye movement data (i.e. ground truth or GT) is combined across all fixations from participants' data, being represented as binary maps (called fixation maps), according to the fixation localizations in the visual space for each corresponding image, or as density distributions (alternatively named density maps) from these fixations considering eye-movement localization probabilities (\hyperref[fig:qualitative]{Fig. \ref*{fig:example}}). Fixation density maps are computed accordingly from fixation maps with a Gaussian filter \cite{LeMeur2012}.
	
	\subsection*{Saliency Metrics}
	
	Prediction scores are calculated using spatially dependent metrics \cite{mit-saliency-benchmark}\cite{Bylinskii2016} which compare either fixation maps or fixation density maps to saliency map predictions from the models. For the case of AUC, it computes the Area Under ROC considering true positive (TP) values for the saliency predictions inside the locations from the fixation maps and false positive (FP) values for saliency outside the maps. The Normalized Scanpath Saliency (NSS) is calculated by standarizing the saliency map of the TP. Other metrics such as Correlation Coefficient (CC) or Similarity (SIM), compare correlations of pixels between fixation density maps and predicted saliency maps. Also using the fixation density maps as GT, the Kullback-Leibler divergence (KL) measures the statistical difference between the two maps (the density map of GT and the saliency map), therefore the lower score is the better.
	
	Other metrics compare saliency maps with a baseline set of other image fixation maps in order to prevent behavioral tendencies such as center biases (see \cite{Berga2018a}\cite{Hayes2019}), which are not representative data for saliency prediction. For instance, the shuffled AUC (sAUC) is calculated as the proportion between TP of the current GT and penalizes for TP of GT from other images. For the case of Information Gain (InfoGain) a Gaussian baseline of all GT (adding up fixations for all dataset to one unique map) is substracted from the prediction for penalizing for center biases.

	\subsection{Predicting human eye movements in natural images}
	
	We have computed the saliency maps\footnote{Code for model evaluations can be downloaded in \url{https://github.com/dberga/saliency}} for images from distinct eye-tracking datasets, corresponding to 120 real scenes (Toronto) \cite{Bruce2005}, 40 nature scenes (KTH) \cite{Kootstra2011}, 100 synthetic patterns (CAT2000$_{Pattern}$)\cite{CAT2000} and 230 synthetic images with specific feature contrast (SID4VAM) \cite{Berga2018a}\cite{Berga2019c}. We have computed these image datasets with deep supervised artificial saliency models that specifically compute high-level features (OpenSalicon \cite{Huang2015}\cite{christopherleethomas2016}, DeepGazeII \cite{Kummerer2016}, SAM \cite{Cornia2016}, SalGan \cite{Pan_2017_SalGAN}), and models that extract low-level features, corresponding to the cases with artificial (SUN \cite{Zhang2008}, GBVS \cite{Harel2006}) and biological inspiration (IKN \cite{Itti1998}, AIM \cite{Bruce2009}, SSR \cite{Seo2009}, AWS \cite{GarciaDiaz2012} and SIM \cite{Murray2011}). The Saliency WAvelet Model (SWAM) and Neurodynamic SWAM (NSWAM) corresponds to our model excluding or including lateral interactions explained in \hyperref[sec:feat2]{Section \ref{sec:feat2}}.

\begin{figure*}
		\centering
		\makebox[\textwidth][c]{
			\begin{tabular}{@{}ccccccc@{}}
				Image & GT & IKN & AIM & SWAM & SIM  & NSWAM \\
				& (Human Fix.)  & & & \footnotesize{(Ours)}  & \footnotesize{(SWAM+CS\&eCSF)} & \footnotesize{(Ours)} \\
				\includegraphics[width=.12\textwidth]{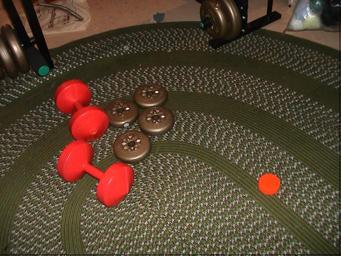} &
				\includegraphics[width=.12\textwidth]{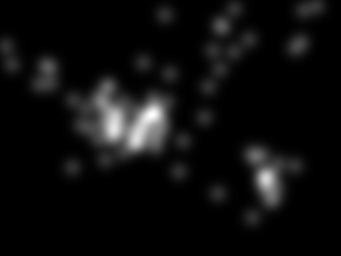} &
				\includegraphics[width=.12\textwidth]{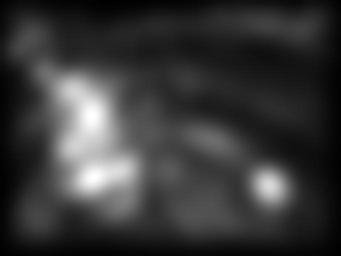} &
				\includegraphics[width=.12\textwidth]{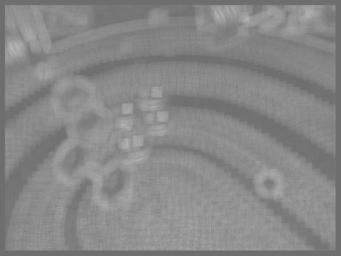}   &
				\includegraphics[width=.12\textwidth]{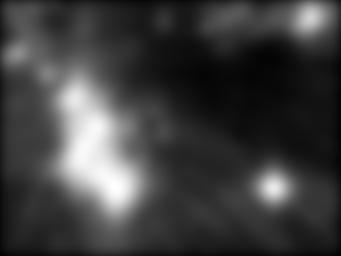}  &
				\includegraphics[width=.12\textwidth]{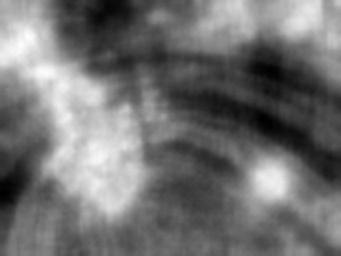}  &
				\includegraphics[width=.12\textwidth]{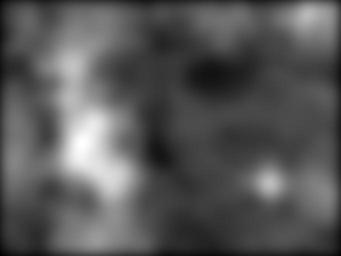}  \\
				\includegraphics[width=.12\textwidth]{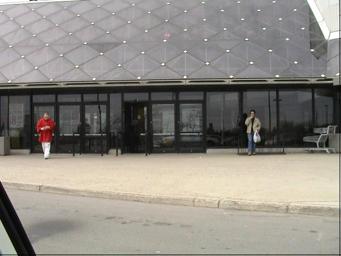} &
				\includegraphics[width=.12\textwidth]{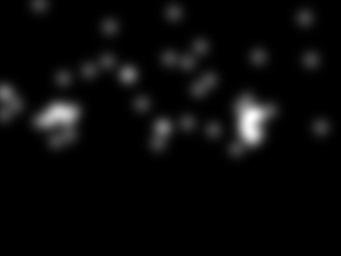} &
				\includegraphics[width=.12\textwidth]{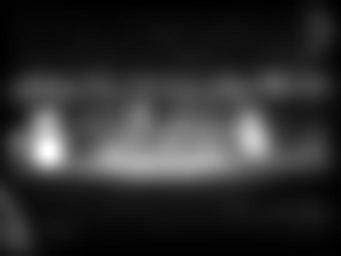} &
				\includegraphics[width=.12\textwidth]{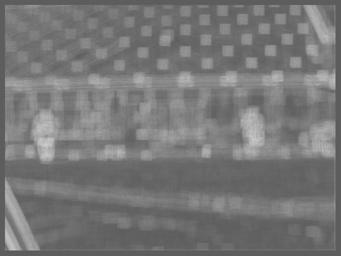}   &
				\includegraphics[width=.12\textwidth]{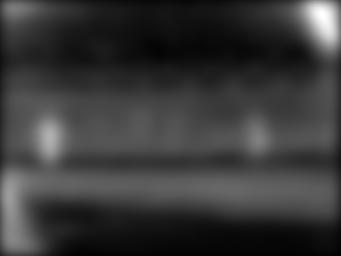}  &
				\includegraphics[width=.12\textwidth]{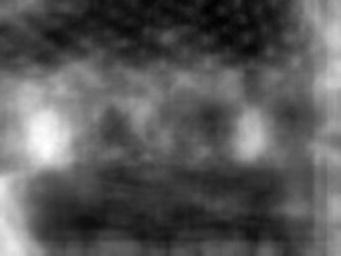}  &
				\includegraphics[width=.12\textwidth]{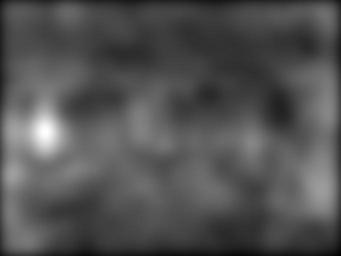}  \\
				\includegraphics[width=.12\textwidth]{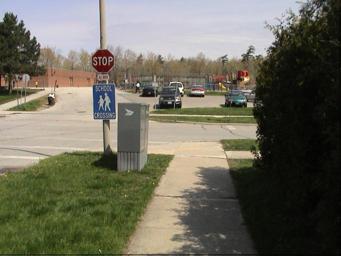} &
				\includegraphics[width=.12\textwidth]{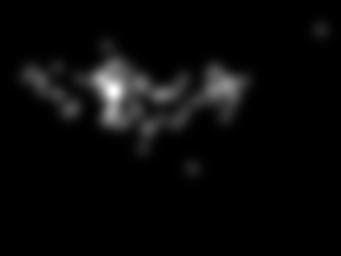} &
				\includegraphics[width=.12\textwidth]{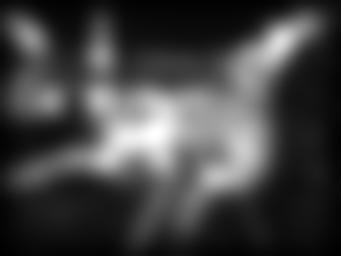} &
				\includegraphics[width=.12\textwidth]{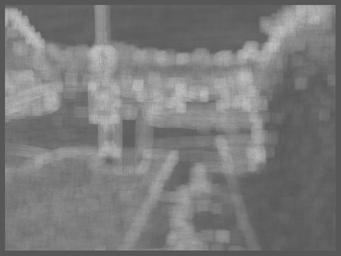}   &
				\includegraphics[width=.12\textwidth]{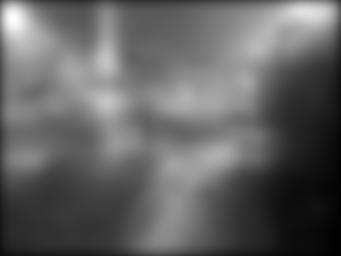}  &
				\includegraphics[width=.12\textwidth]{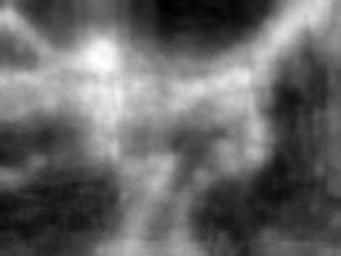}  &
				\includegraphics[width=.12\textwidth]{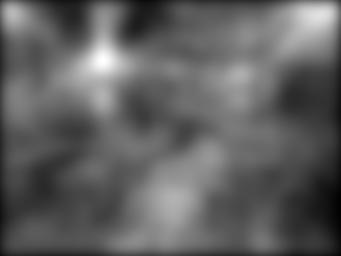}  \\
				\includegraphics[width=.12\textwidth]{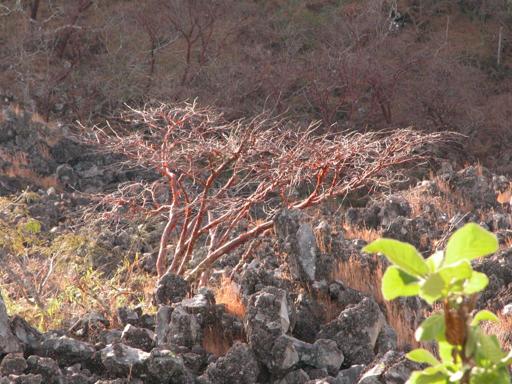} &
				\includegraphics[width=.12\textwidth]{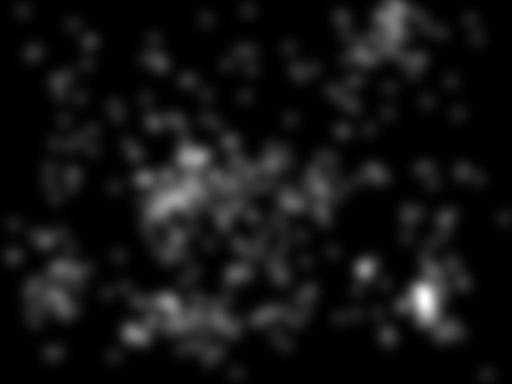} &
				\includegraphics[width=.12\textwidth]{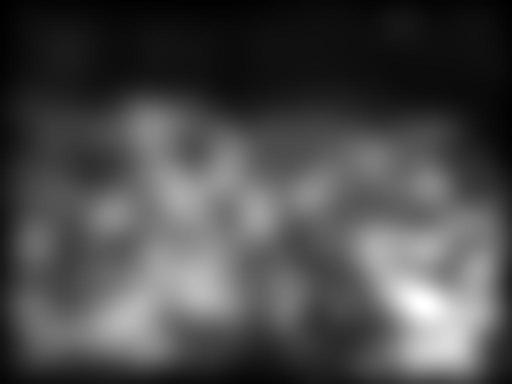} &
				\includegraphics[width=.12\textwidth]{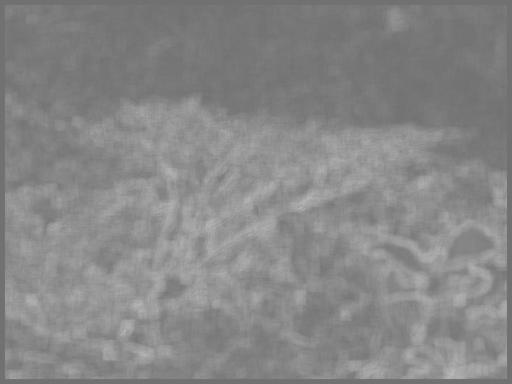} &
				\includegraphics[width=.12\textwidth]{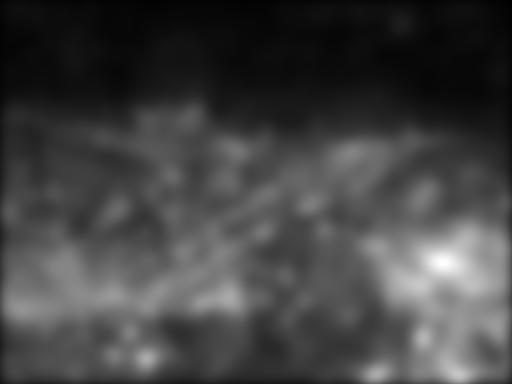}  &
				\includegraphics[width=.12\textwidth]{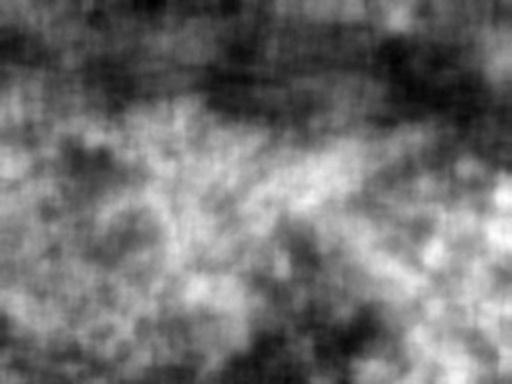}  &
				\includegraphics[width=.12\textwidth]{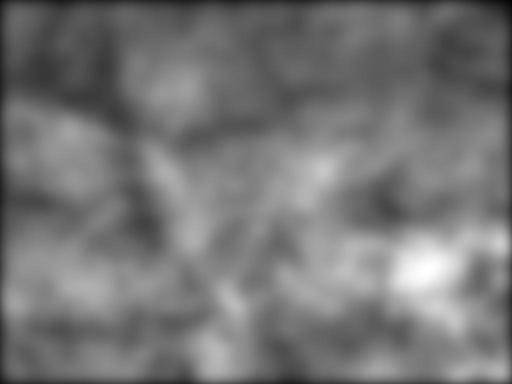}  \\
				\includegraphics[width=.12\textwidth]{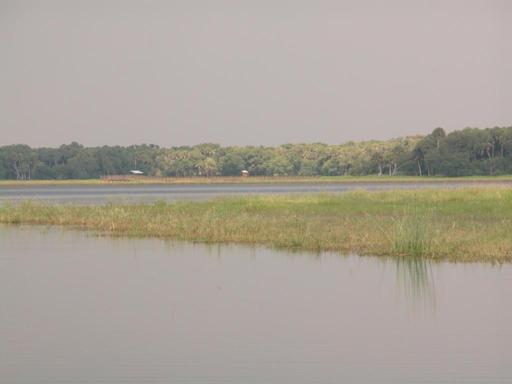} &
				\includegraphics[width=.12\textwidth]{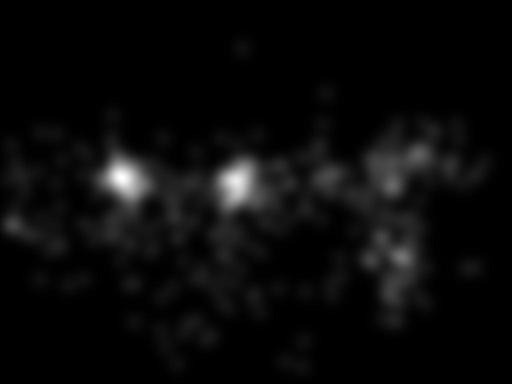} &
				\includegraphics[width=.12\textwidth]{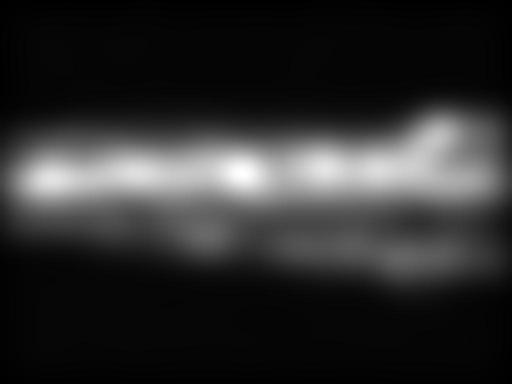} &
				\includegraphics[width=.12\textwidth]{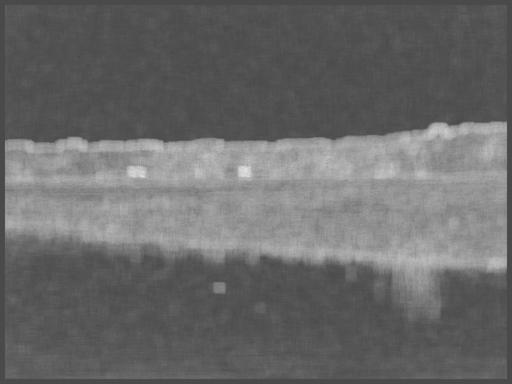} &
				\includegraphics[width=.12\textwidth]{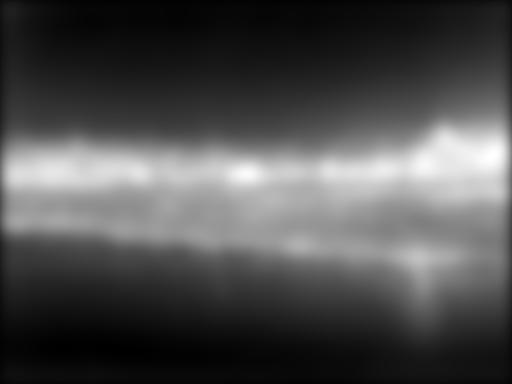}  &
				\includegraphics[width=.12\textwidth]{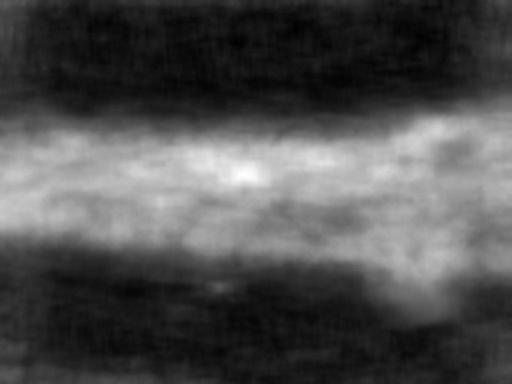}  &
				\includegraphics[width=.12\textwidth]{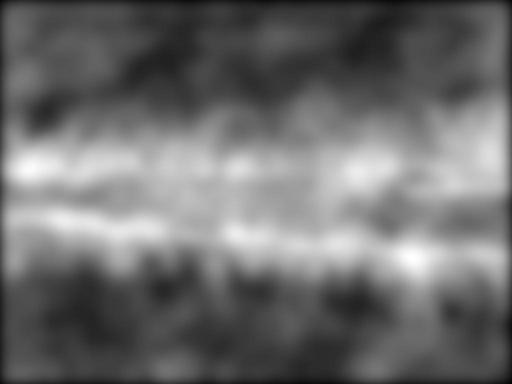}  \\
				\includegraphics[width=.12\textwidth]{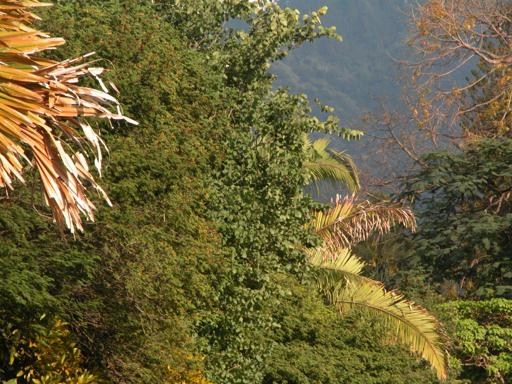} &
				\includegraphics[width=.12\textwidth]{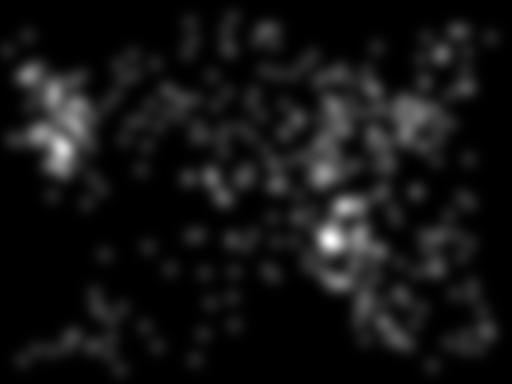} &
				\includegraphics[width=.12\textwidth]{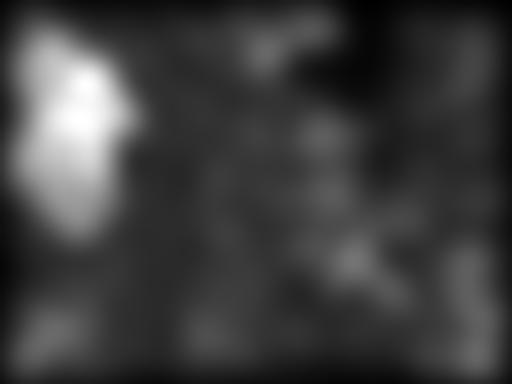} &
				\includegraphics[width=.12\textwidth]{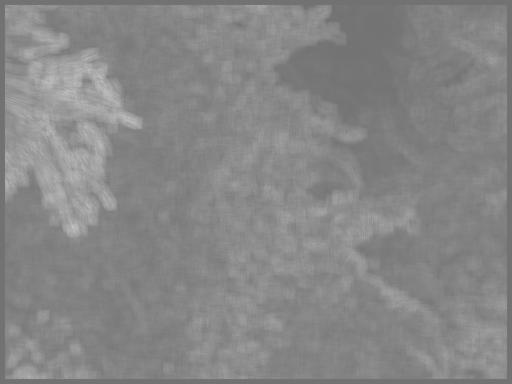} &
				\includegraphics[width=.12\textwidth]{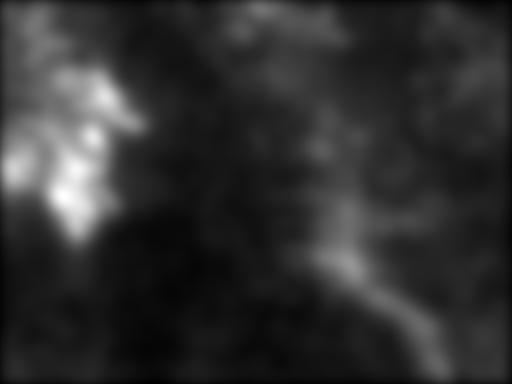}  &
				\includegraphics[width=.12\textwidth]{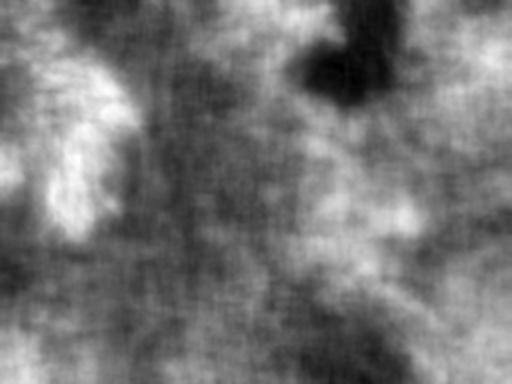}  &
				\includegraphics[width=.12\textwidth]{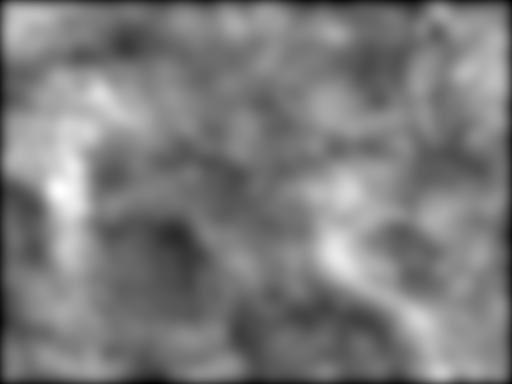}  \\
				\includegraphics[width=.12\textwidth]{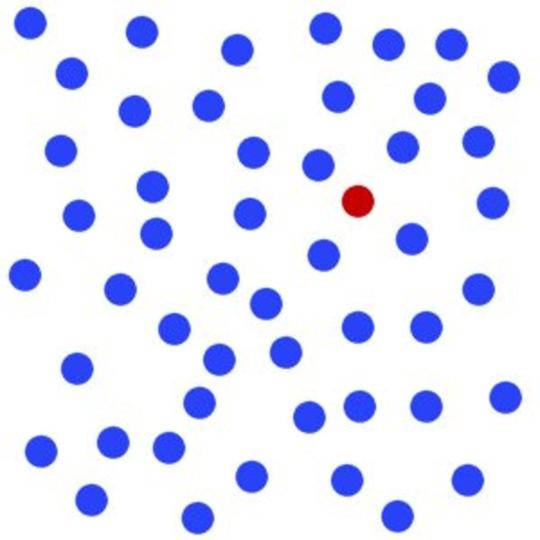} &
				\includegraphics[width=.12\textwidth]{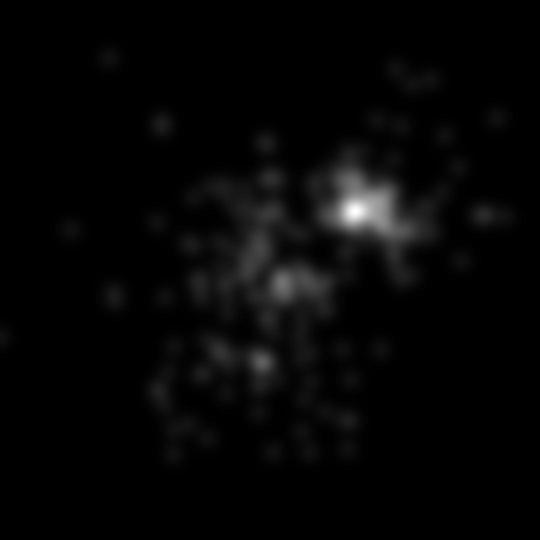} &
				\includegraphics[width=.12\textwidth]{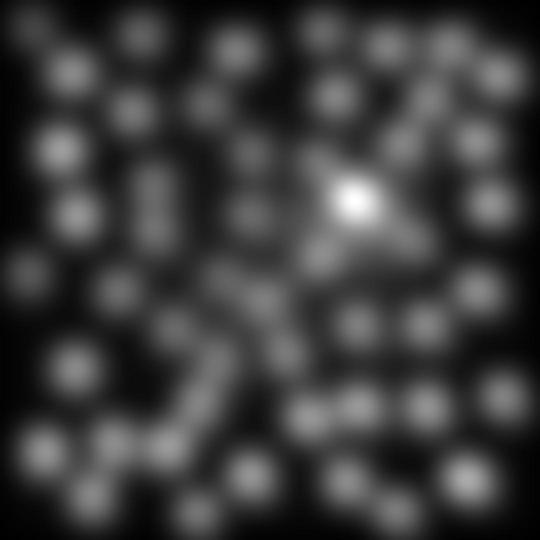} &
				\includegraphics[width=.12\textwidth]{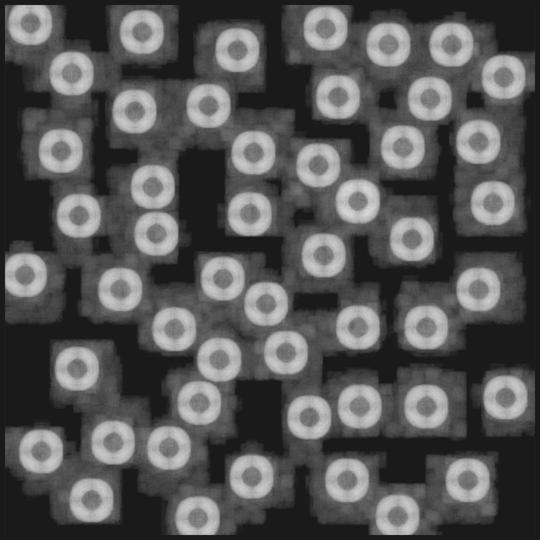}   &
				\includegraphics[width=.12\textwidth]{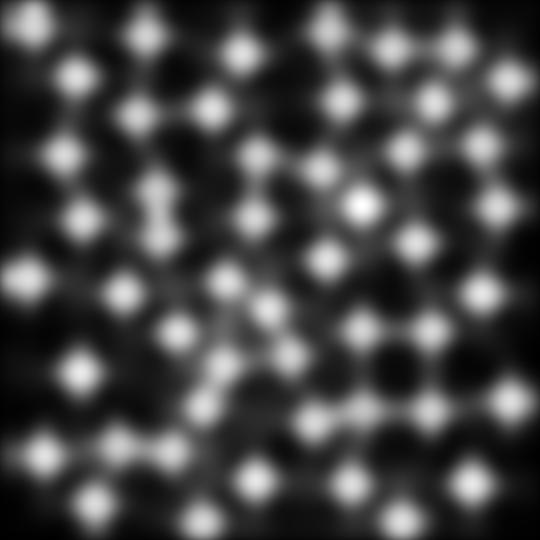}   &
				\includegraphics[width=.12\textwidth]{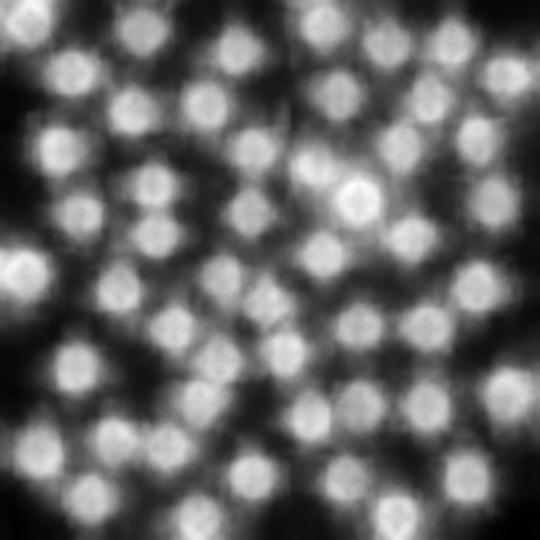}   &
				\includegraphics[width=.12\textwidth]{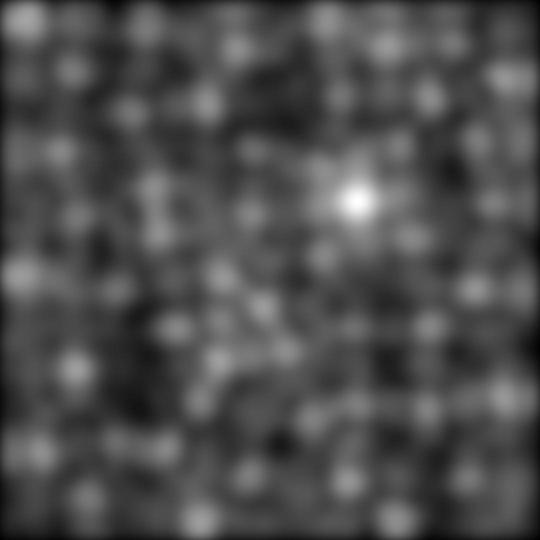}  \\
				\includegraphics[width=.12\textwidth]{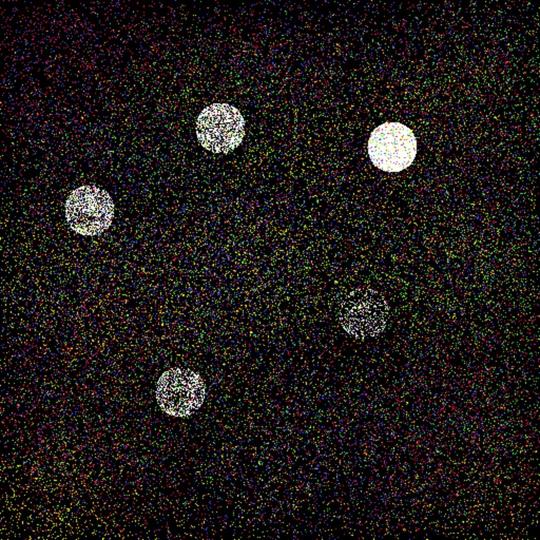} &
				\includegraphics[width=.12\textwidth]{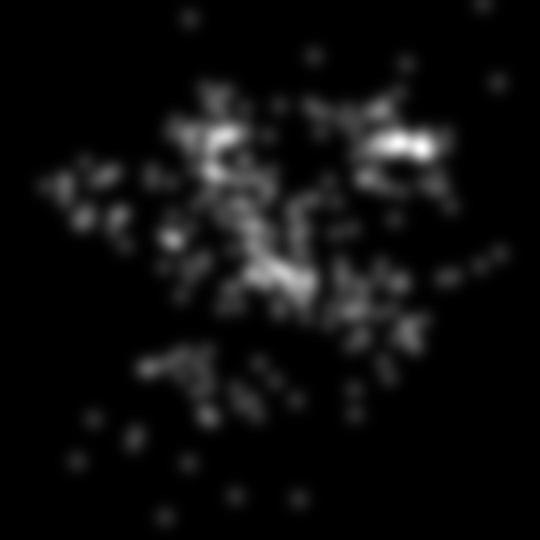} &
				\includegraphics[width=.12\textwidth]{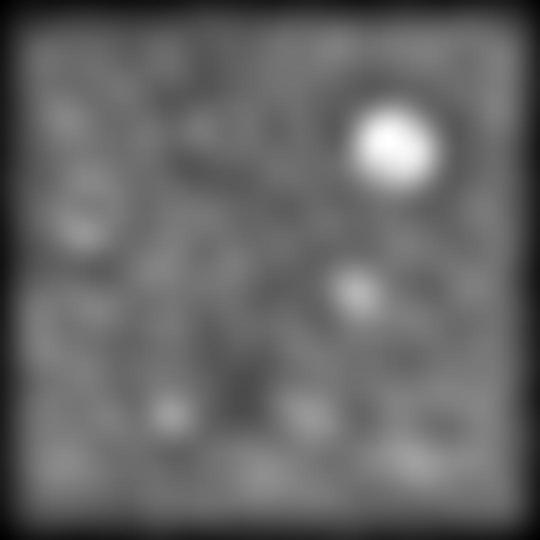} &
				\includegraphics[width=.12\textwidth]{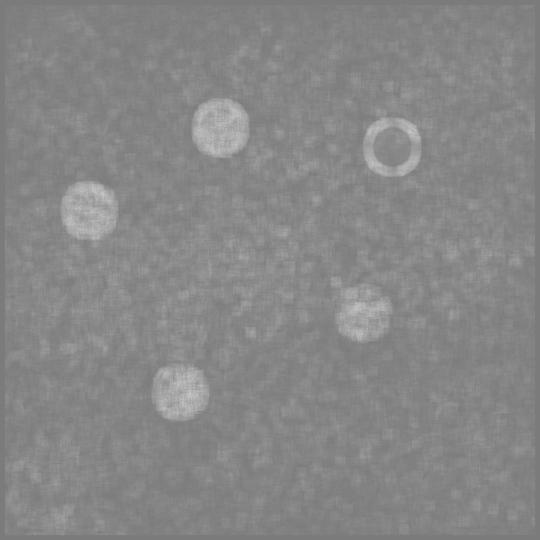} &
				\includegraphics[width=.12\textwidth]{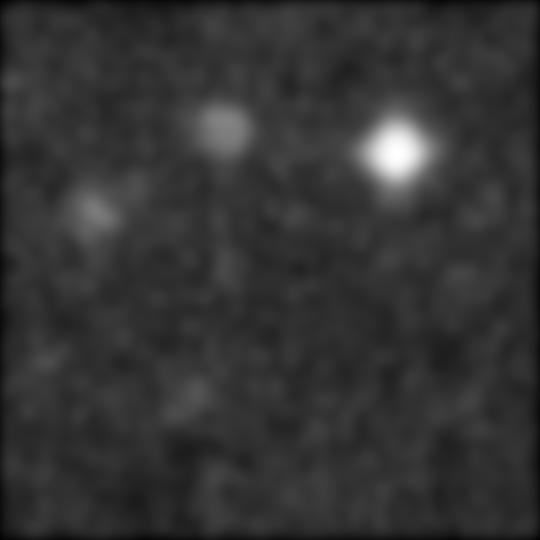} &
				\includegraphics[width=.12\textwidth]{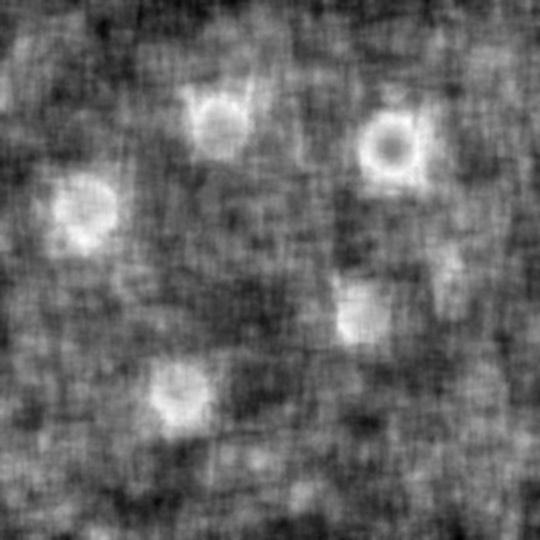} &
				\includegraphics[width=.12\textwidth]{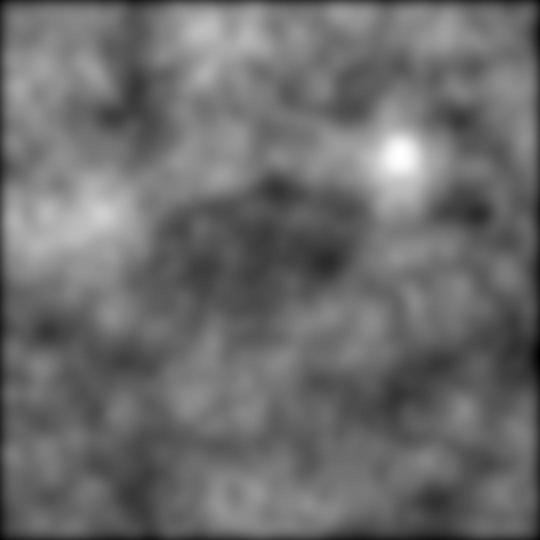} \\
				\includegraphics[width=.12\textwidth]{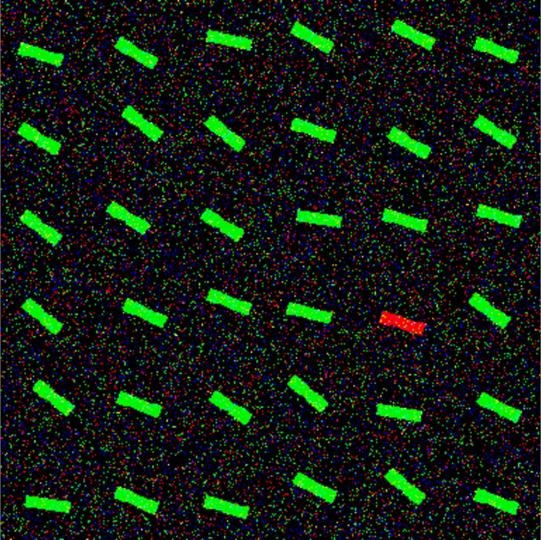} &
				\includegraphics[width=.12\textwidth]{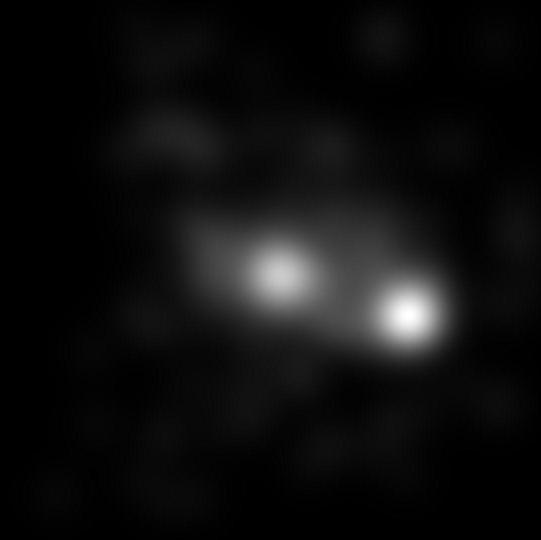} &
				\includegraphics[width=.12\textwidth]{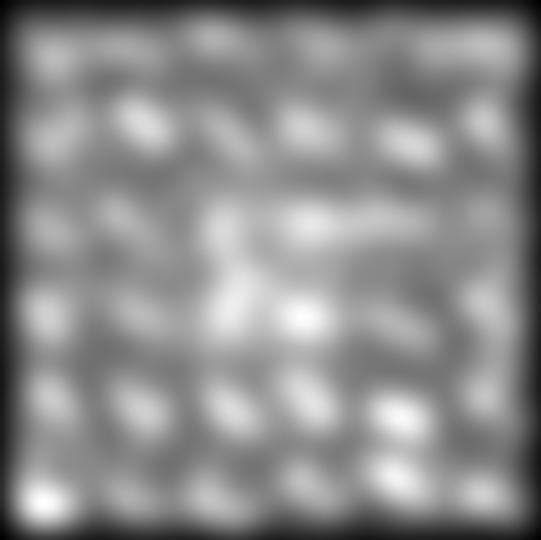} &
				\includegraphics[width=.12\textwidth]{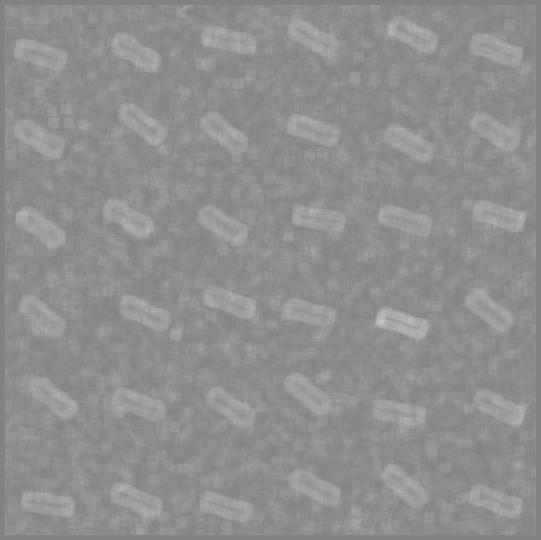}   &
				\includegraphics[width=.12\textwidth]{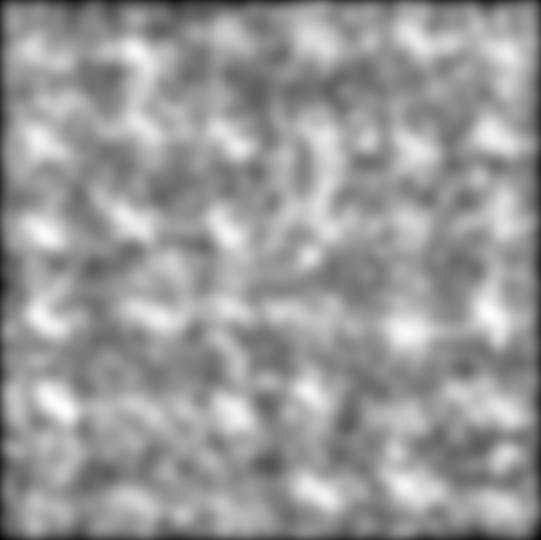}   &
				\includegraphics[width=.12\textwidth]{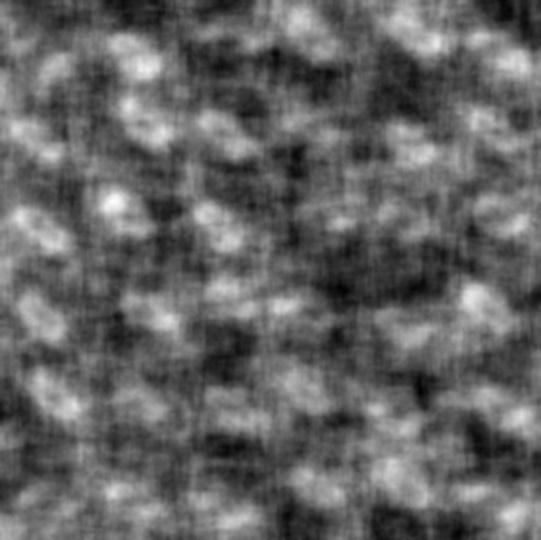}   &
				\includegraphics[width=.12\textwidth]{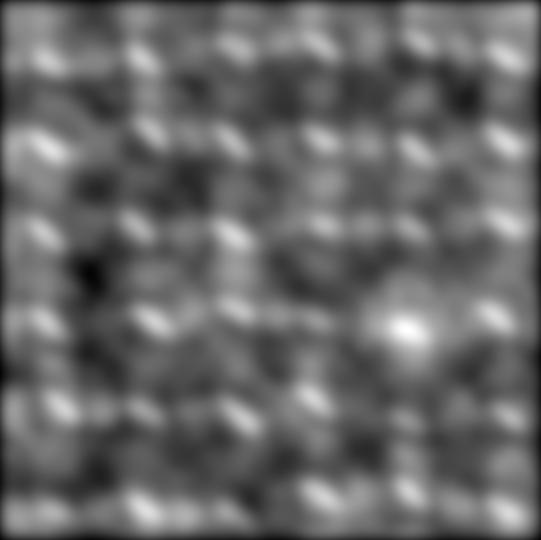}  \\
			\end{tabular}
		}
		\caption{For images showing distinct contexts (first column), we show the eye-tracking psychophysical experimentation (GT, column 2)} and several examples of saliency maps from Itti et al. (IKN), Bruce \& Tsotsos (AIM), Saliency WAvelet Model (SWAM), Murray et al.'s model (SIM) and our Neurodynamic model (columns 5 to 7, respectively).
		\label{fig:qualitative}
	\end{figure*}

	Our results show that our model performance is similar to other saliency models, outperforming previous Murray's SIM model for the cases of SID4VAM, CAT2000 and KTH (\hyperref[tab:kth]{Tables \ref*{tab:sid4vam}, \ref*{tab:cat2000} and \ref*{tab:kth}}), corresponding to synthetic and nature images, as well as showing stable metric scores for distinct contexts (similarly as AWS and GBVS). NSWAM outperforms SWAM as well as other biologically-inspired models (IKN, AIM, SSR \& SIM) specially for metrics that account for center biases. These center biases are qualitatively present even for images where the salient region is conspicuous \hyperref[fig:qualitative]{(Fig. \ref*{fig:qualitative}, rows 8-9)}. 
	
	Saliency models that compute high-level visual features are shown to perform better with real image scenes (\hyperref[tab:toronto]{Table \ref*{tab:toronto}}). However, the image contexts that lack of high-level visual information should be more representative indicators of saliency, due to the absence of semantically or contextually-relevant visual information (nature images), or to be characterized to uniquely contain low-level features (synthetic images) presenting clear pop-out spots to direct participants fixations (which would cause lower inter-participant differences and therefore lower center biases).

	Although AWS and GBVS perform better on predicting fixations at distinct contexts, we remark the plausibility of our unified design for modeling distinct \hyperref[sec:multitask]{HVS' functionality}. NSWAM shows a new insight of applying a more biologically plausible computation of the aforementioned steps. First, we transform image values to color opponencies, found in RGC. Second, we model LGN projections to V1 simple cells using a multiresolution wavelet transform. Third, conspicuity is computed with the Penacchio's dynamical model of the lateral interactions between these cells. Fourth, these channels are integrated to a unique map which will represent SC activity. Using a neurodynamic model with firing-rate neurons allows a more detailed understanding of the dependency of saliency on lateral connections and a potential further study in terms of single neuron dynamics using real image scenes.
	
	\begin{table} [h]
		\centering
		\includegraphics[clip,trim=1.9cm 19.5cm 2cm 2.6cm, width=\linewidth,height=5cm,keepaspectratio]{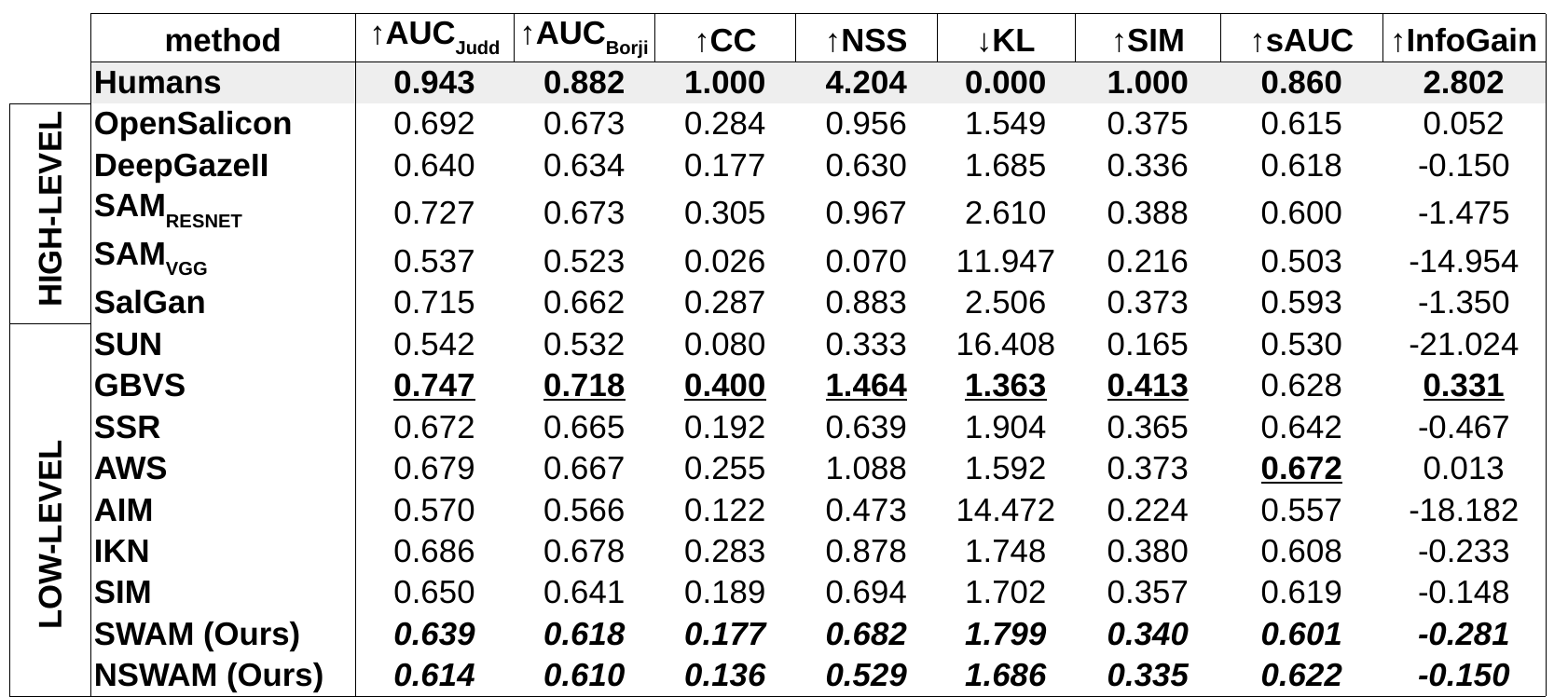}
		\caption{Results for prediction metrics (columns) with SID4VAM dataset \cite{Berga2018a} with synthetic images for different computational models (rows). An up/down arrow ($\uparrow/\downarrow$) besides a metric name means that the highest/lowest the value of this metric, the better the prediction of the particular method. Best results for every metric is shown in bold and underlined. We can see that the GBVS method is usually the one obtaining the best results. Our models (SWAM and NSWAM) are shown in the last rows in bold and italics.}
		\label{tab:sid4vam}
	\end{table}
	
	\begin{table} [h]
		\centering
		\includegraphics[clip,trim=1.9cm 19.5cm 2cm 2.6cm, width=\linewidth,height=5cm,keepaspectratio]{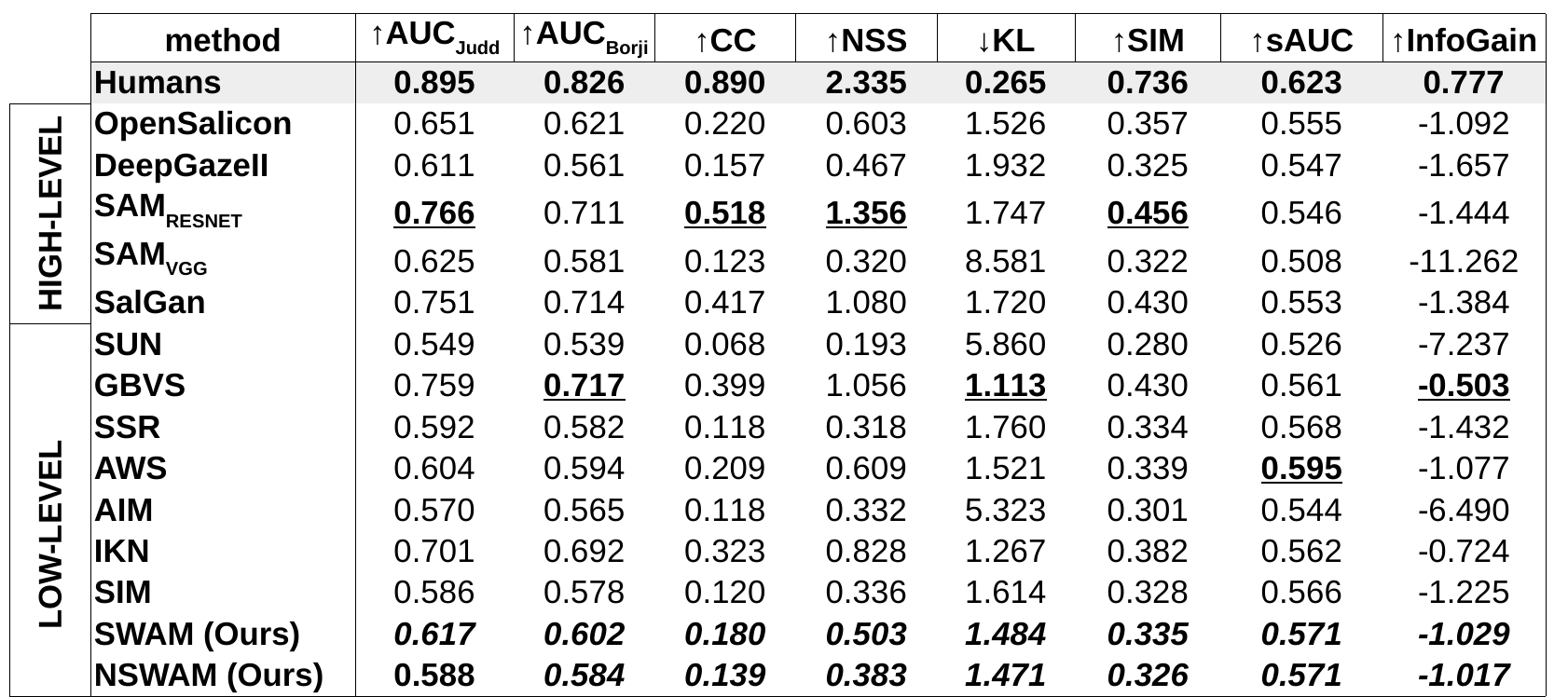}
		\caption{Results for prediction metrics with CAT2000 dataset \cite{CAT2000} training subset (Pattern) of uniquely synthetic images. Best results for every metric is shown in bold and underlined}.
		\label{tab:cat2000}
	\end{table}
	\begin{table} [h]
		\centering
		\includegraphics[clip,trim=1.9cm 19.5cm 2cm 2.6cm, width=\linewidth,height=5cm,keepaspectratio]{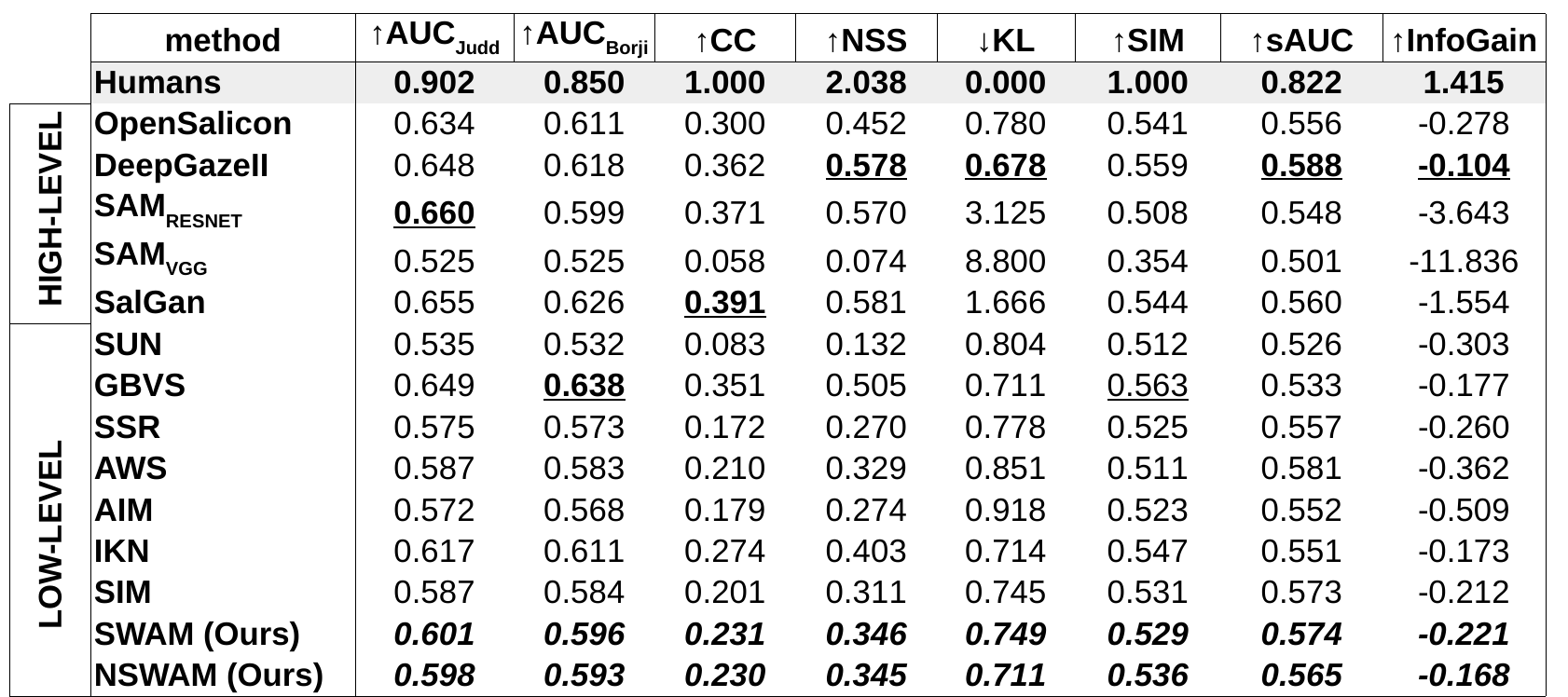}
		\caption{Results for prediction metrics with KTH dataset \cite{Kootstra2011} subset of uniquely nature images. Best results for every metric is shown in bold and underlined}
		\label{tab:kth}
	\end{table}
	\begin{table}[H]
		\centering
		\includegraphics[clip,trim=1.9cm 19.5cm 2cm 2.6cm, width=\linewidth,height=5cm,keepaspectratio]{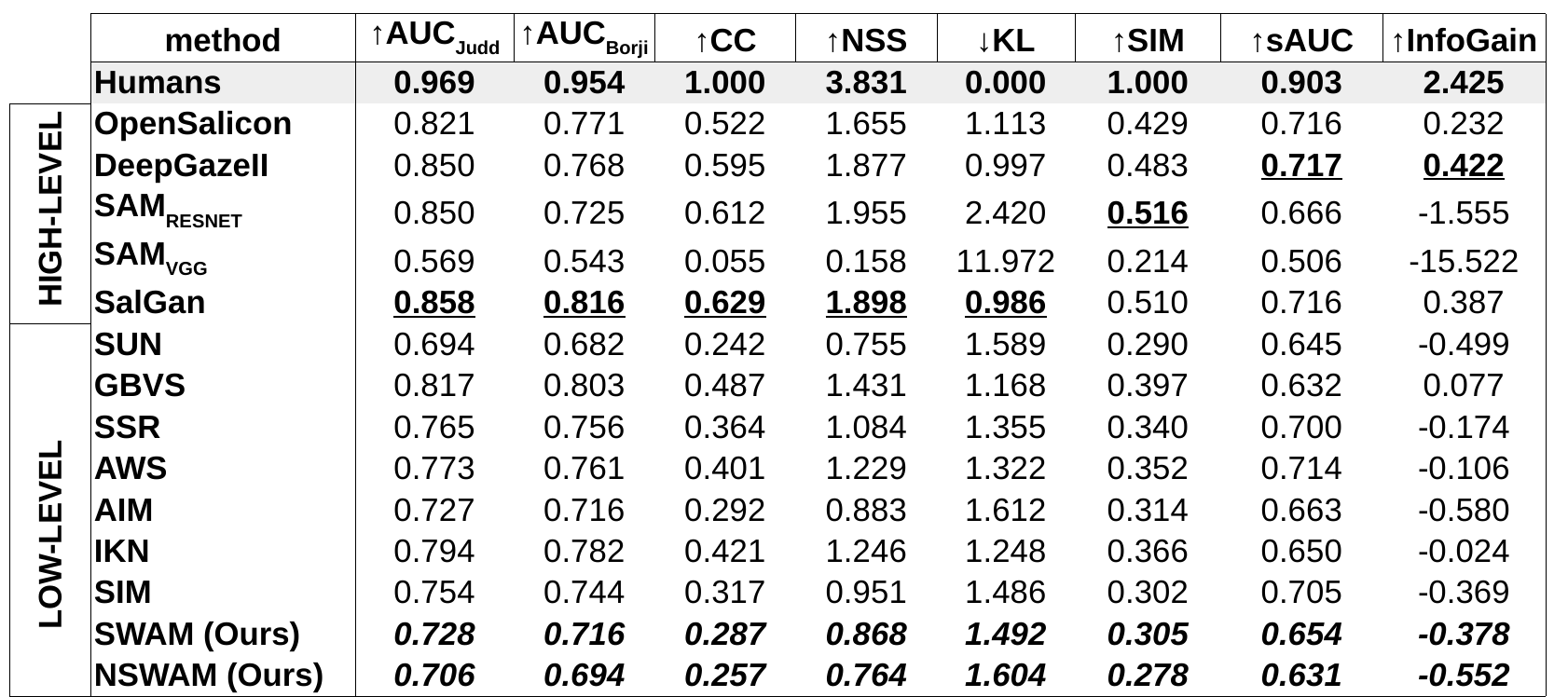}
		\caption{Results for prediction metrics with Toronto dataset \cite{Bruce2009}, corresponding to real (indoor and outdoor) images. Best results for every metric is shown in bold and underlined.}
		\label{tab:toronto}
	\end{table}

	\subsection{Psychophysical study with low-level visual features}
	
	In the previous section we studied the accuracy of the computational architecture to predict eye movements for natural images. But one of the open questions is how every low-level visual feature, e.g. contrast, size, orientation, etc, contributes to conspicuity of feature maps. Acknowledging that the HVS process visual information according to the visual context, human performance on detecting a salient object on a scene may also vary according to the visual properties of such object. With a synthetic image dataset \cite{Berga2018a}\cite{Berga2019c} a specific analysis of how each individual feature influences saliency can be done. In this study we will show how fixation data is predicted when varying feature contrast, concretely on parametrizing Set Size, and Brightness, Color, Size and Orientation contrast between a target salient object \hyperref[fig:example]{Fig. \ref*{fig:example}B} and the rest of distractors (feature singleton search).
	
	
	In order to quantitatively estimate the accuracy of the computational model predictions, we used the shuffled AUC (sAUC) metric. It computes the area under ROC considering TP as fixations inside the saliency map, similarly to the AUC metric. In contrast to AUC, sAUC does not evaluate FP at random areas of the image but instead uses fixations inside other random images from the same dataset over several trials (10 by default). The sAUC metric gives a more accurate evaluation of predicted maps with respect human fixations but penalizing for higher model center biases (which are or can be present for distinct images in the ground truth).
	
	
	\subsubsection{Brightness differences}
	
\begin{figure}[h!]
		\centering
		\setlength{\fboxsep}{0pt}%
		\setlength{\fboxrule}{0.5pt}%
		\begin{subfigure}{\linewidth} \centering A
			\fbox{\includegraphics[width=.12\textwidth]{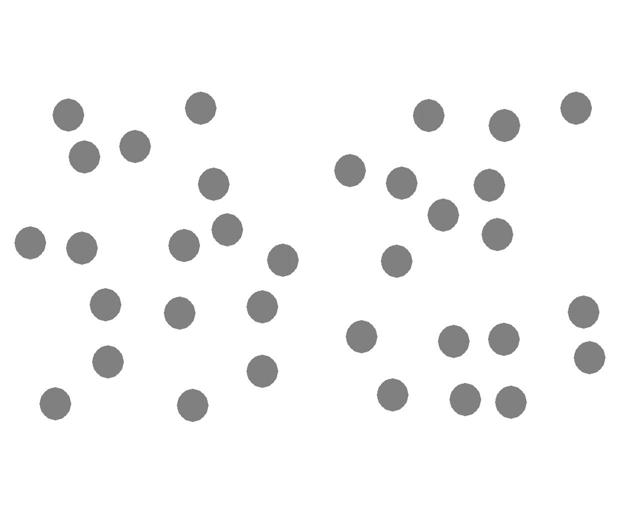}}
			\fbox{\includegraphics[width=.12\textwidth]{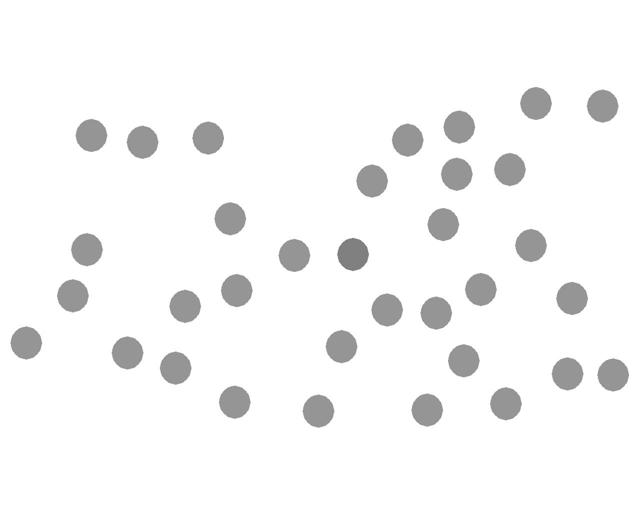}}
			\fbox{\includegraphics[width=.12\textwidth]{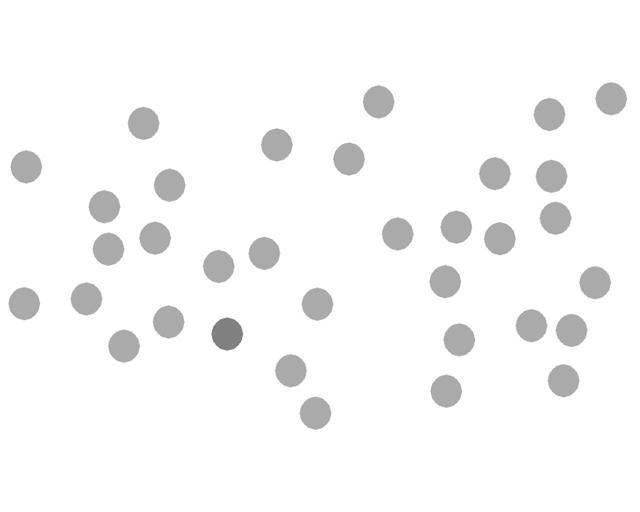}}
			\fbox{\includegraphics[width=.12\textwidth]{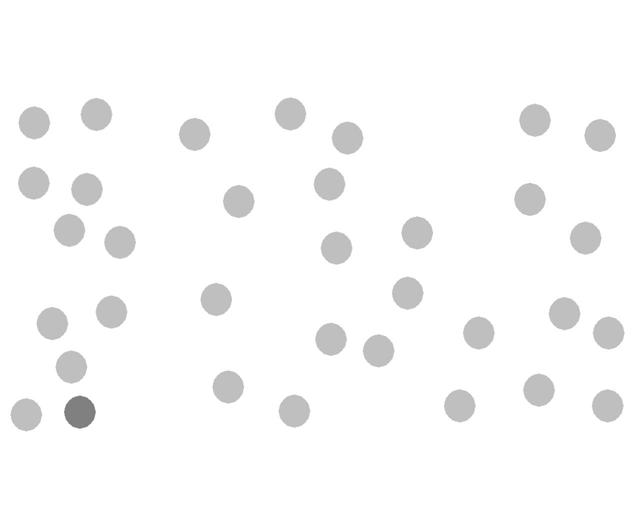}}
			\fbox{\includegraphics[width=.12\textwidth]{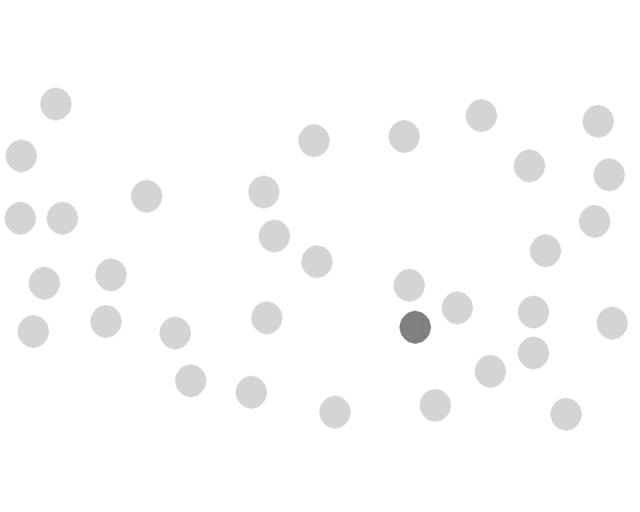}}
			\fbox{\includegraphics[width=.12\textwidth]{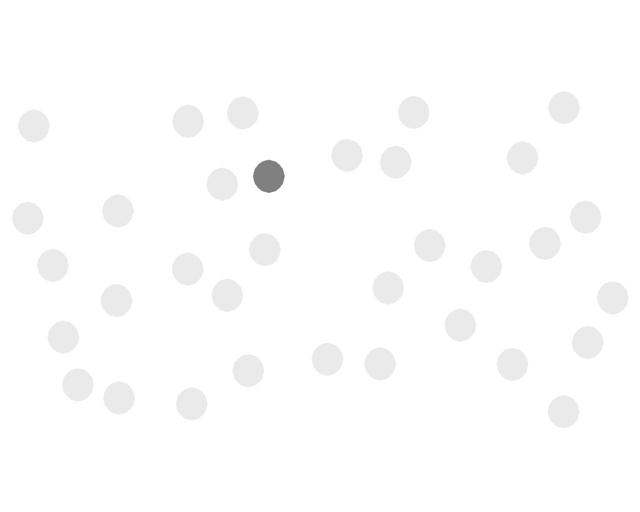}}
			\fbox{\includegraphics[width=.12\textwidth]{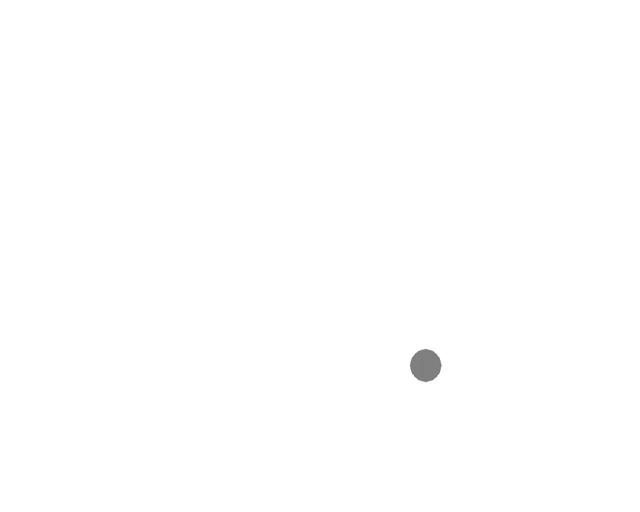}}
			\\ \hspace{2mm}
			\fbox{\includegraphics[width=.12\textwidth]{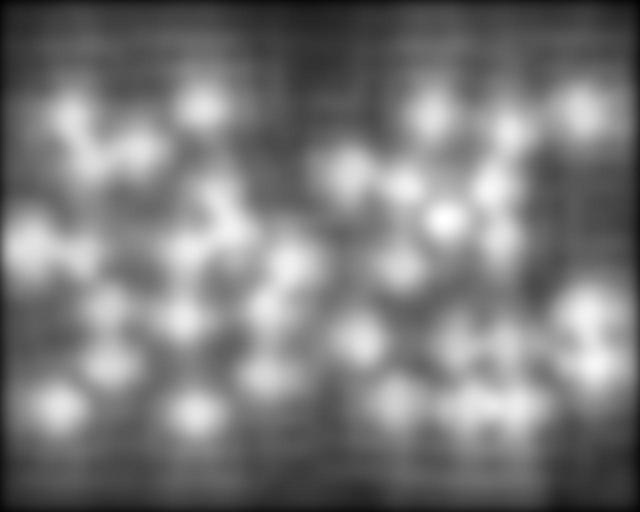}}
			\fbox{\includegraphics[width=.12\textwidth]{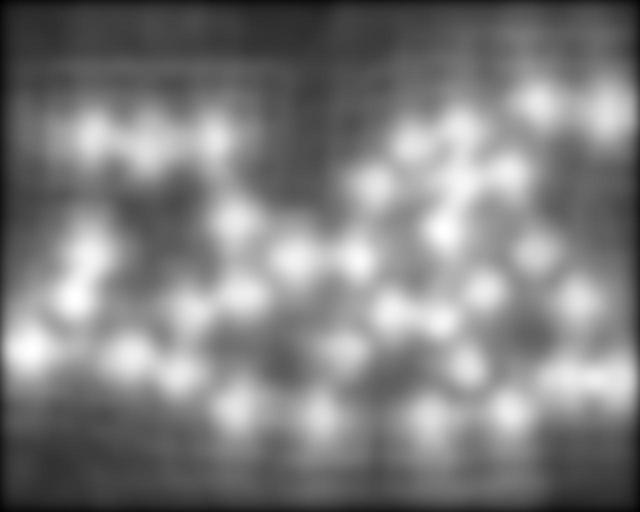}}
			\fbox{\includegraphics[width=.12\textwidth]{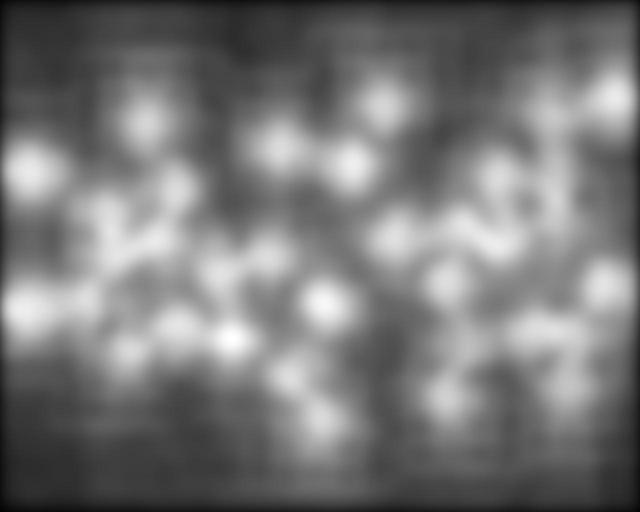}}
			\fbox{\includegraphics[width=.12\textwidth]{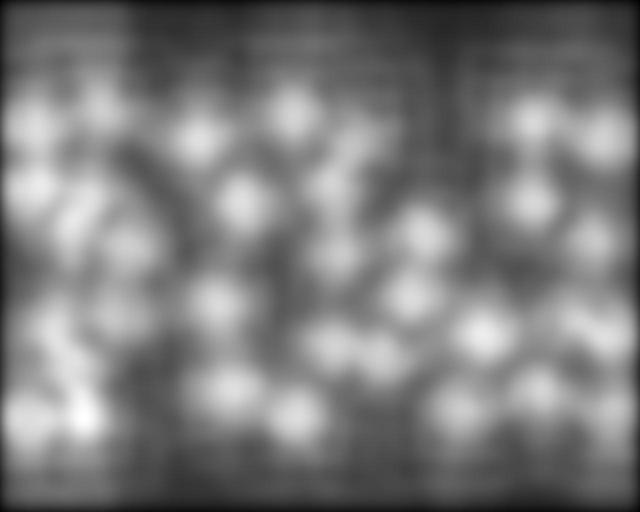}}
			\fbox{\includegraphics[width=.12\textwidth]{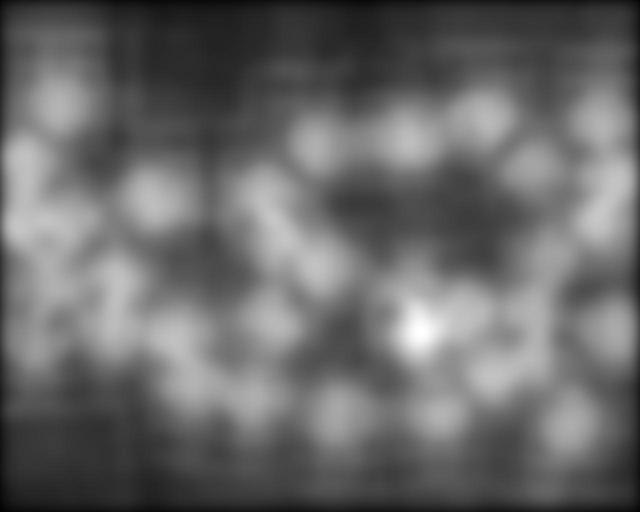}}
			\fbox{\includegraphics[width=.12\textwidth]{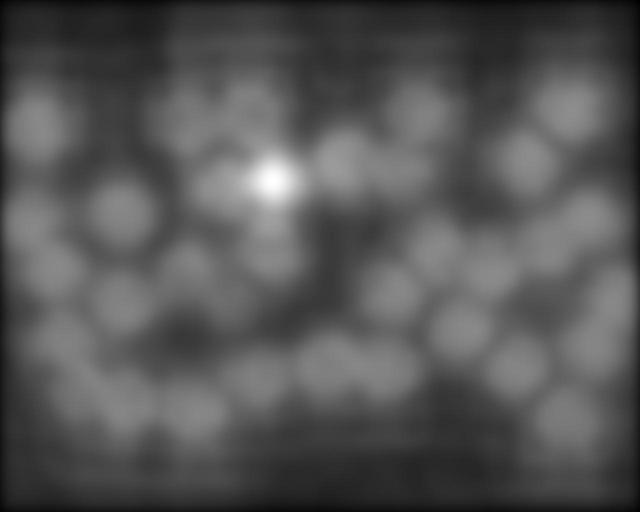}}
			\fbox{\includegraphics[width=.12\textwidth]{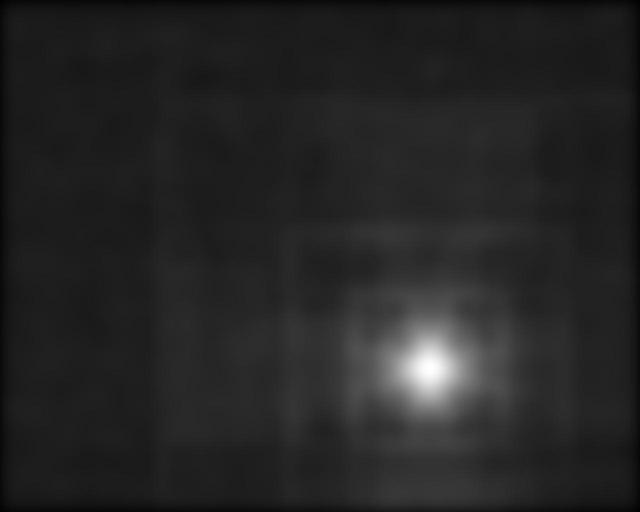}}
			\\    B
			\fbox{\includegraphics[width=.12\textwidth]{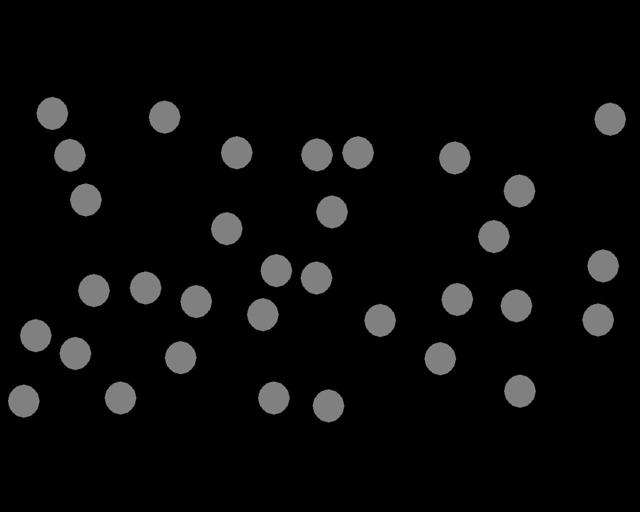}}
			\fbox{\includegraphics[width=.12\textwidth]{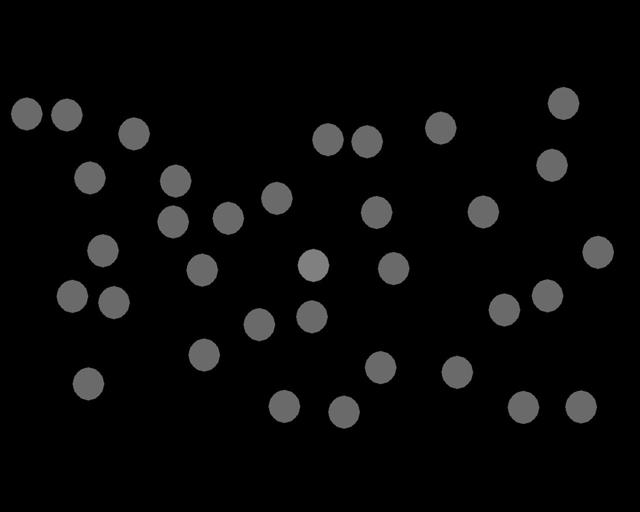}}
			\fbox{\includegraphics[width=.12\textwidth]{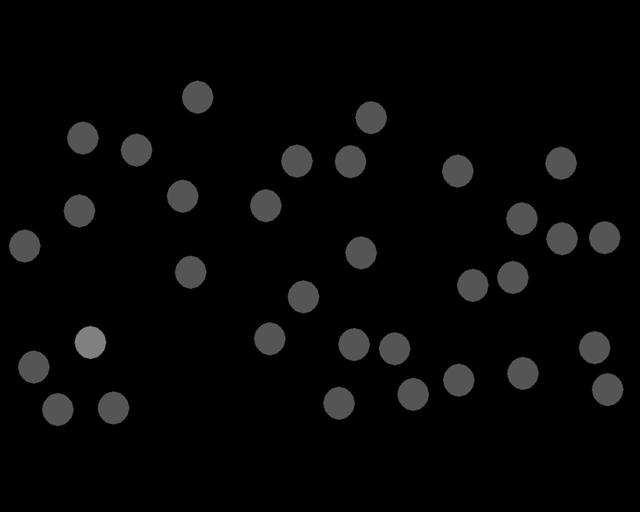}}
			\fbox{\includegraphics[width=.12\textwidth]{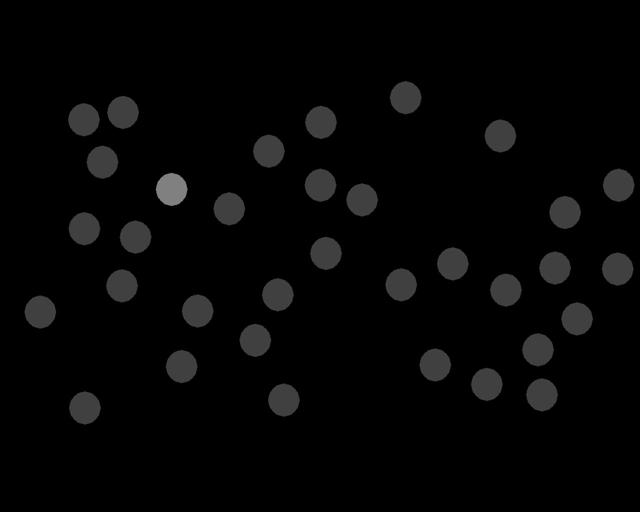}}
			\fbox{\includegraphics[width=.12\textwidth]{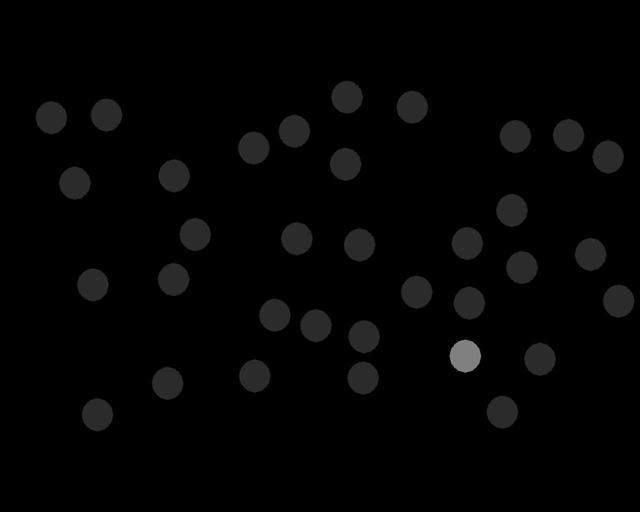}}
			\fbox{\includegraphics[width=.12\textwidth]{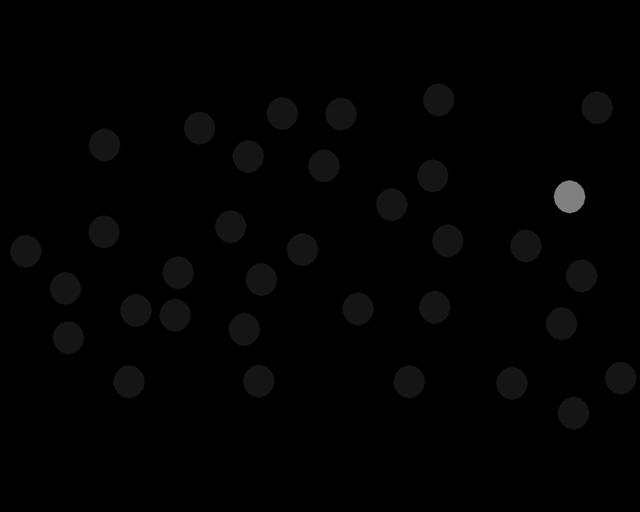}}
			\fbox{\includegraphics[width=.12\textwidth]{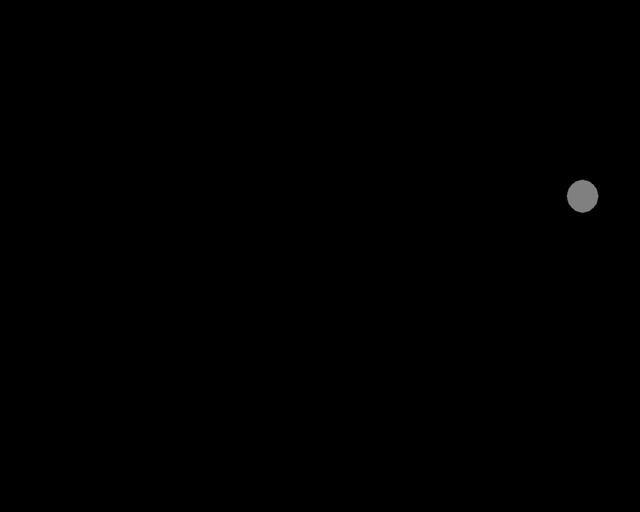}}
			\\      \hspace{2.5mm}
			\fbox{\includegraphics[width=.12\textwidth]{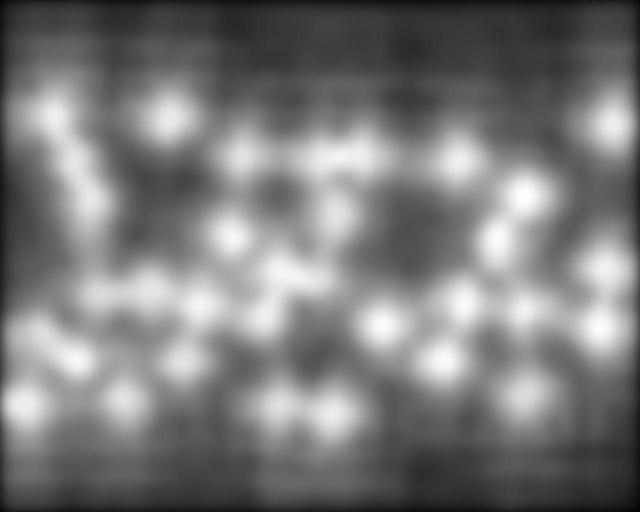}}
			\fbox{\includegraphics[width=.12\textwidth]{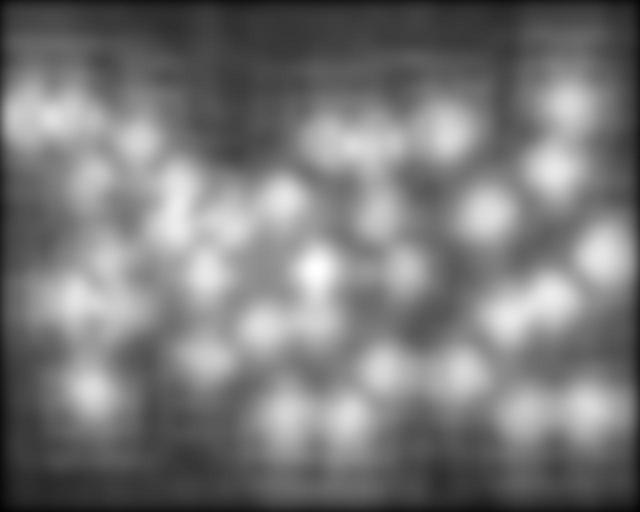}}
			\fbox{\includegraphics[width=.12\textwidth]{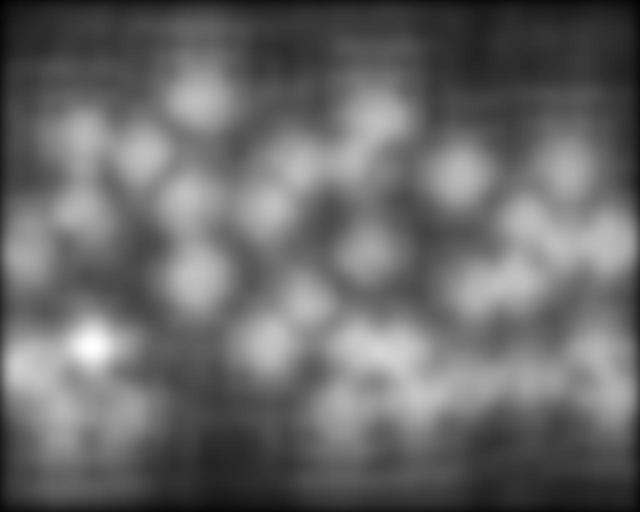}}
			\fbox{\includegraphics[width=.12\textwidth]{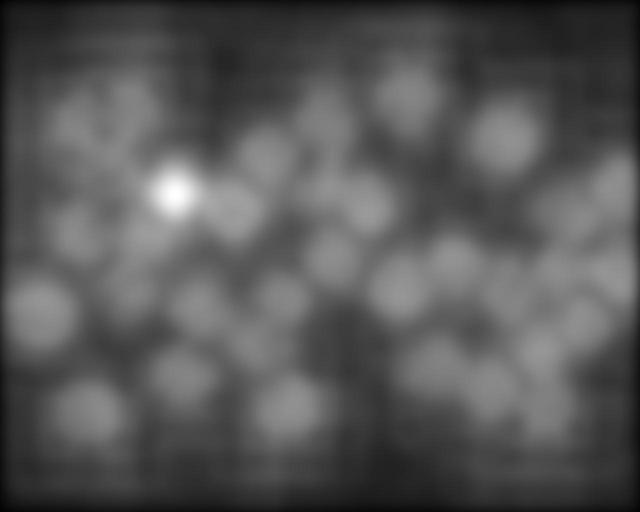}}
			\fbox{\includegraphics[width=.12\textwidth]{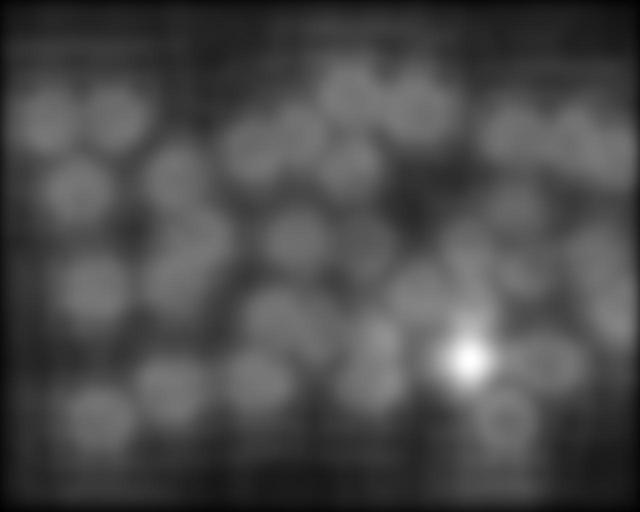}}
			\fbox{\includegraphics[width=.12\textwidth]{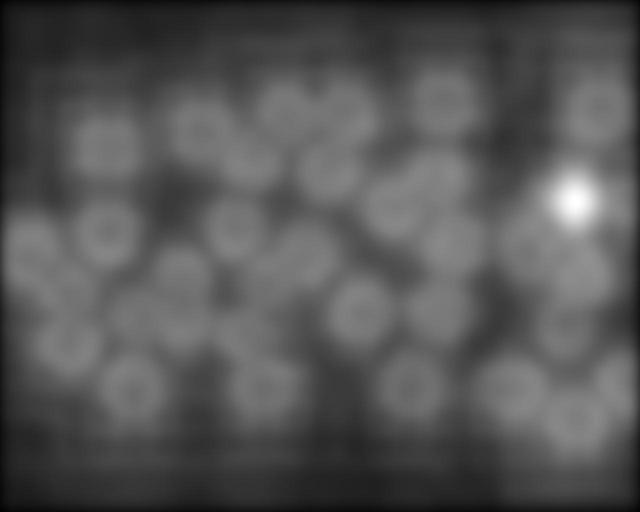}}
			\fbox{\includegraphics[width=.12\textwidth]{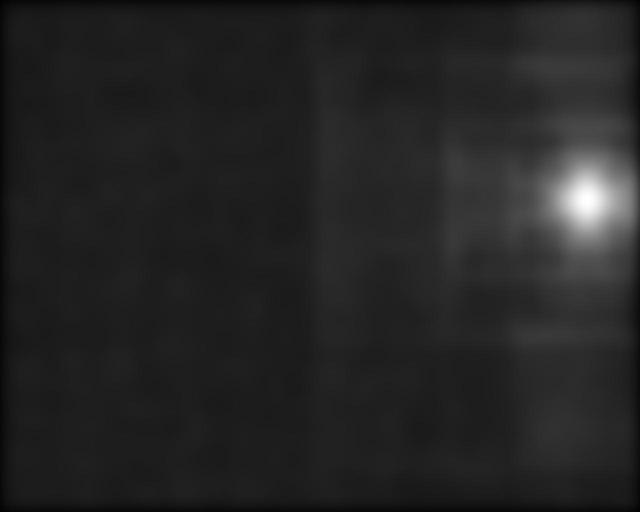}}
			\\
			\hspace{.08\textwidth}0\hspace{.10\textwidth}0.08\hspace{.08\textwidth}0.17\hspace{.06\textwidth}0.25\hspace{.06\textwidth}0.33\hspace{.08\textwidth}0.41\hspace{.08\textwidth}0.5
		\end{subfigure}
		\caption{Synthetic stimuli representing distinct brightness contrasts (HSL luminance differences) from target and distractors ($\Delta L_{D,T}$) with \textbf{(A)} bright background ($L_T=0.5, L_B=1, L_D=0.5..1$) and \textbf{(B)} dark background ($L_T=0.5, L_B=0, L_D=0..0.5$). Rows below \textbf{A,B} are NSWAM predictions.} \label{fig:vs5} 
	\end{figure}

\begin{figure}[h!]
		\centering
		\begin{subfigure}{.9\linewidth}
			\includegraphics[width=\textwidth]{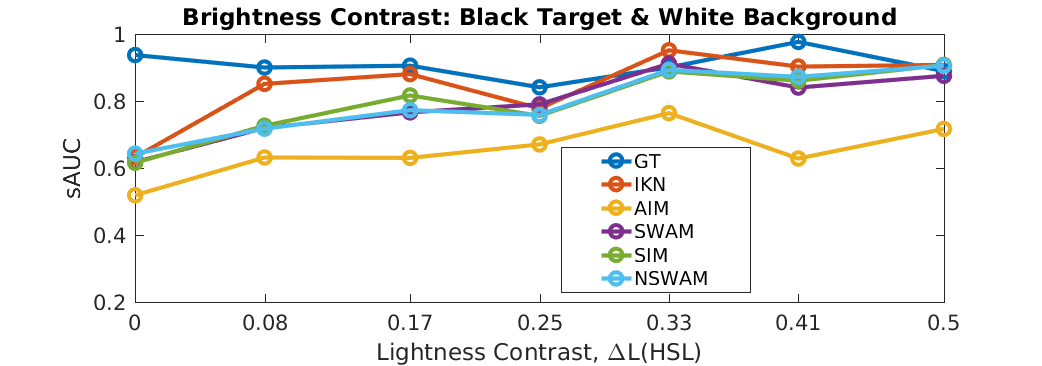}
			\caption*{\centering \textbf{A}}
		\end{subfigure}
		\begin{subfigure}{.9\linewidth}
			\includegraphics[width=\textwidth]{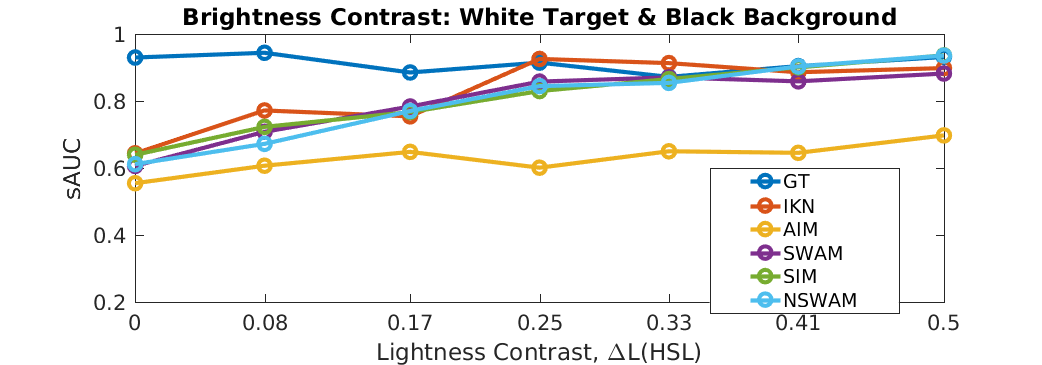}
			\caption*{\centering \textbf{B}}
		\end{subfigure}
		\caption{Results of sAUC upon brightness contrast, $\Delta L_{D,T}$) with \textbf{(A)} bright and \textbf{(B)} dark background. We can see that our models SWAM, SIM and NSWAM are usually among the best methods.}
		\label{fig:res_b10}
	\end{figure}

	Differences in brightness are major factors for making an object to attract attention. That is, a bright object is less salient as luminance of other surround objects increase (\hyperref[fig:vs5]{Fig. \ref*{fig:vs5}}). Conversely, a dark target in a bright background will be more salient as surround distractors have higher luminance  \cite{Pashler2004}\cite{Nothdurft2000}.  
	NSWAM processes luminance signals separately from chromatic ones using the L channel (feature conspicuity from a distinctively bright object upon a dark background will be processed similarly to a dark object upon a bright background). We compare sAUC metrics for both conditions and NSWAM is shown to acquire similar performance to SIM and SWAM, with higher sAUC than IKN \hyperref[fig:res_b10]{Fig. \ref*{fig:res_b10},A-B}, specially for stimulus with higher contrasts ($\Delta L_{D,T}>.25$). Results on sAUC for NSWAM correlates with brightness contrast, for both cases of bright ($\rho=.941, p=1.6 \times 10^{-3}$) and dark ($\rho=.986, p=4.7 \times 10^{-5}$) background.

	\vfill
	
	\subsubsection{Color differences}
	
\begin{figure}[h!]
		\centering
		\setlength{\fboxsep}{0pt}%
		\setlength{\fboxrule}{0.5pt}%
		\begin{subfigure}{\linewidth} \centering A
			\fbox{\includegraphics[width=.12\textwidth]{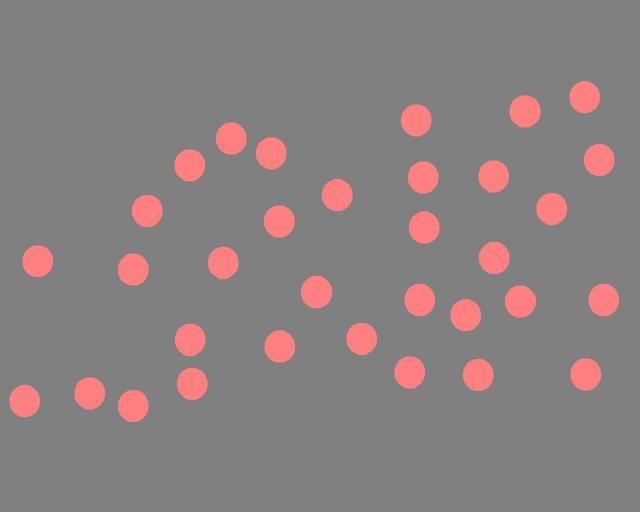}}
			\fbox{\includegraphics[width=.12\textwidth]{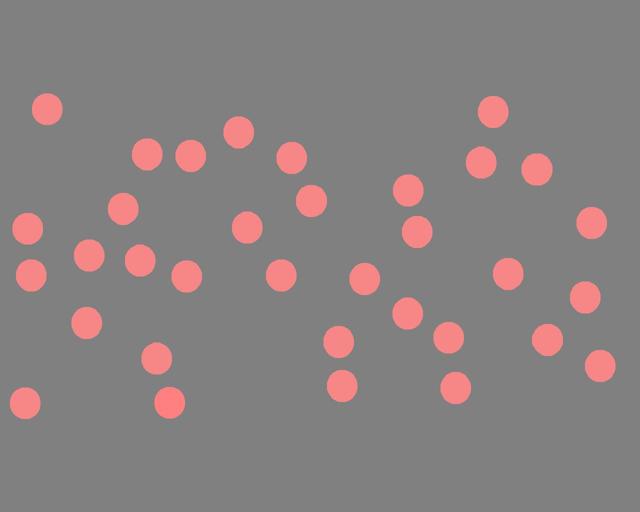}}
			\fbox{\includegraphics[width=.12\textwidth]{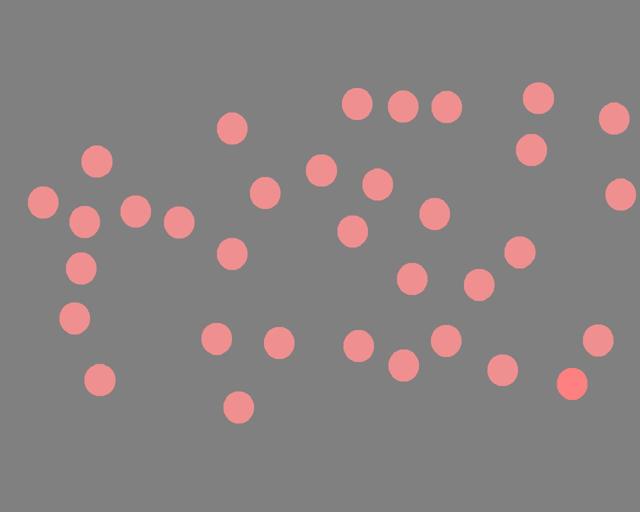}}
			\fbox{\includegraphics[width=.12\textwidth]{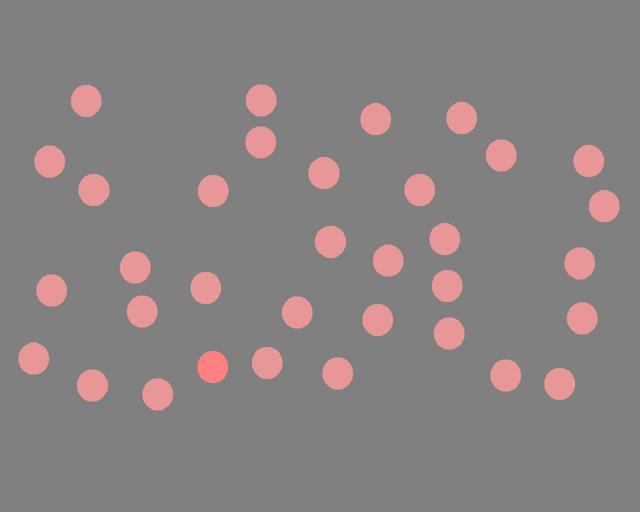}}
			\fbox{\includegraphics[width=.12\textwidth]{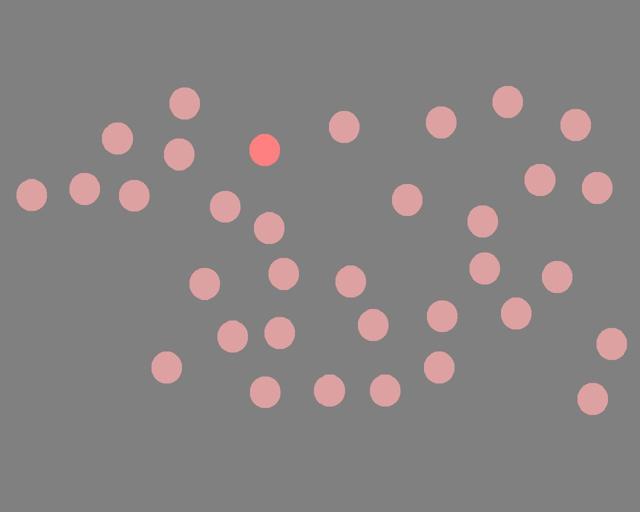}}
			\fbox{\includegraphics[width=.12\textwidth]{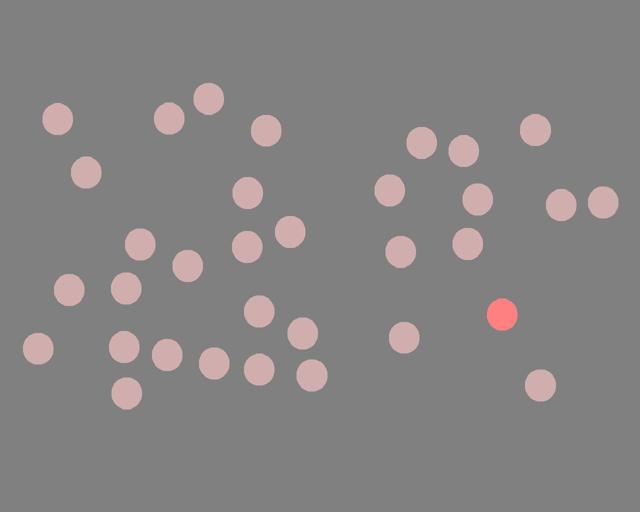}}
			\fbox{\includegraphics[width=.12\textwidth]{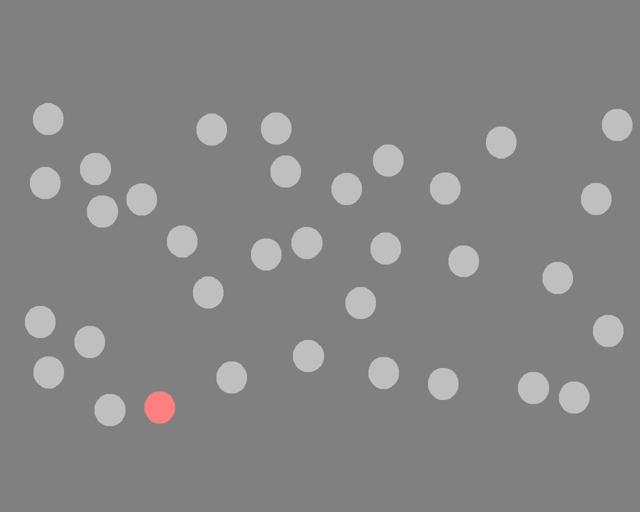}}
			\\   \hspace{2.5mm}
			\fbox{\includegraphics[width=.12\textwidth]{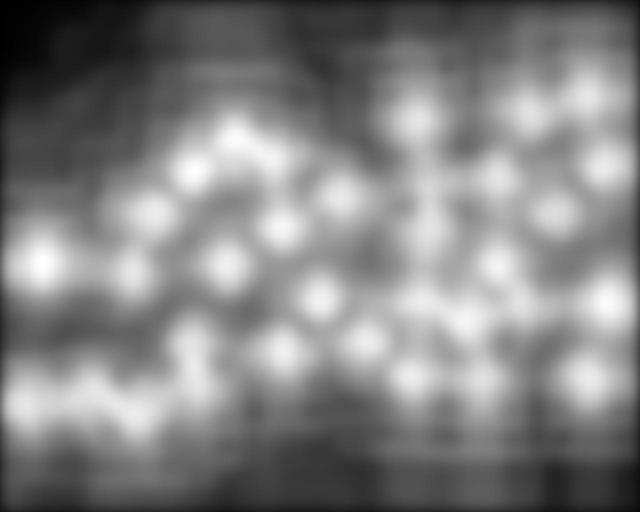}}
			\fbox{\includegraphics[width=.12\textwidth]{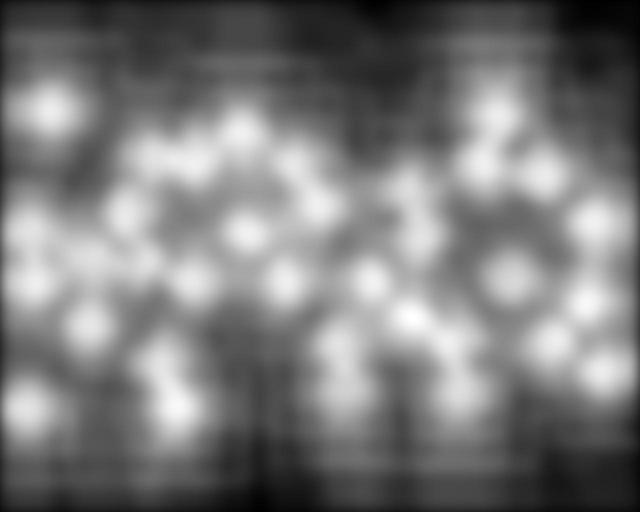}}
			\fbox{\includegraphics[width=.12\textwidth]{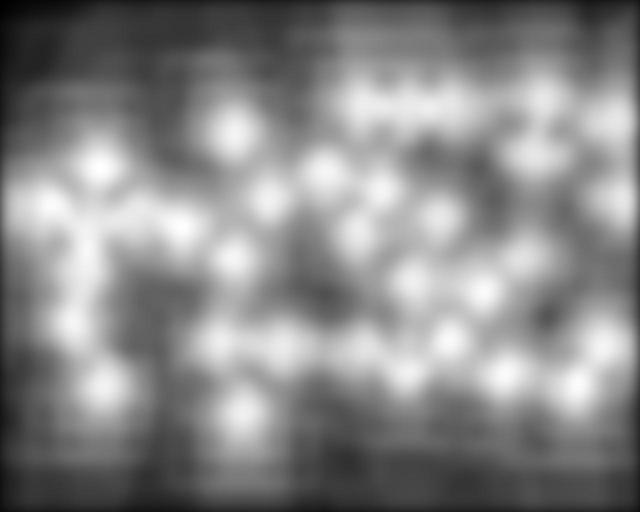}}
			\fbox{\includegraphics[width=.12\textwidth]{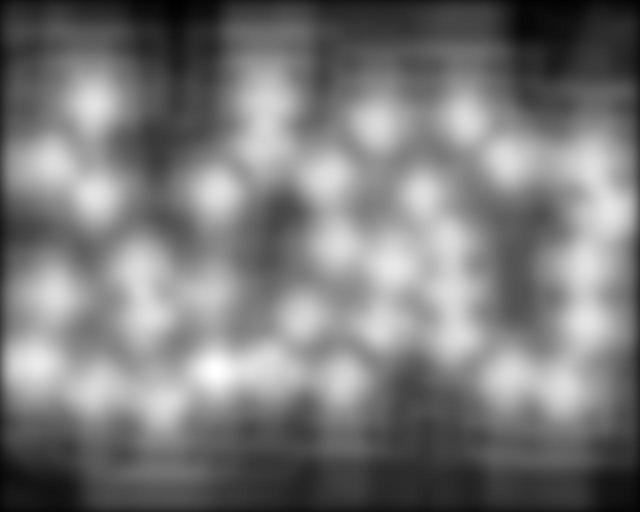}}
			\fbox{\includegraphics[width=.12\textwidth]{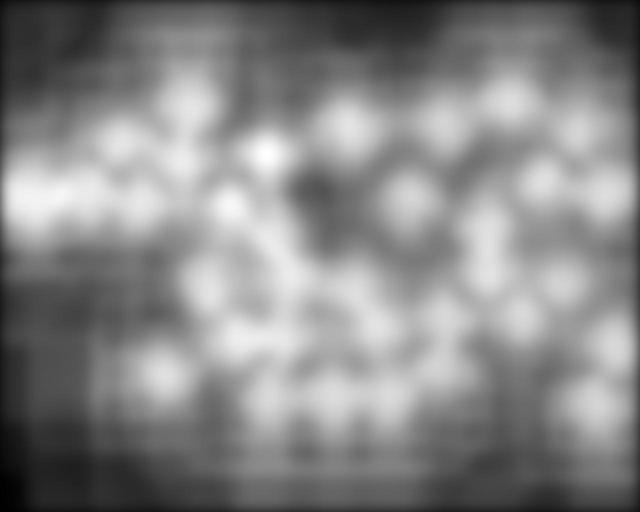}}
			\fbox{\includegraphics[width=.12\textwidth]{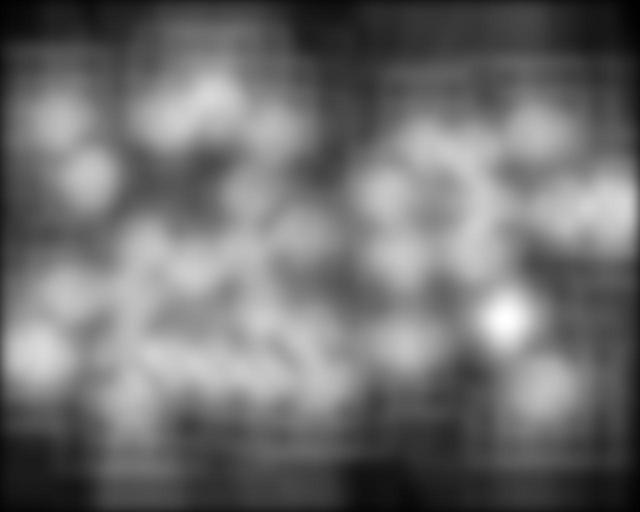}}
			\fbox{\includegraphics[width=.12\textwidth]{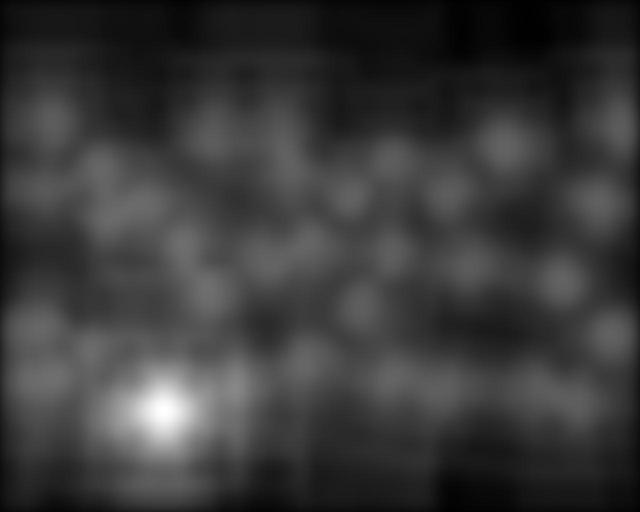}}
			\\   B
			\fbox{\includegraphics[width=.12\textwidth]{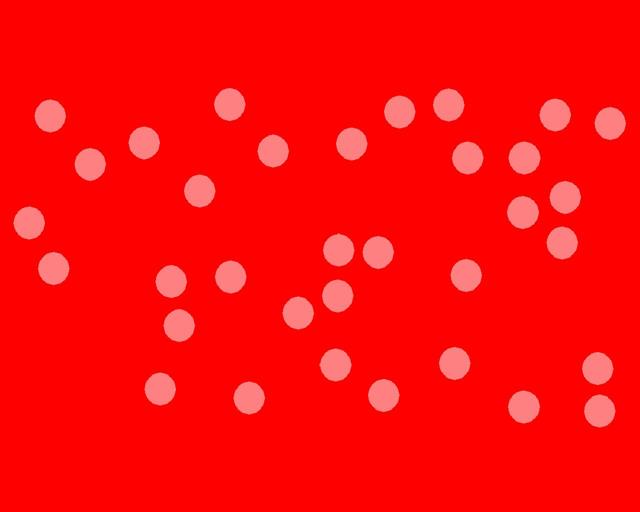}}
			\fbox{\includegraphics[width=.12\textwidth]{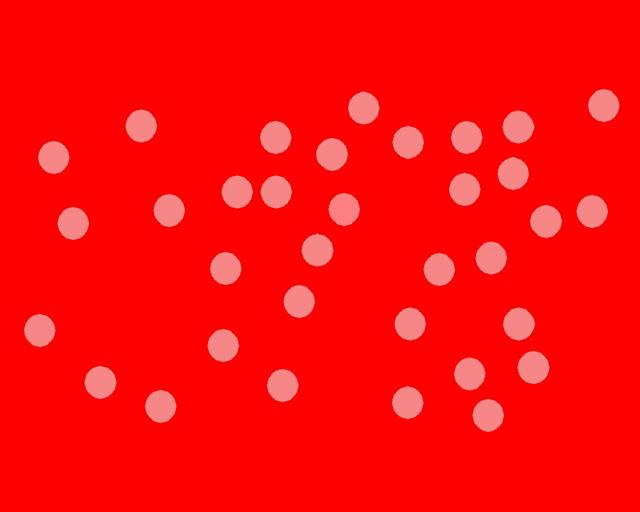}}
			\fbox{\includegraphics[width=.12\textwidth]{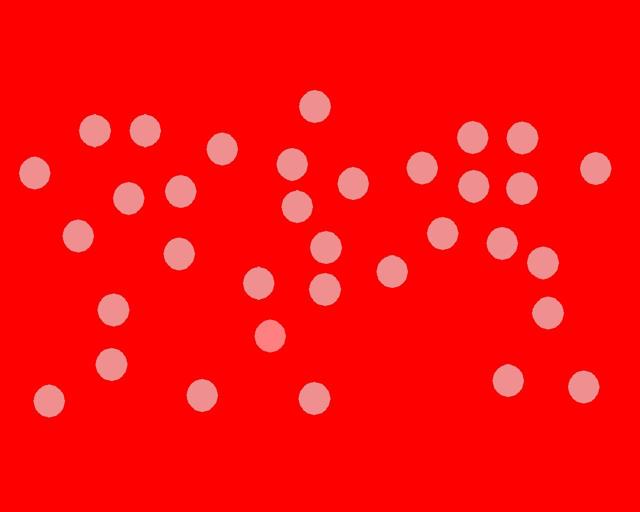}}
			\fbox{\includegraphics[width=.12\textwidth]{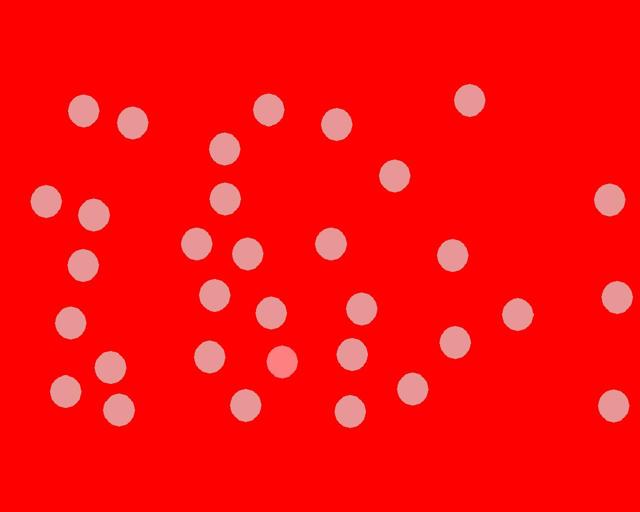}}
			\fbox{\includegraphics[width=.12\textwidth]{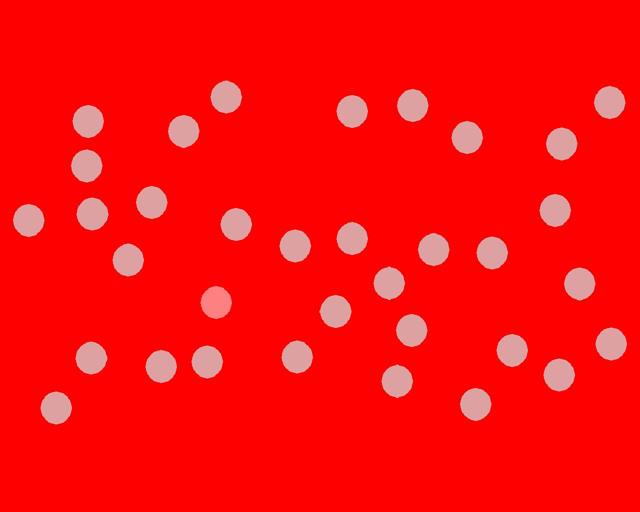}}
			\fbox{\includegraphics[width=.12\textwidth]{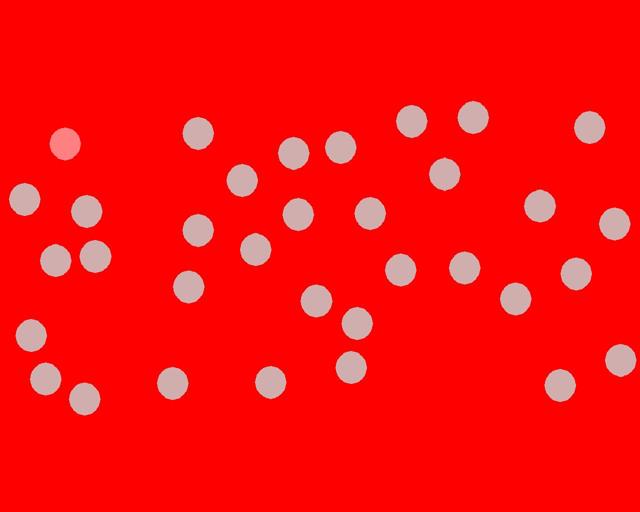}}
			\fbox{\includegraphics[width=.12\textwidth]{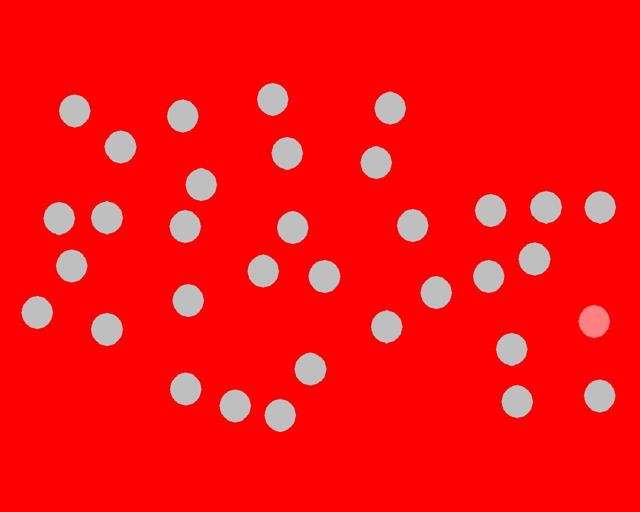}}
			\\   \hspace{2.5mm}
			\fbox{\includegraphics[width=.12\textwidth]{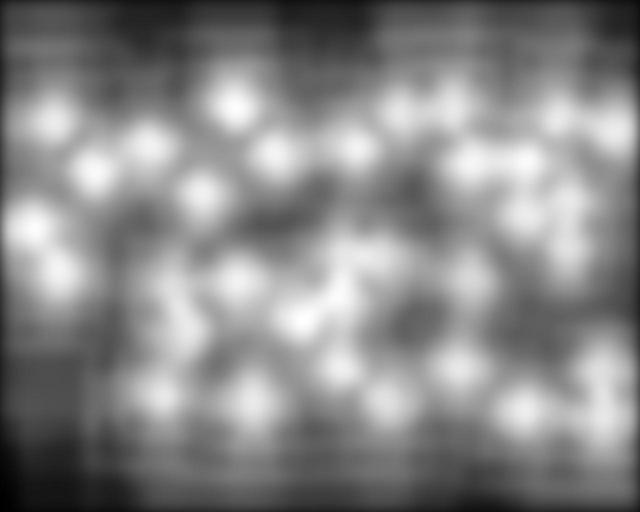}}
			\fbox{\includegraphics[width=.12\textwidth]{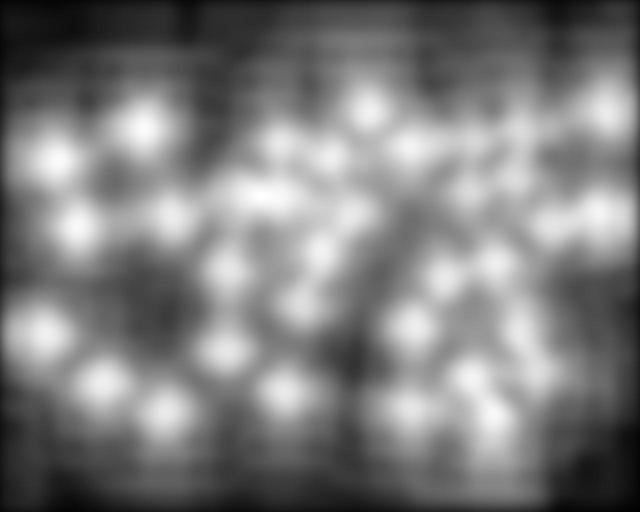}}
			\fbox{\includegraphics[width=.12\textwidth]{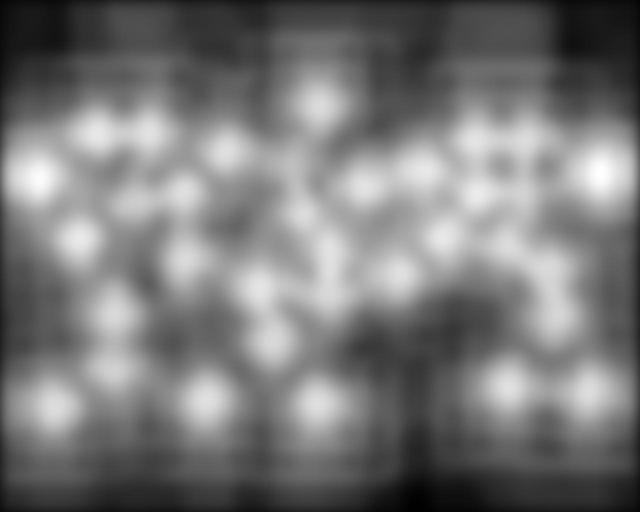}}
			\fbox{\includegraphics[width=.12\textwidth]{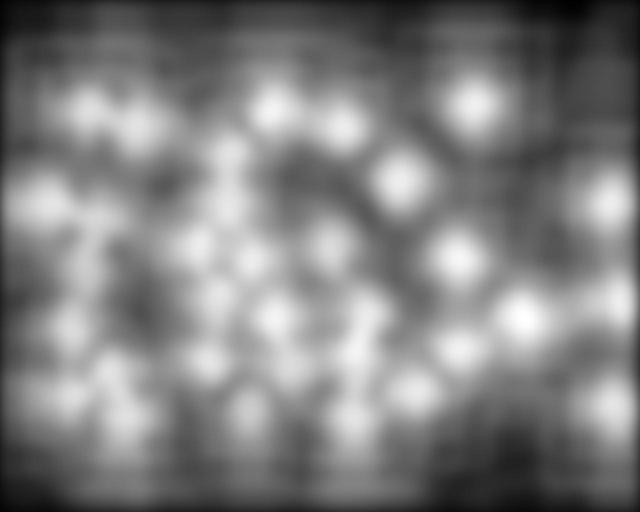}}
			\fbox{\includegraphics[width=.12\textwidth]{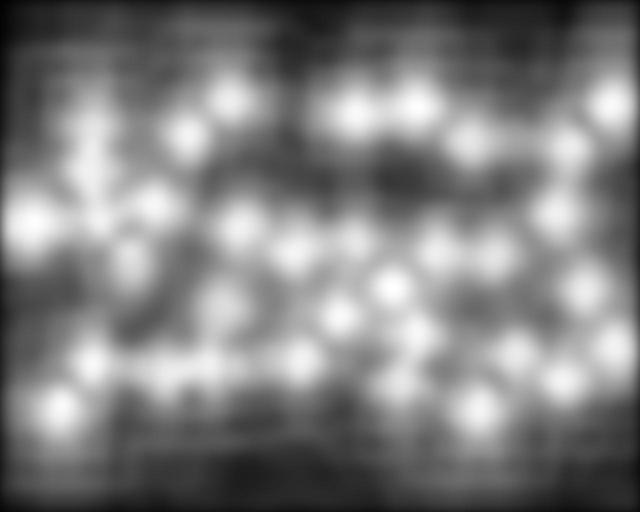}}
			\fbox{\includegraphics[width=.12\textwidth]{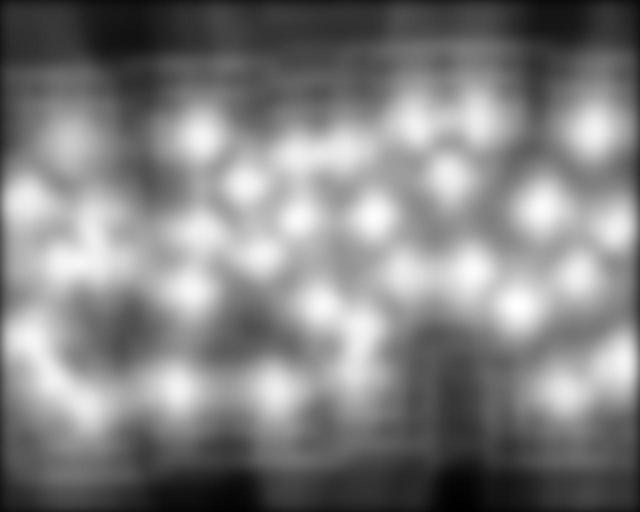}}
			\fbox{\includegraphics[width=.12\textwidth]{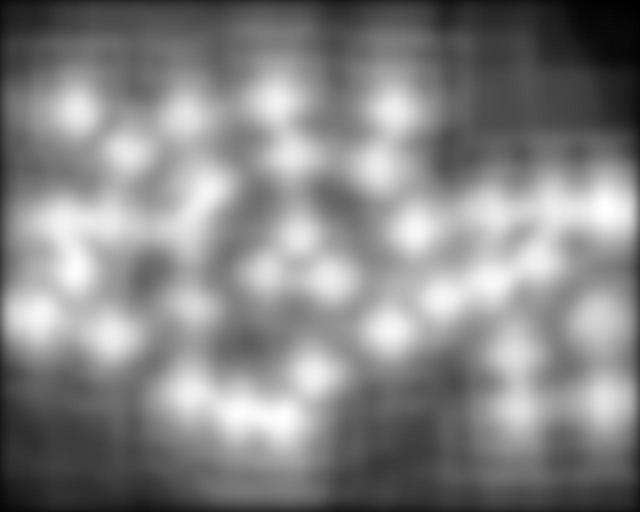}}
			\\ C
			\fbox{\includegraphics[width=.12\textwidth]{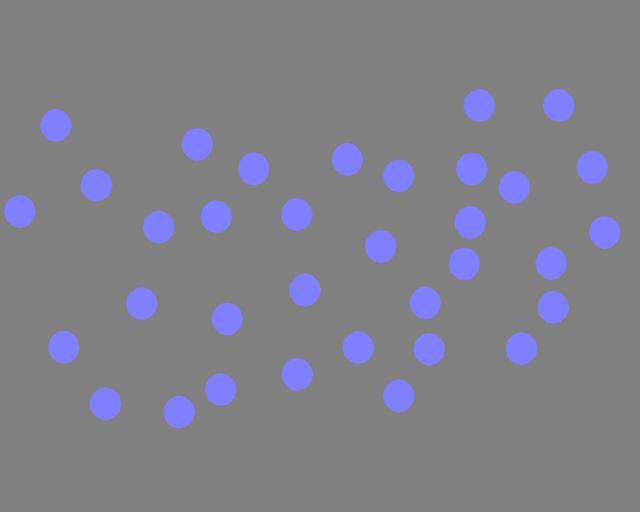}}
			\fbox{\includegraphics[width=.12\textwidth]{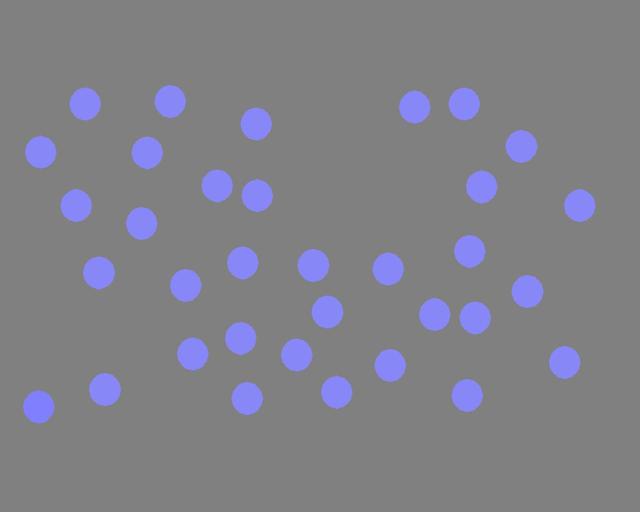}}
			\fbox{\includegraphics[width=.12\textwidth]{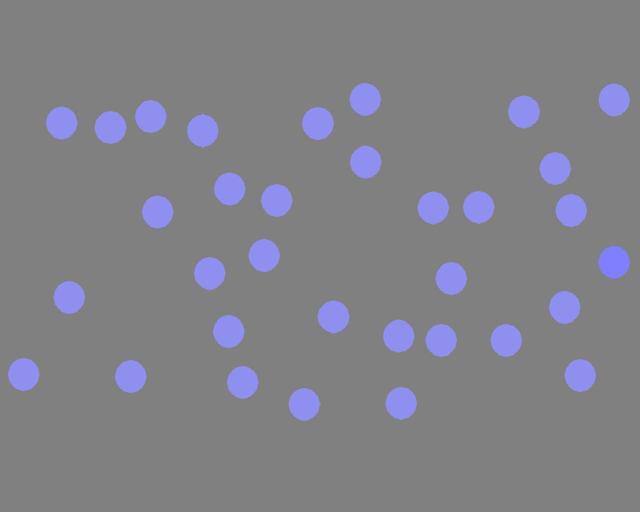}}
			\fbox{\includegraphics[width=.12\textwidth]{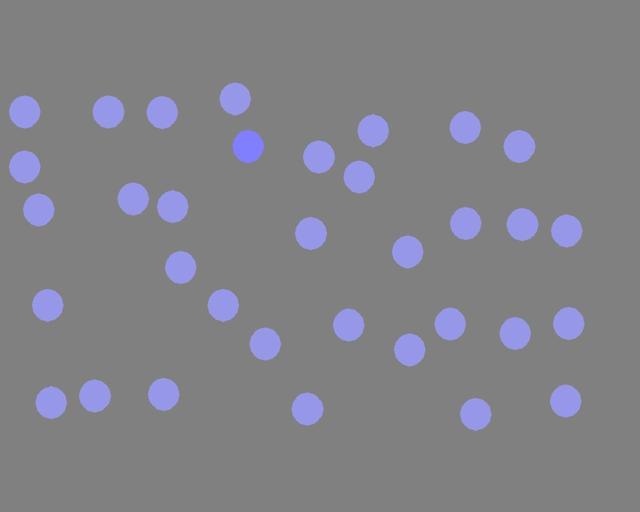}}
			\fbox{\includegraphics[width=.12\textwidth]{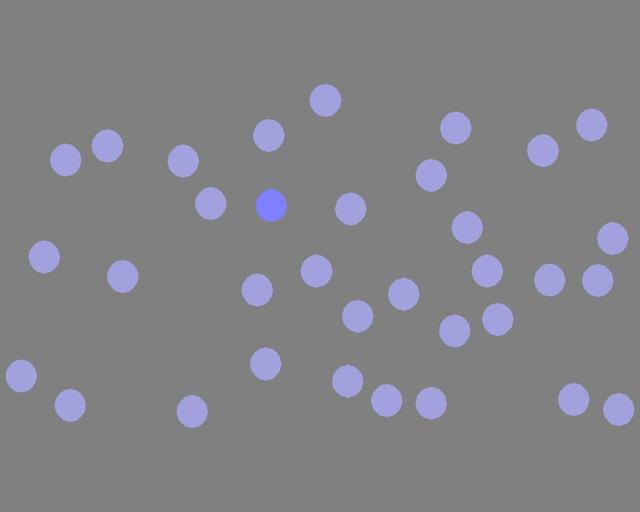}}
			\fbox{\includegraphics[width=.12\textwidth]{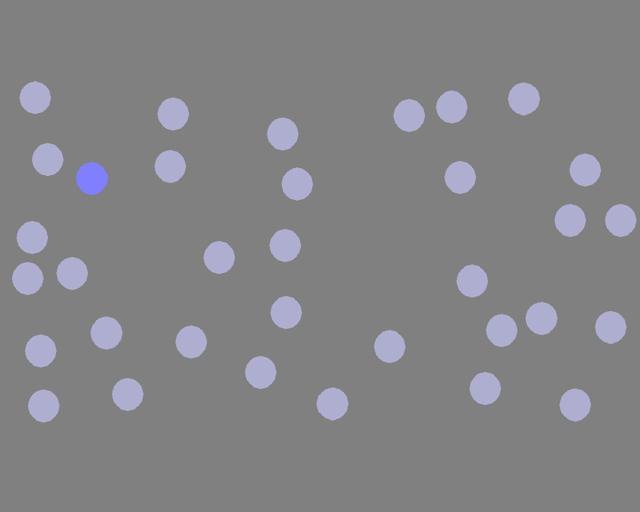}}
			\fbox{\includegraphics[width=.12\textwidth]{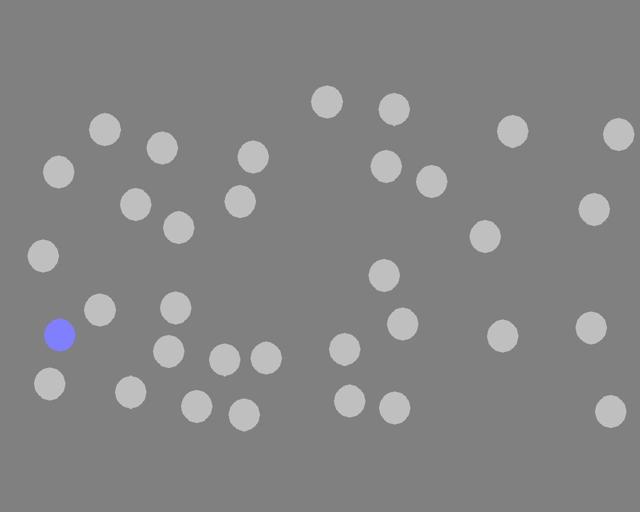}}
			\\ \hspace{2.5mm}
			\fbox{\includegraphics[width=.12\textwidth]{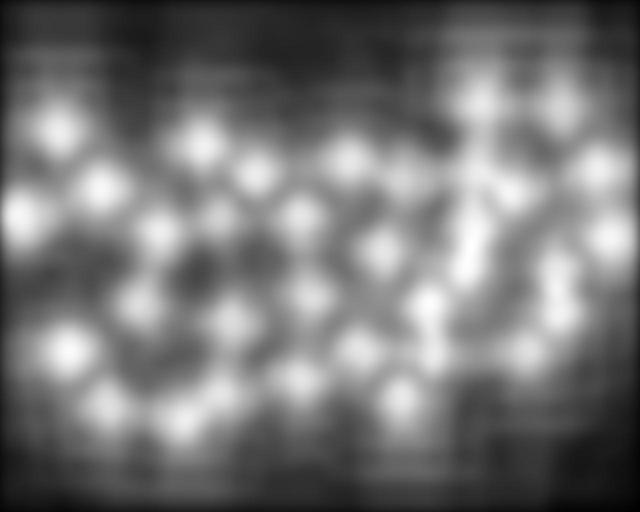}}
			\fbox{\includegraphics[width=.12\textwidth]{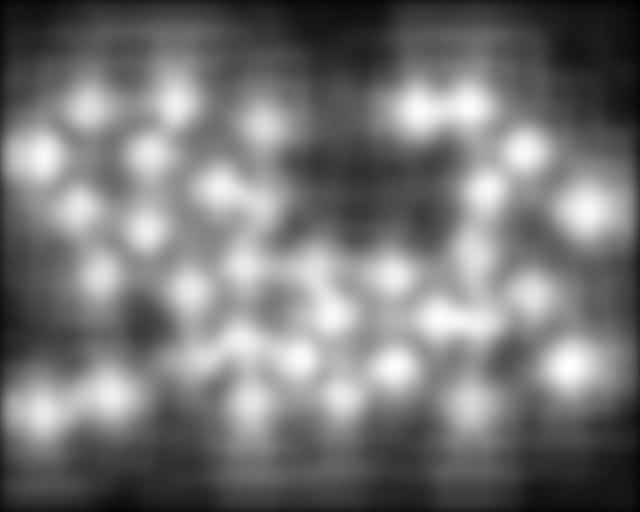}}
			\fbox{\includegraphics[width=.12\textwidth]{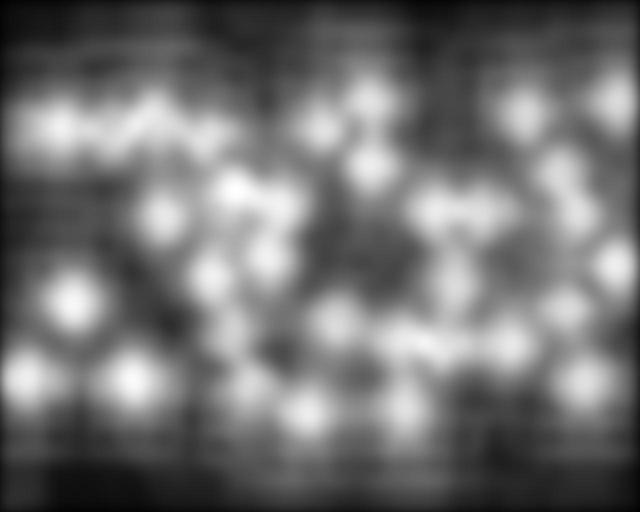}}
			\fbox{\includegraphics[width=.12\textwidth]{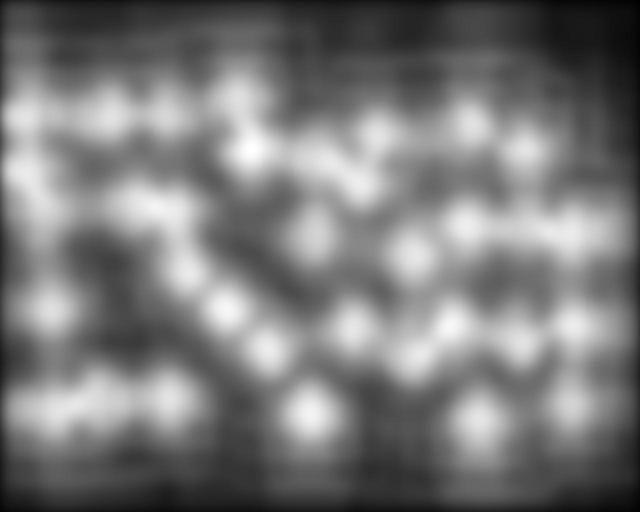}}
			\fbox{\includegraphics[width=.12\textwidth]{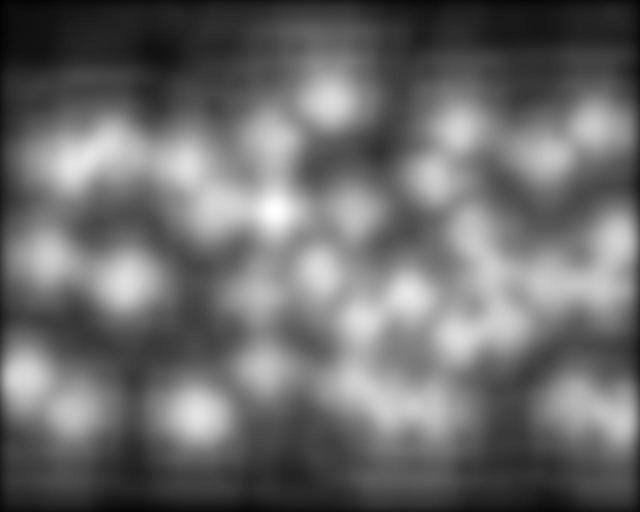}}
			\fbox{\includegraphics[width=.12\textwidth]{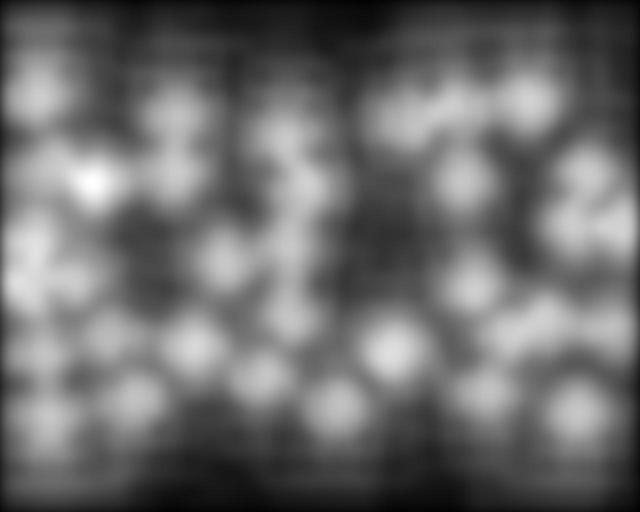}}
			\fbox{\includegraphics[width=.12\textwidth]{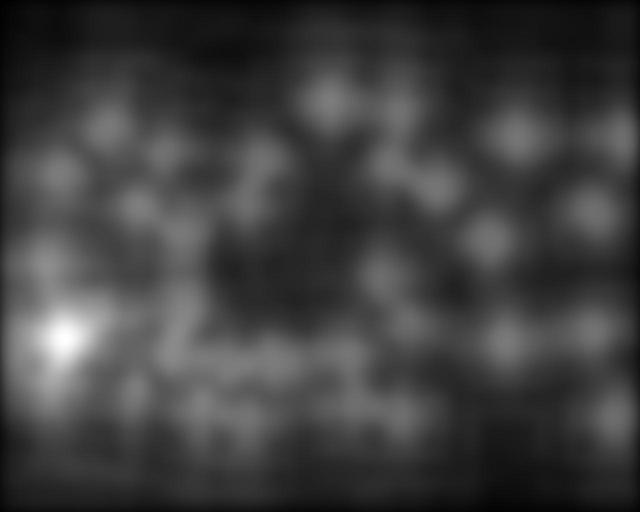}}
			\\   D
			\fbox{\includegraphics[width=.12\textwidth]{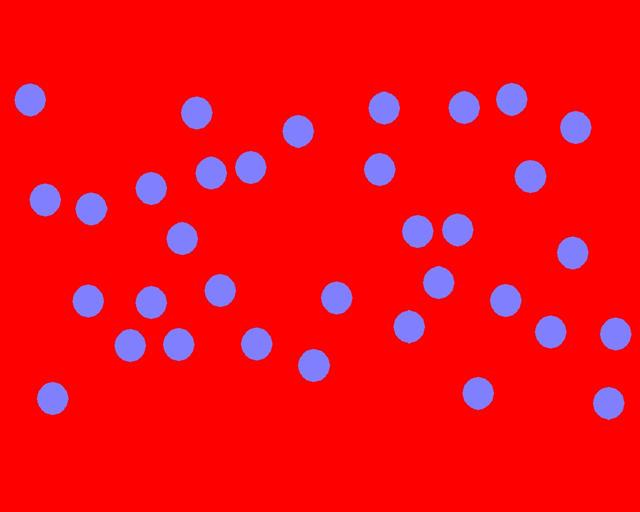}}
			\fbox{\includegraphics[width=.12\textwidth]{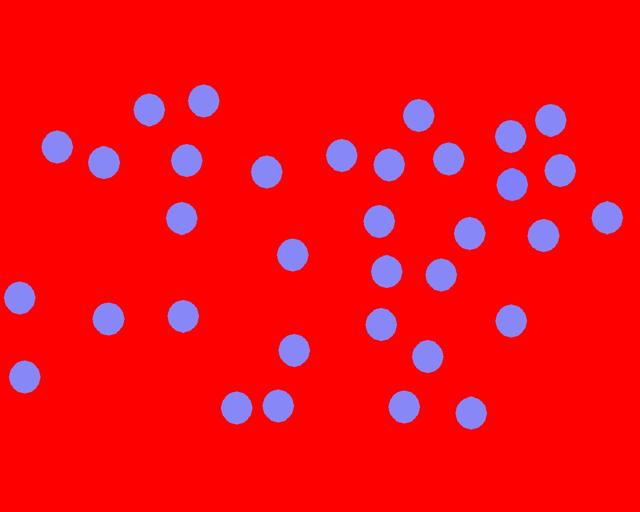}}
			\fbox{\includegraphics[width=.12\textwidth]{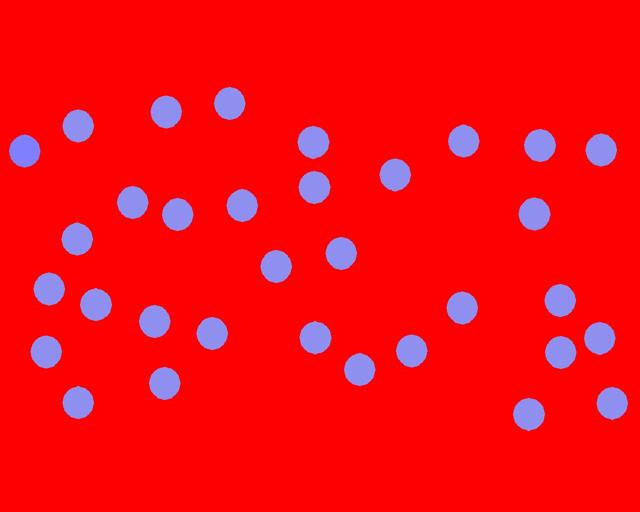}}
			\fbox{\includegraphics[width=.12\textwidth]{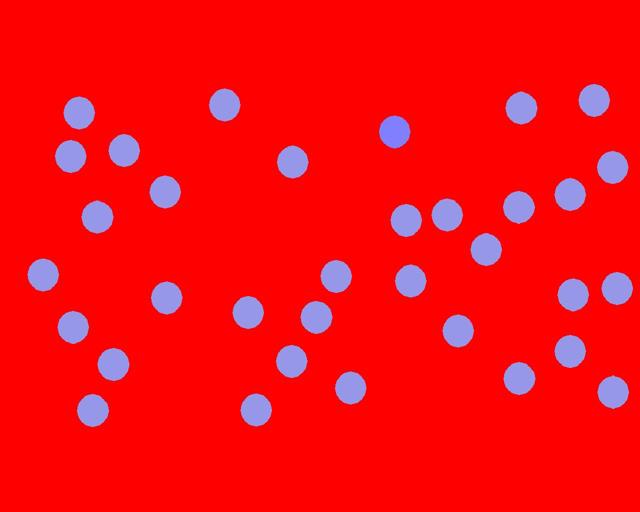}}
			\fbox{\includegraphics[width=.12\textwidth]{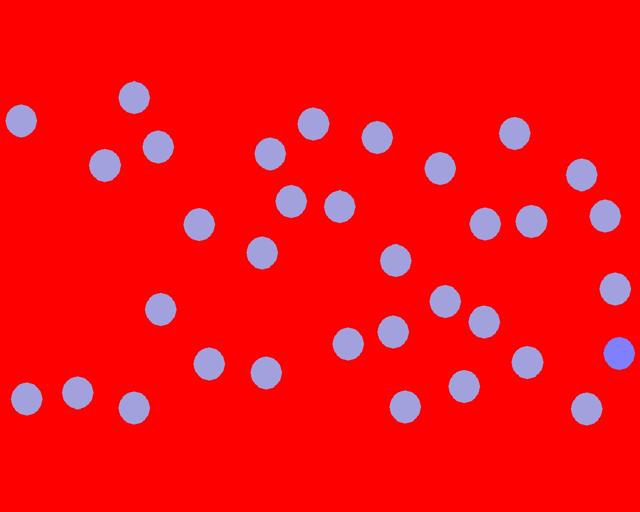}}
			\fbox{\includegraphics[width=.12\textwidth]{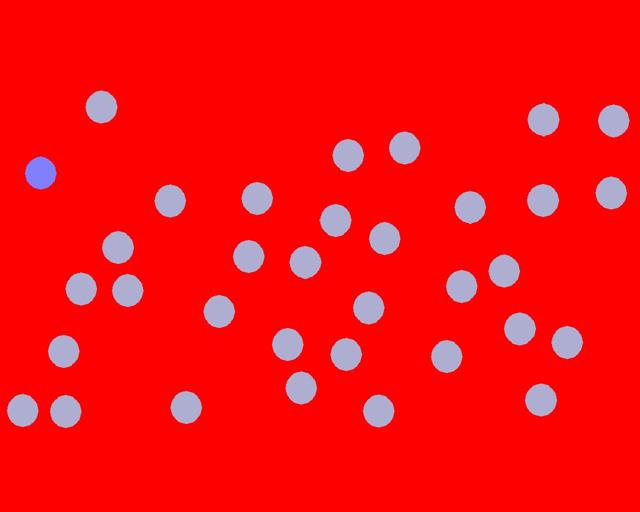}}
			\fbox{\includegraphics[width=.12\textwidth]{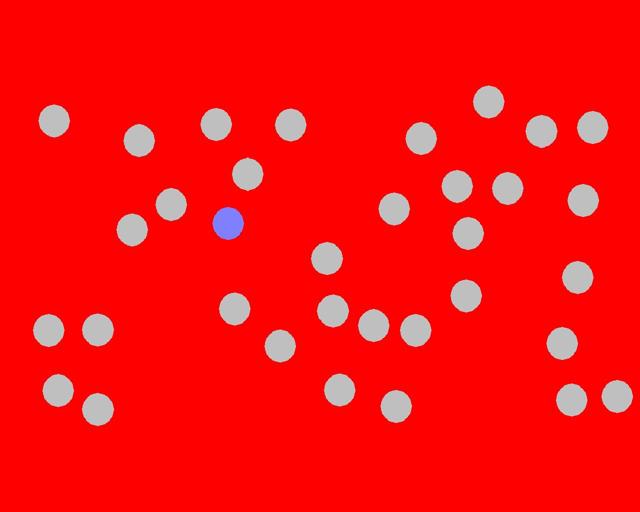}}
			\\    \hspace{2.5mm}
			\fbox{\includegraphics[width=.12\textwidth]{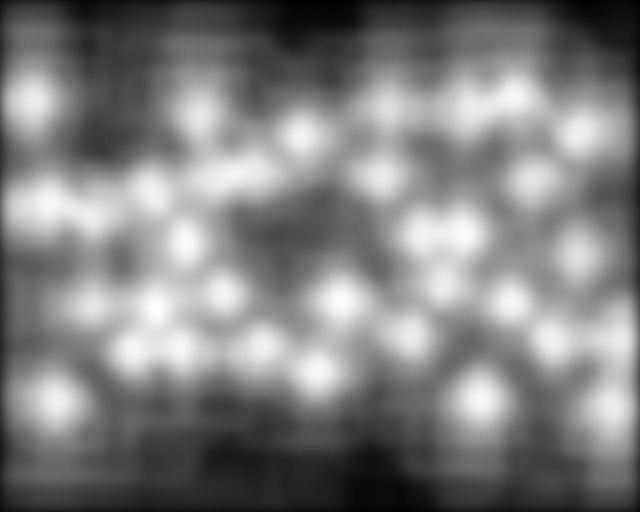}}
			\fbox{\includegraphics[width=.12\textwidth]{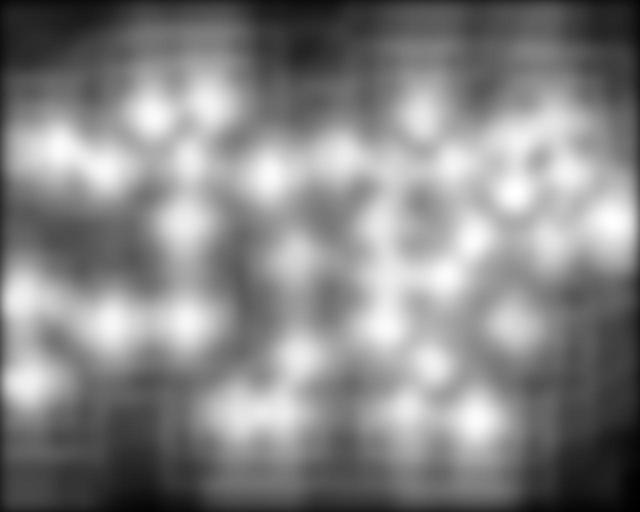}}
			\fbox{\includegraphics[width=.12\textwidth]{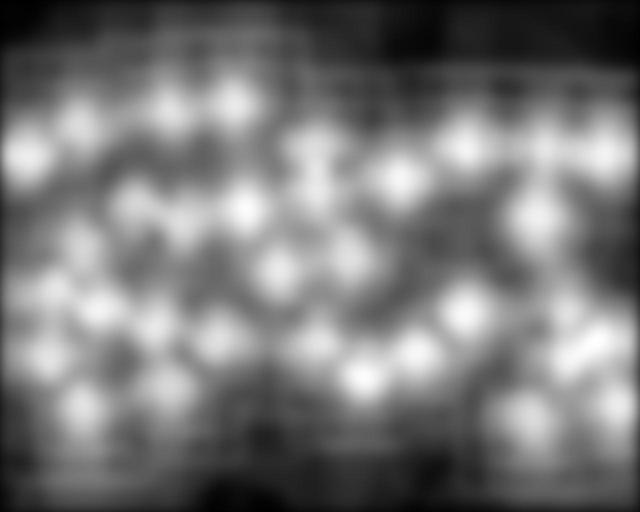}}
			\fbox{\includegraphics[width=.12\textwidth]{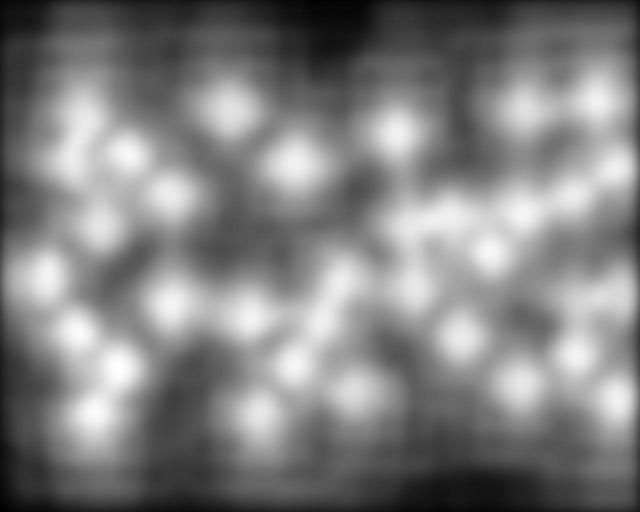}}
			\fbox{\includegraphics[width=.12\textwidth]{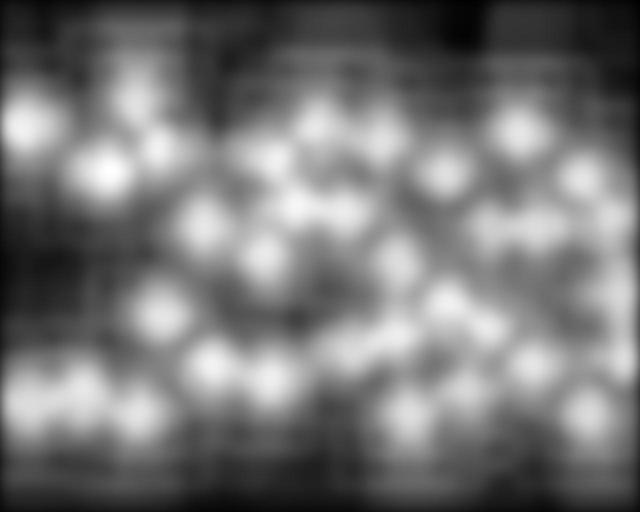}}
			\fbox{\includegraphics[width=.12\textwidth]{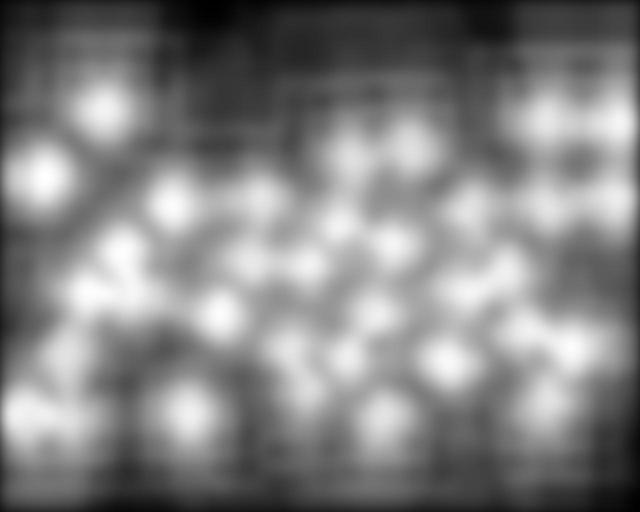}}
			\fbox{\includegraphics[width=.12\textwidth]{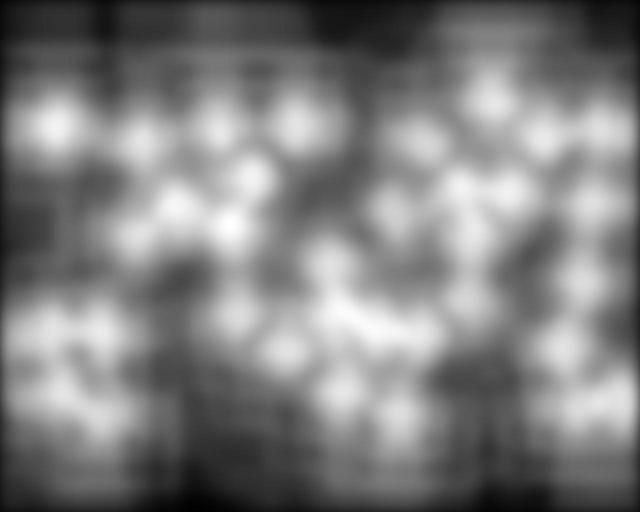}}
			\\ 
			\hspace{.04\textwidth}0\hspace{.08\textwidth}0.121\hspace{.05\textwidth}0.246\hspace{.05\textwidth}0.368\hspace{.05\textwidth}0.528\hspace{.05\textwidth}0.728\hspace{.06\textwidth}1
		\end{subfigure}
		\caption{Chromatic stimuli upon saturation contrast ($\Delta S_{D,T}$) between a red target ($H_T=0^{\circ}$) and a \textbf{(A)} grey background or a \textbf{(B)} saturated red background. Other cases \textbf{(C,D)} present a blue target ($H_T=240^{\circ}$) with same background properties to \textbf{(A)} and \textbf{(B)} respectively. Rows below \textbf{A-D} correspond to NSWAM's predicted saliency maps.}
		\label{fig:vs6}
	\end{figure}

	Color changes spatial and temporal behavior of eye movements, influencing conspicuity of specific objects on a scene \cite{DZmura1991}\cite{Bauer1996}. Similarly to previous section, here we vary the chromaticity of the background, which can alter search efficiency \cite{Nagy1999}\cite{Danilova2014}. In this section, we used stimuli similar to Rosenholtz's experimentation \cite{Rosenholtz2004}, with red and blue singletons for achromatic or saturated backgrounds \hyperref[fig:vs6]{Fig. \ref*{fig:vs6}}. Here, chromatic contrast is defined as the HSL saturation differences ($\Delta S_{D,T}$) between a salient target and the rest of distractors.

\begin{figure}[h!]
		\centering
		\begin{subfigure}{.9\linewidth}
			\includegraphics[width=\textwidth]{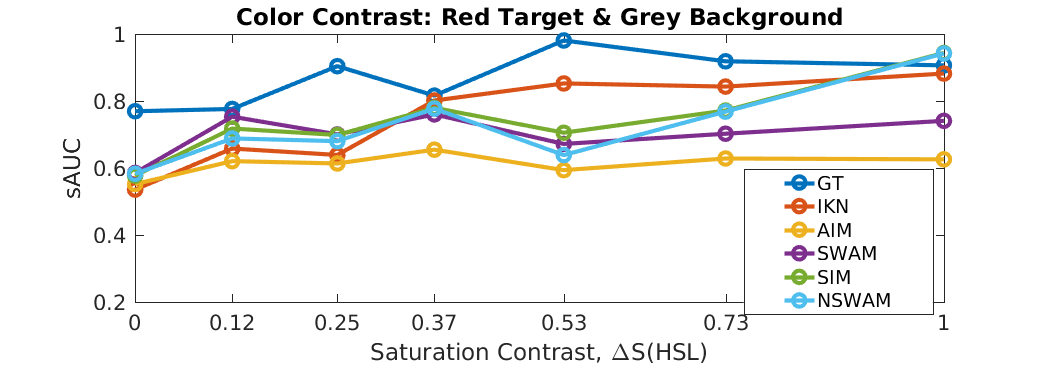}
			\caption*{\centering \textbf{A}}
		\end{subfigure}
		\begin{subfigure}{.9\linewidth}
			\includegraphics[width=\textwidth]{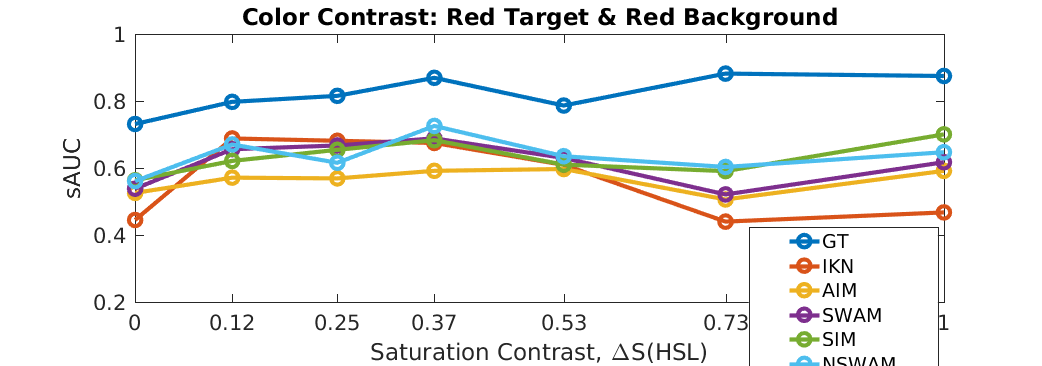}
			\caption*{\centering \textbf{B}}
		\end{subfigure}
		\begin{subfigure}{.9\linewidth}
			\includegraphics[width=\textwidth]{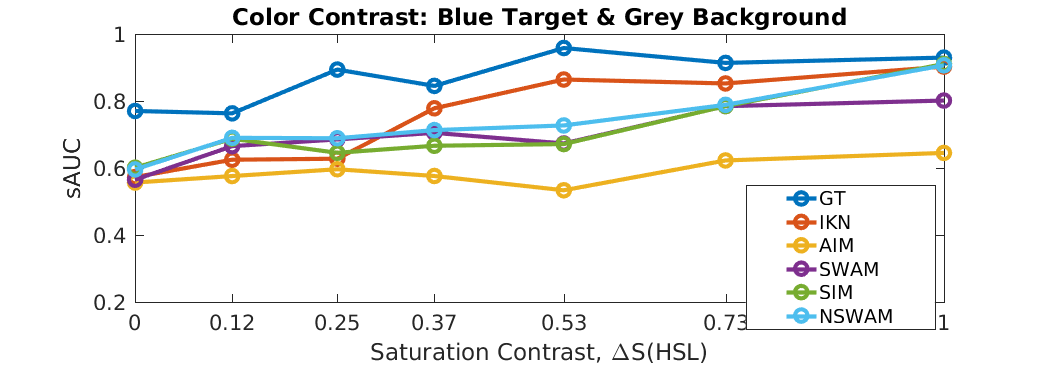}
			\caption*{\centering \textbf{C}}
		\end{subfigure}
		\begin{subfigure}{.9\linewidth}
			\includegraphics[width=\textwidth]{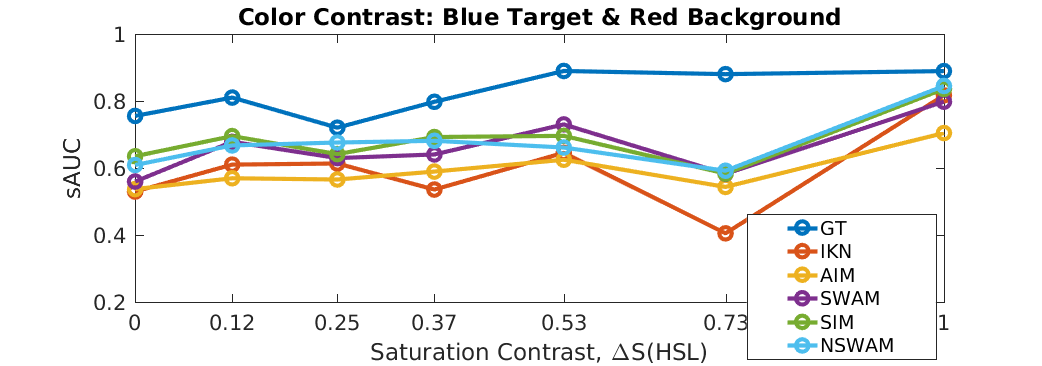}
			\caption*{\centering \textbf{D}}
		\end{subfigure}
		\caption{Results of the sAUC metric upon saturation contrast ($\Delta S_{D,T}$) on a red singleton with \textbf{(A)} achromatic or \textbf{(B)} saturated red background, or either a blue singleton with \textbf{(C)} achromatic or \textbf{(D)} saturated red background. We can see that our models SWAM, SIM and NSWAM are usually among the best methods.}
		\label{fig:res_b9}
	\end{figure}

	Similarly to \hyperref[fig:vs5]{Fig. \ref*{fig:vs5}}, NSWAM has similar sAUC to SIM for all background conditions (\hyperref[fig:res_b9]{Fig. \ref*{fig:res_b9},A-D}). Achromatic backgrounds contribute to salient object detection by increasing sAUC of the pop-out singleton. That effect is present for visual search results and our saliency prediction. Results comparing target search fixation maps and sAUC show distinct performance upon saturation contrast depending on background conditions. Cases where stimulus background is achromatic, distinct from the feature singleton, have higher correlation than with saturated background. For the cases of grey (achromatic) background, there is a correlation between sAUC results for our model and $\Delta S_{D,T}$ with a red ($\rho=.864, p=1.2 \times 10^{-2}$) and blue ($\rho=.944, p=1.4 \times 10^{-3}$) target singleton. However, when background color is saturated red, while targets are either red ($\rho=.106, p=.82$) or blue ($\rho=.483, p=.27$), then saturation contrast do not correlate with sAUC.
	
	\vfill
	
	\subsubsection{Size contrast} 
	
\begin{figure} [H]
		\centering
		\setlength{\fboxsep}{0pt}%
		\setlength{\fboxrule}{0.5pt}%
		\begin{subfigure}{\linewidth} \centering
			\fbox{\includegraphics[width=.12\textwidth]{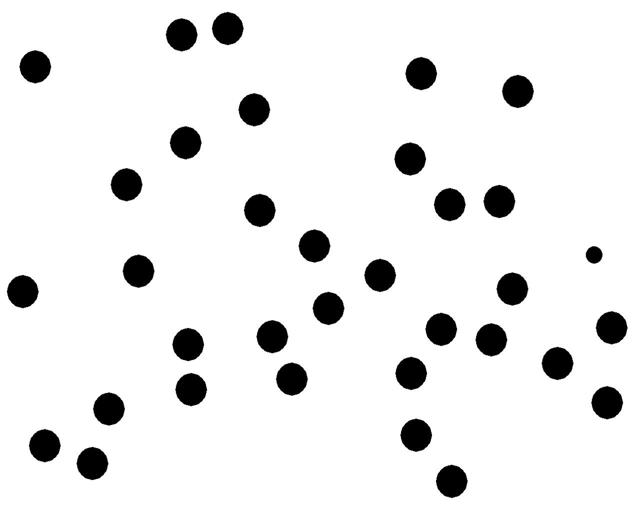}}
			\fbox{\includegraphics[width=.12\textwidth]{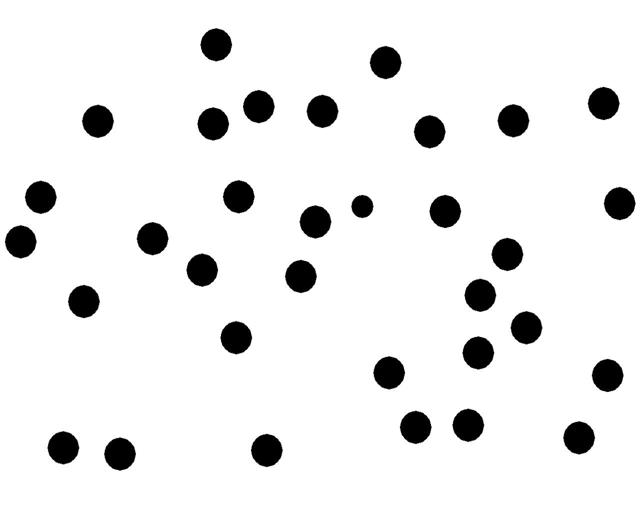}}
			\fbox{\includegraphics[width=.12\textwidth]{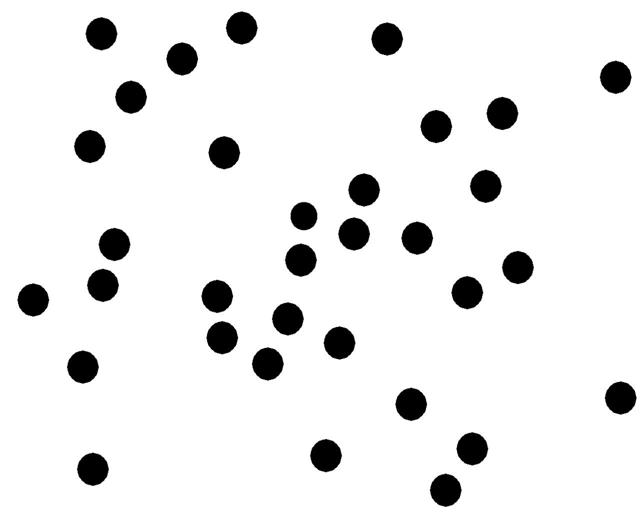}}
			\fbox{\includegraphics[width=.12\textwidth]{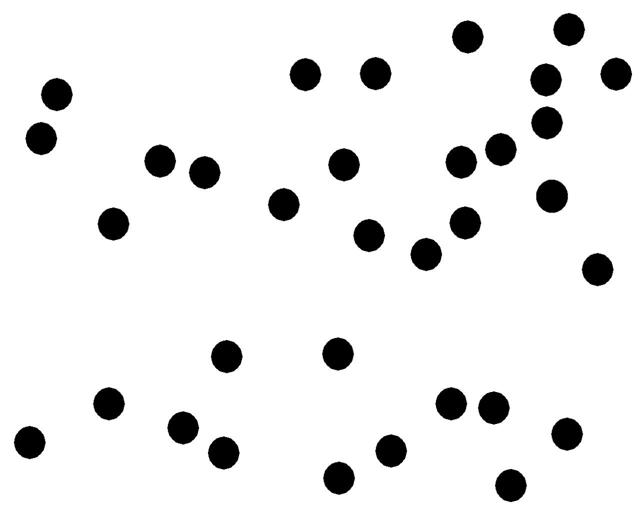}}
			\fbox{\includegraphics[width=.12\textwidth]{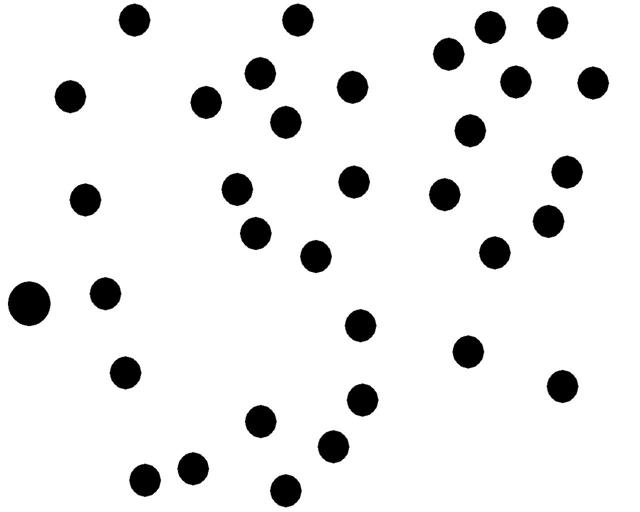}}
			\fbox{\includegraphics[width=.12\textwidth]{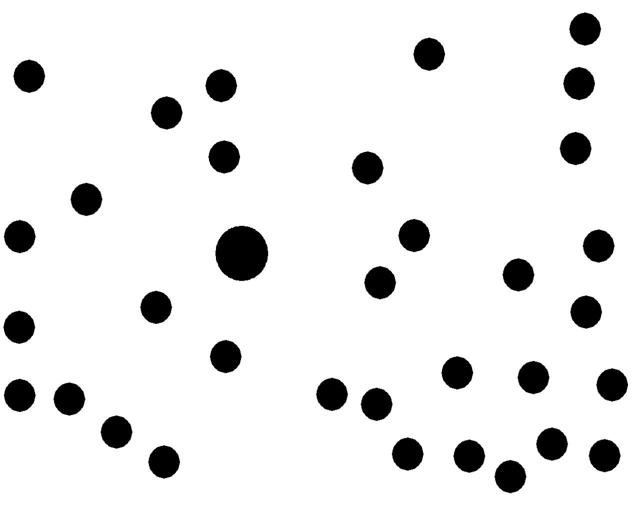}}
			\fbox{\includegraphics[width=.12\textwidth]{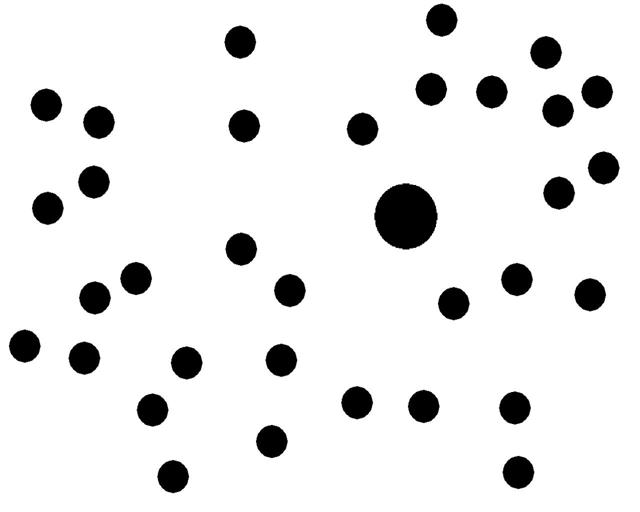}}
			\\   
			\fbox{\includegraphics[width=.12\textwidth]{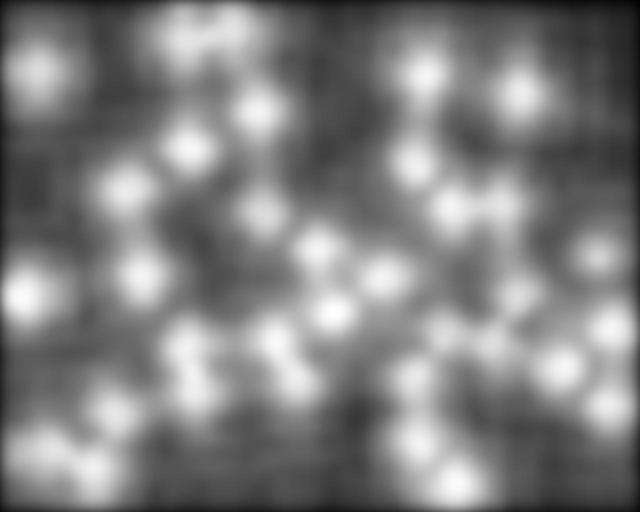}}
			\fbox{\includegraphics[width=.12\textwidth]{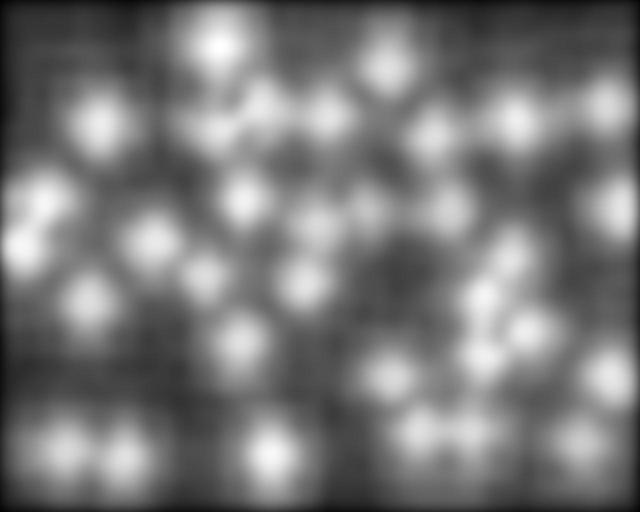}}
			\fbox{\includegraphics[width=.12\textwidth]{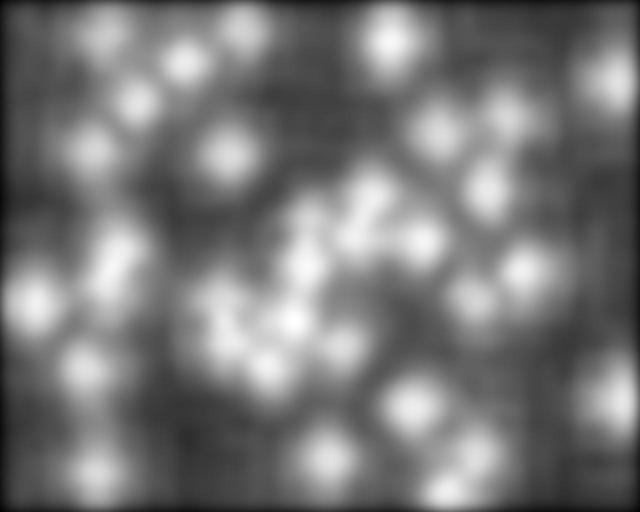}}
			\fbox{\includegraphics[width=.12\textwidth]{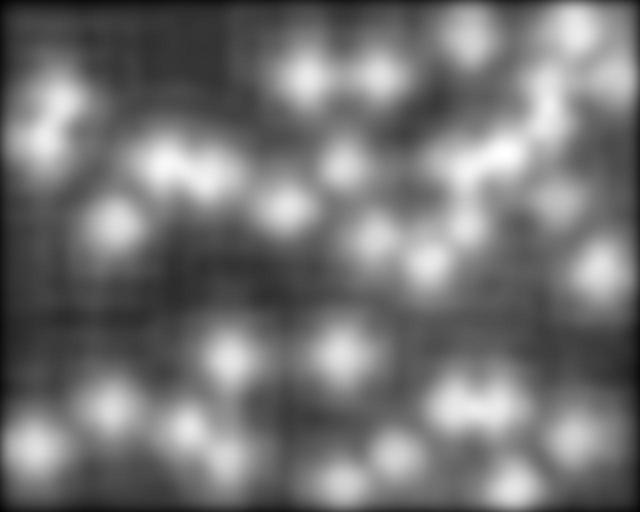}}
			\fbox{\includegraphics[width=.12\textwidth]{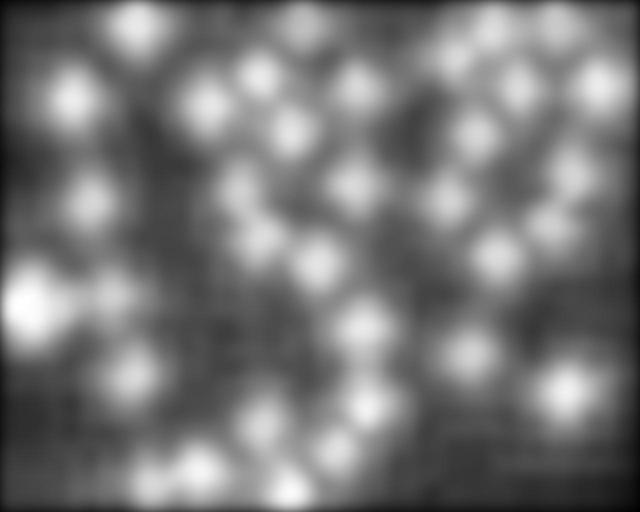}}
			\fbox{\includegraphics[width=.12\textwidth]{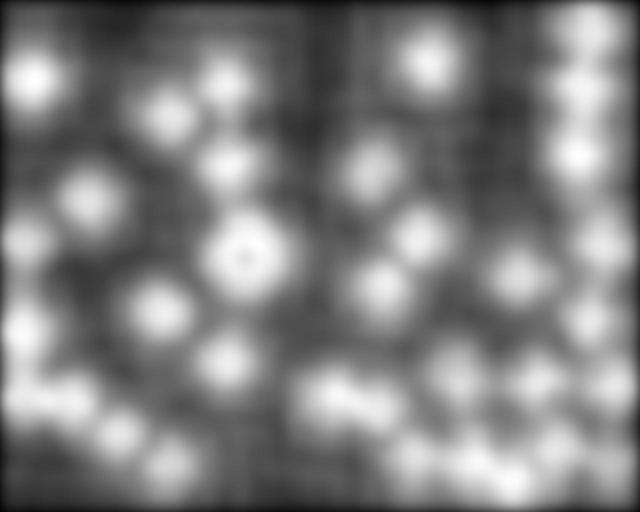}}
			\fbox{\includegraphics[width=.12\textwidth]{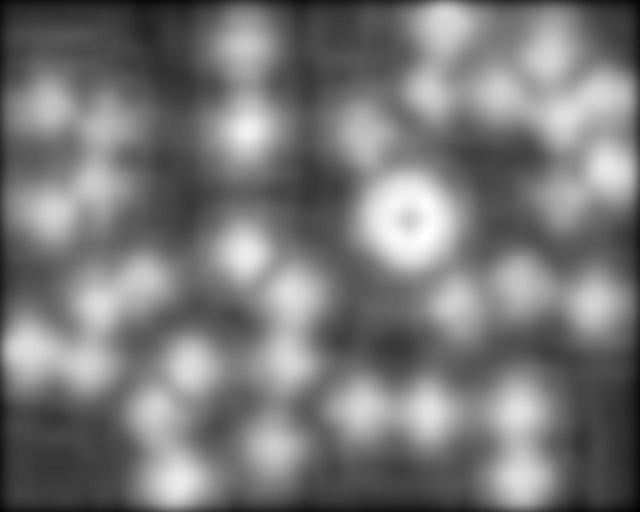}}
			\\ 
			1.25\hspace{.06\textwidth}1.67\hspace{.08\textwidth}2.08\hspace{.08\textwidth}2.5\hspace{.06\textwidth}3.34\hspace{.08\textwidth}4.17\hspace{.10\textwidth}5
		\end{subfigure}
		\caption{ Examples of circle distractors with equal diameter ($\diameter_D=2.5\deg$), containing a salient one with dissimilar size ($\diameter_T=1.25..5\deg$) with respect the rest. In lower row there are NSWAM's predicted saliency maps. }
		\label{fig:vs8}
	\end{figure}

\begin{figure}[h!]
		\centering
		\begin{subfigure}{1\linewidth}
			\includegraphics[width=\textwidth]{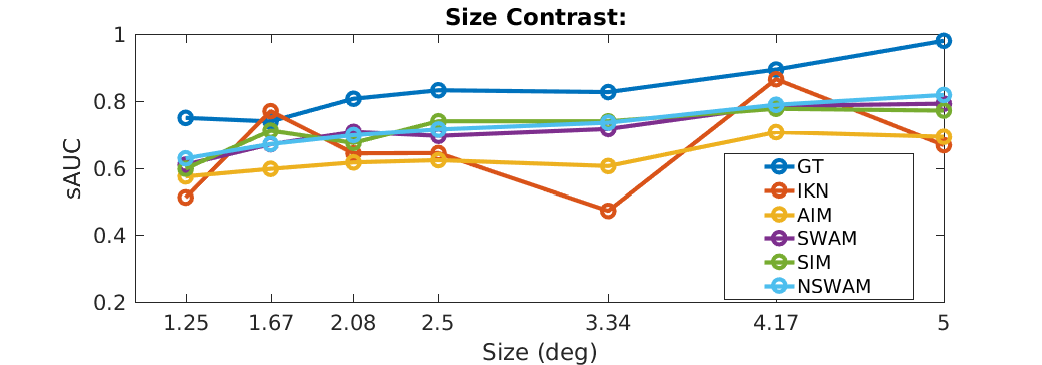}
		\end{subfigure}
		\caption{Results of the sAUC metric for Size Contrast stimuli. We can see that our models SWAM, SIM and NSWAM are usually among the best methods.}
		\label{fig:res_b11}
	\end{figure}

	Feature distinctiveness using feature singletons have been tested by varying set size, object orientation and/or color. Here, we test how object size affects its saliency, previously tested with visual search experimentation \cite{Goolkasian1997}\cite{Tavassoli2009}\cite{Proulx2010}. A set of 34 symmetric objects (with a dark circle shape) are distributed randomly around the image \hyperref[fig:vs8]{Fig. \ref*{fig:vs8}}, preserving equal diameter. One of the circles is defined with dissimilar size, either with higher or lower diameter with respect the rest (which are defined with a diameter of $2.5\deg$).  Performance for NSWAM's sAUC improves with size dissimilarity. When the diameter of the dissimilar circle is higher, sAUC is higher for that particular region. For the highest scaling factor (when the dissimilar object is bigger), NSWAM has higher sAUC compared to previous biologically-inspired models (\hyperref[fig:res_b11]{Fig. \ref*{fig:res_b11}}). In addition, there is a significant correlation between circle diameter and our model's results of sAUC ($\rho=.955,p=8.3\times 10^{-4}$).

	\subsubsection{Orientation contrast}\label{sec:contrast_orientation}
	
\begin{figure} [H]
		\centering
		\setlength{\fboxsep}{0pt}%
		\setlength{\fboxrule}{0.5pt}%
		\begin{subfigure}{\linewidth} \centering
			\fbox{\includegraphics[width=.12\textwidth]{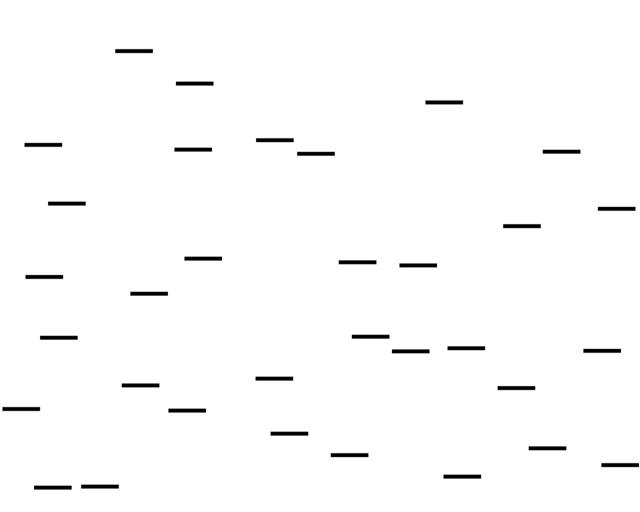}}
			\fbox{\includegraphics[width=.12\textwidth]{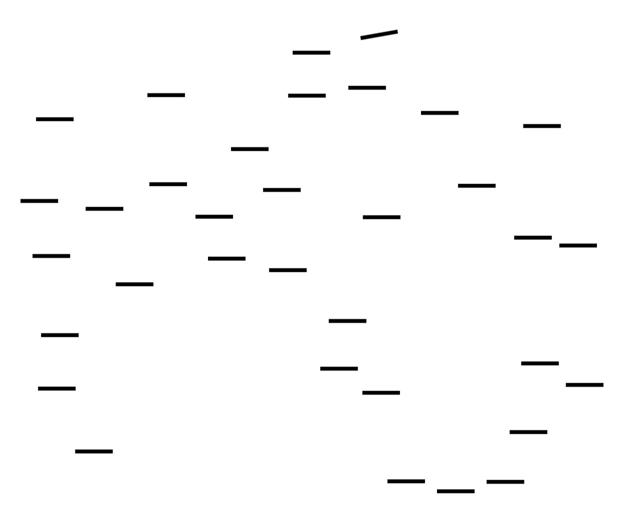}}
			\fbox{\includegraphics[width=.12\textwidth]{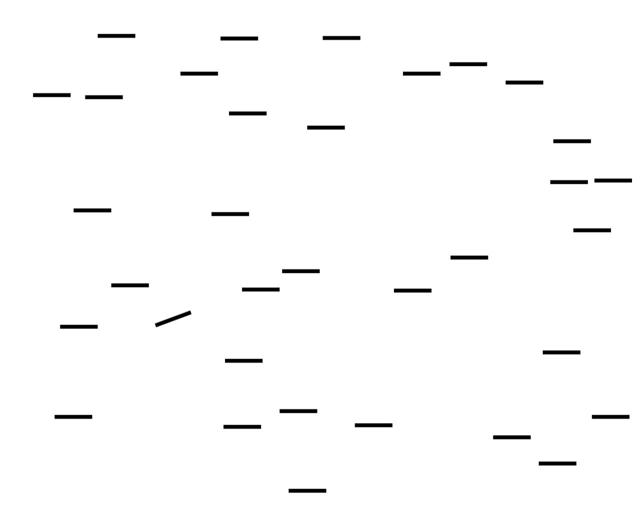}}
			\fbox{\includegraphics[width=.12\textwidth]{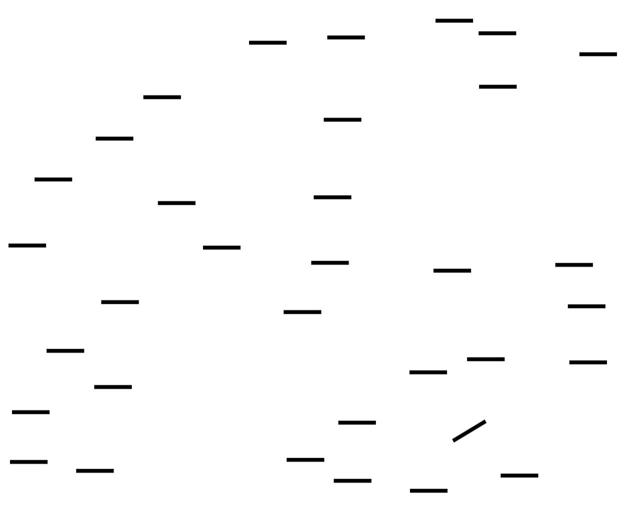}}
			\fbox{\includegraphics[width=.12\textwidth]{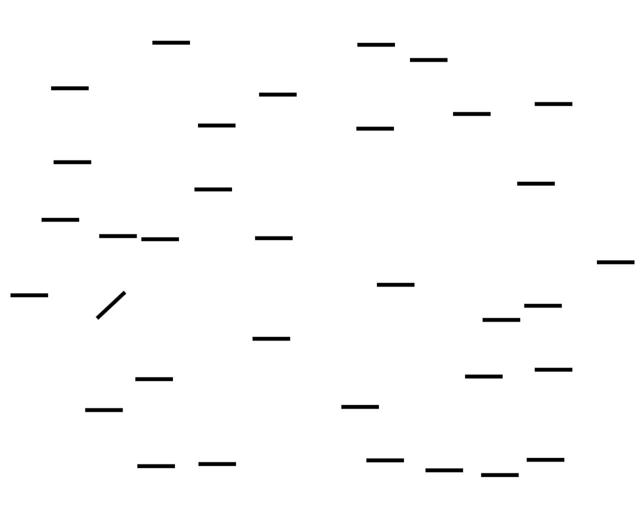}}
			\fbox{\includegraphics[width=.12\textwidth]{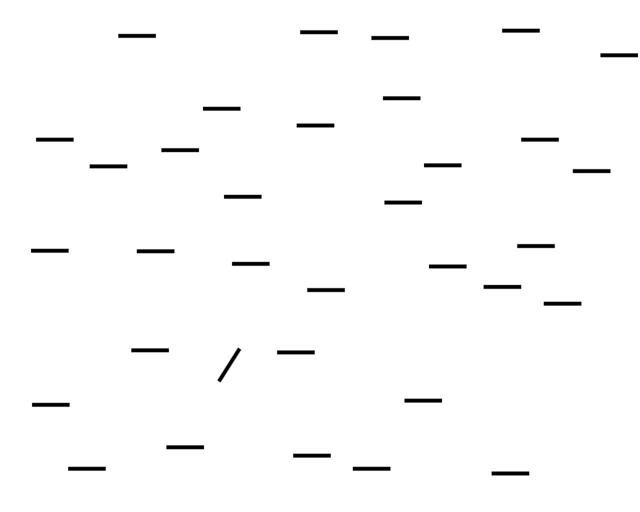}}
			\fbox{\includegraphics[width=.12\textwidth]{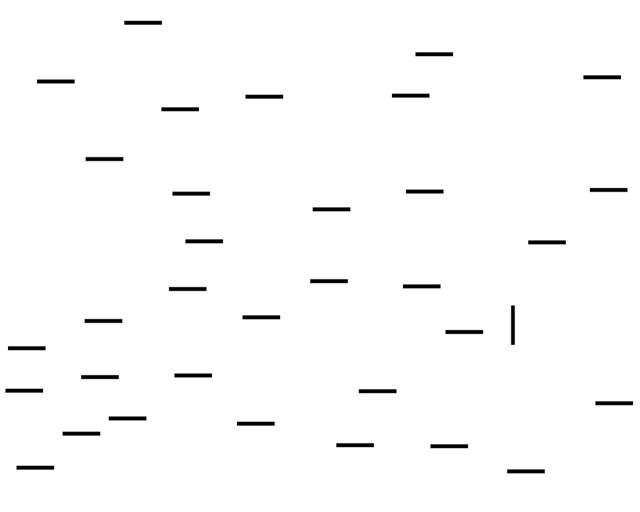}}
			\\   
			\fbox{\includegraphics[width=.12\textwidth]{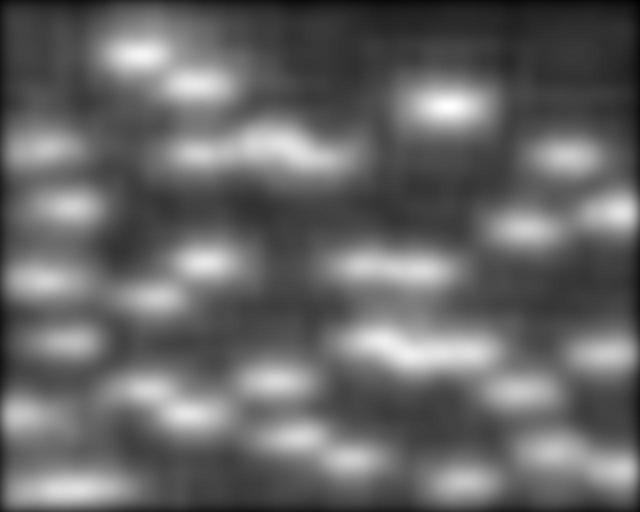}}
			\fbox{\includegraphics[width=.12\textwidth]{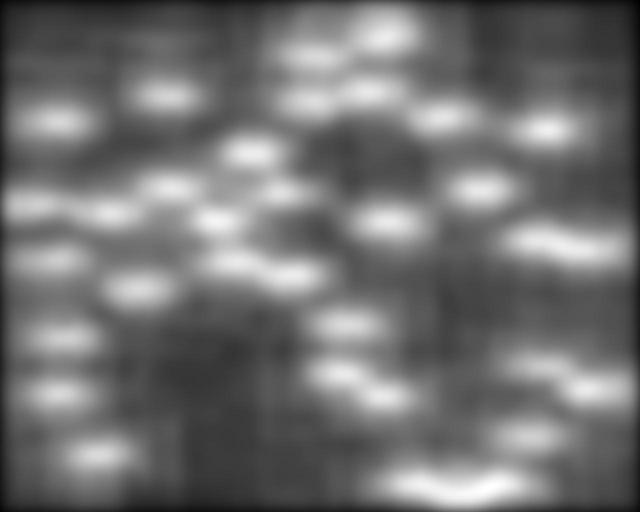}}
			\fbox{\includegraphics[width=.12\textwidth]{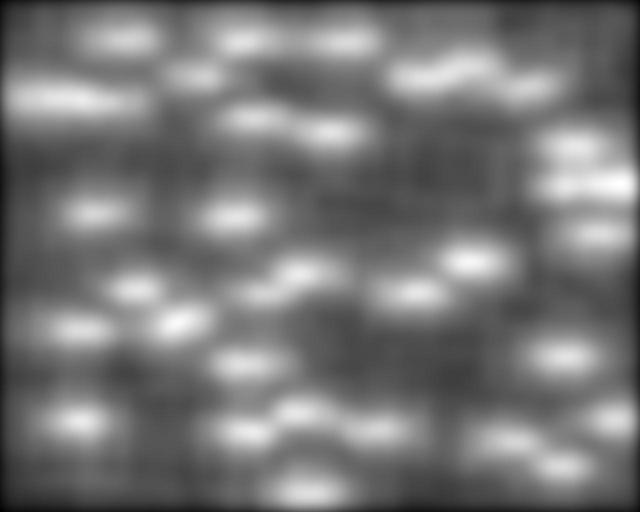}}
			\fbox{\includegraphics[width=.12\textwidth]{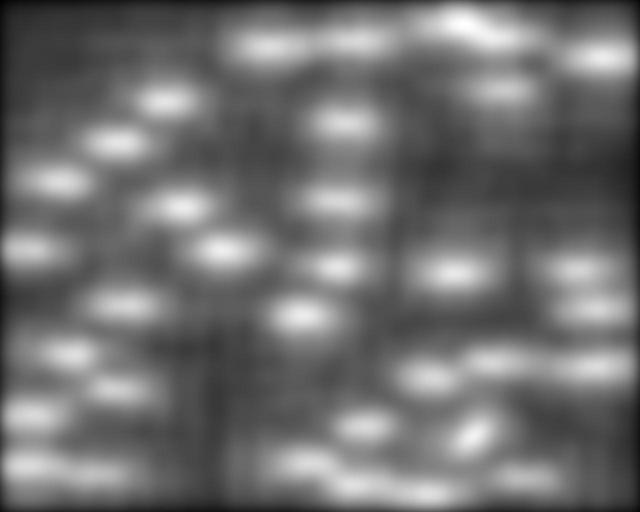}}
			\fbox{\includegraphics[width=.12\textwidth]{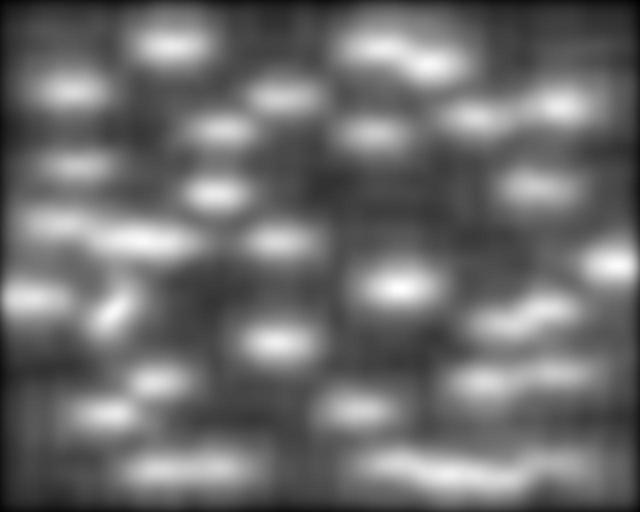}}
			\fbox{\includegraphics[width=.12\textwidth]{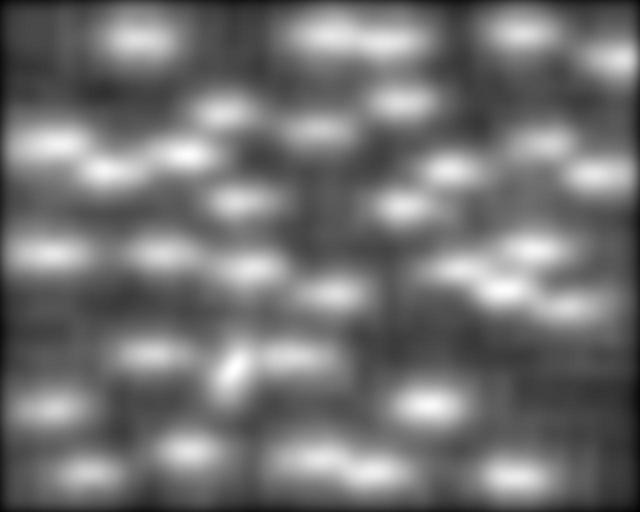}}
			\fbox{\includegraphics[width=.12\textwidth]{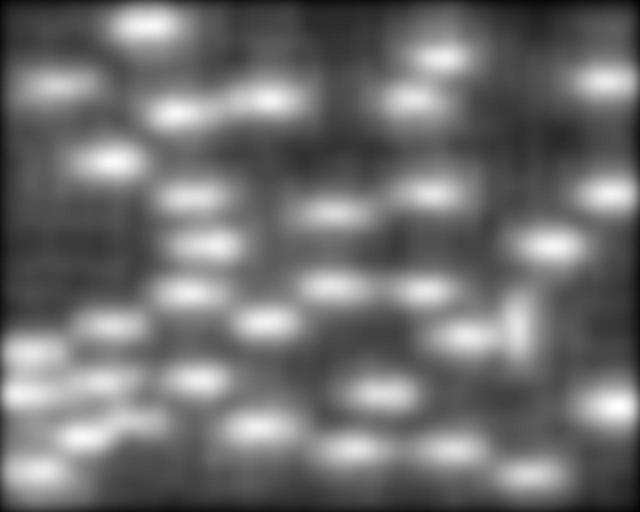}}
			\\  
			\hspace{.00\textwidth}0\hspace{.10\textwidth}10\hspace{.10\textwidth}20\hspace{.10\textwidth}30\hspace{.10\textwidth}42\hspace{.10\textwidth}56\hspace{.10\textwidth}90
		\end{subfigure}
		\caption{An oriented bar with an orientation contrast of $\Delta\Phi=0..90^{\circ}$ with respect to a set of bars oriented at $\Phi_D=0^{\circ}$. In lower row there are NSWAM's predicted saliency maps.}
		\label{fig:vs7}
	\end{figure}

\begin{figure}[h!]
		\centering
		\begin{subfigure}{1\linewidth}
			\includegraphics[width=\textwidth]{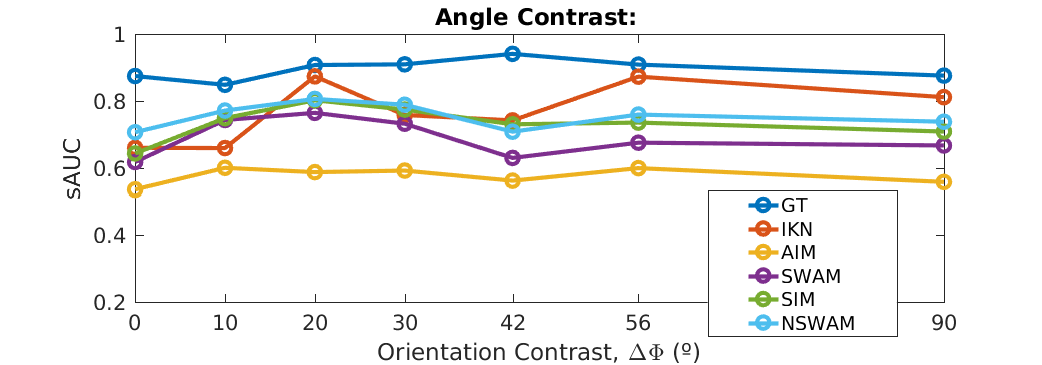}
		\end{subfigure}
		\caption{Results for sAUC metric for Orientation Contrast stimuli.}
		\label{fig:res_b12}
	\end{figure}

	Using visual stimuli defined by oriented bars, varying angle of objects is found to increase search efficiency when angle contrast is increased \cite{Duncan1989}\cite{Nothdurft1993a}\cite{Nothdurft1993b}.  A total of 34 bars were oriented horizontally and randomly displaced around the scene (\hyperref[fig:vs7]{Fig. \ref*{fig:vs7}}). The dissimilar object for this case is a bar oriented with an angle contrast with respect the rest of bars of $\Delta\Phi(1,0)$=$[0, 10, 20, 30,$ $42, 56, 90]^{\circ}$. Although results of sAUC show that NSWAM overperforms SIM's saliency maps, IKN is best for capturing orientation distinctiveness (\hyperref[fig:res_b12]{Fig. \ref*{fig:res_b12}}). In NSWAM, 3 types of orientation selective cells are modeled, corresponding to the orientation for the wavelet coefficients ($\theta=h,v,d$). A higher number of orientation selective cells would provide a higher accuracy, specially for diagonal angles (here we only provide $\theta=d$ for 45/135$^{\circ}$ combined). By modeling orientation selective cells with 2D Gabor and Log-Gabor transforms \cite{TaiSingLee1996}\cite{Fischer2007}\cite{GarciaDiaz2012} it would be possible to correctly build an hypercolumnar organization with a higher number of angle sensitivities. 


	We have to acknowledge that for this experimentation, distractors have been set with same horizontal configuration. Specific connectivity interactions \cite{Asenov2016} between orientation dissimilarities needs to be defined in order to reproduce orientation-dependent visual illusions and conspicuity under heterogeneous, nonlinear and categorical angle configurations (seen to be performed by V2 cells \cite{Anzai2007}), which are previously known to distinctively affect visual attention \cite{Nothdurft1993a}\cite{Nothdurft1993b}\cite{Gao2008}.
	
	\subsubsection{Visual Asymmetries}
	
\begin{figure}[h!]
		\centering
		\setlength{\fboxsep}{0pt}%
		\setlength{\fboxrule}{0.5pt}%
		\begin{subfigure}{\linewidth} \centering A
			\fbox{\includegraphics[width=.12\textwidth]{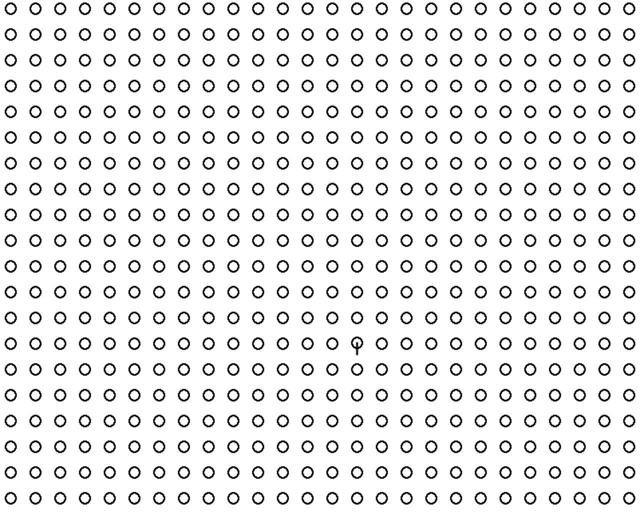}}
			\fbox{\includegraphics[width=.12\textwidth]{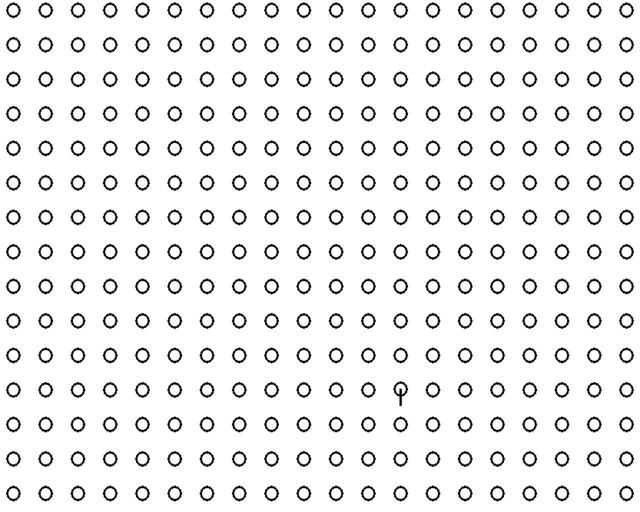}}
			\fbox{\includegraphics[width=.12\textwidth]{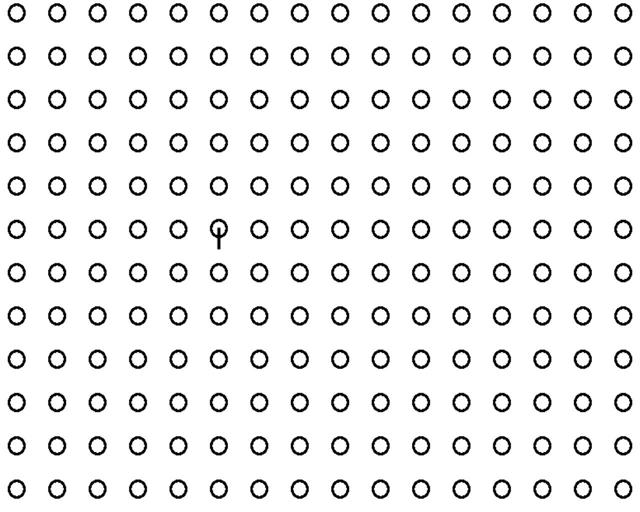}}
			\fbox{\includegraphics[width=.12\textwidth]{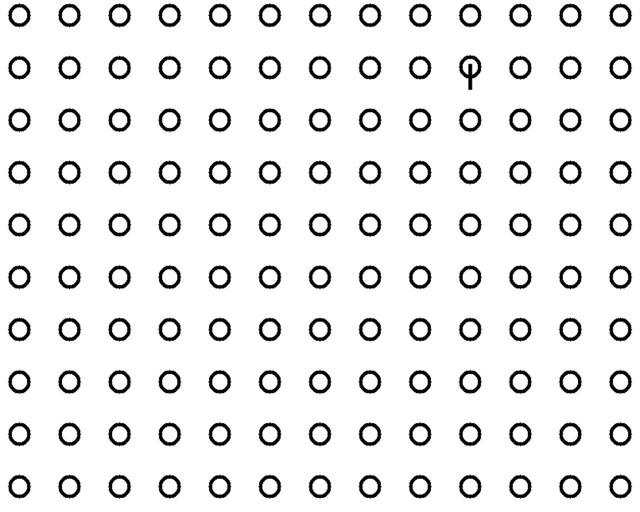}}
			\fbox{\includegraphics[width=.12\textwidth]{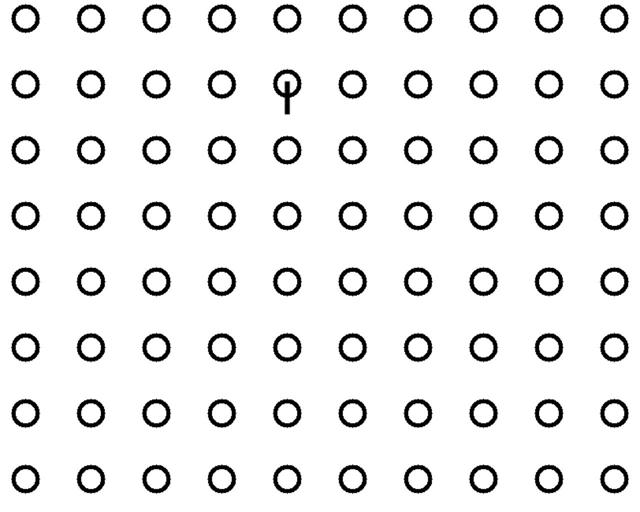}}
			\fbox{\includegraphics[width=.12\textwidth]{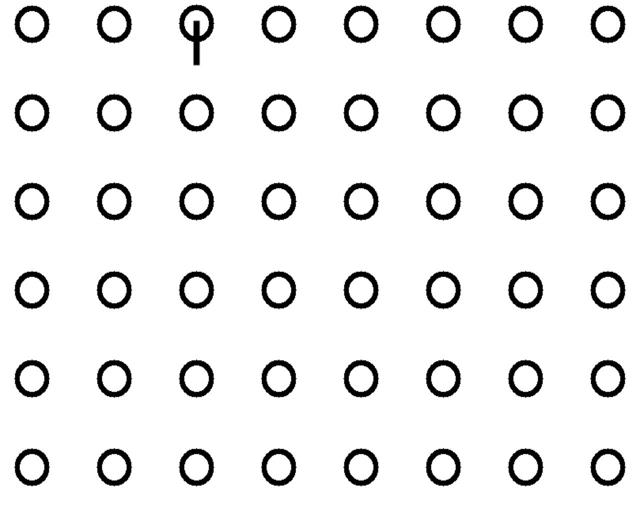}}
			\fbox{\includegraphics[width=.12\textwidth]{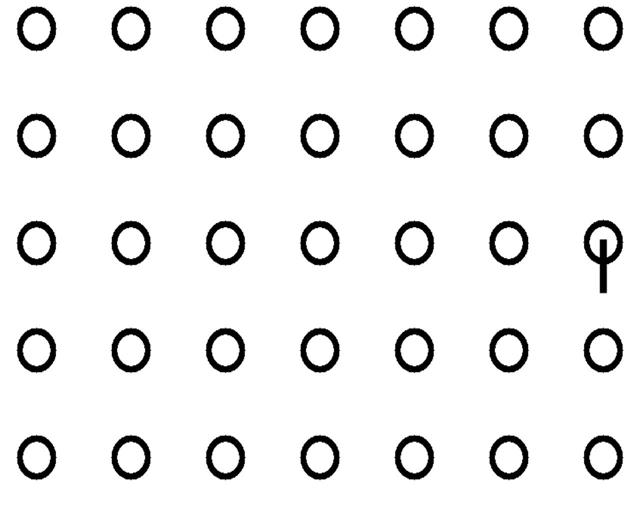}}
			\\   \hspace{2.5mm}
			\fbox{\includegraphics[width=.12\textwidth]{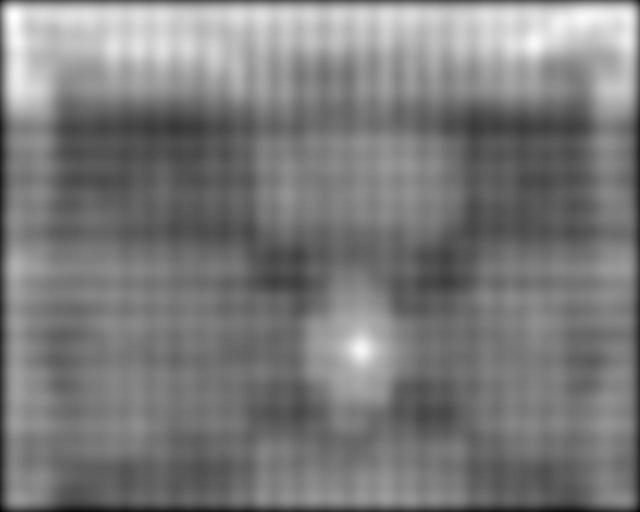}}
			\fbox{\includegraphics[width=.12\textwidth]{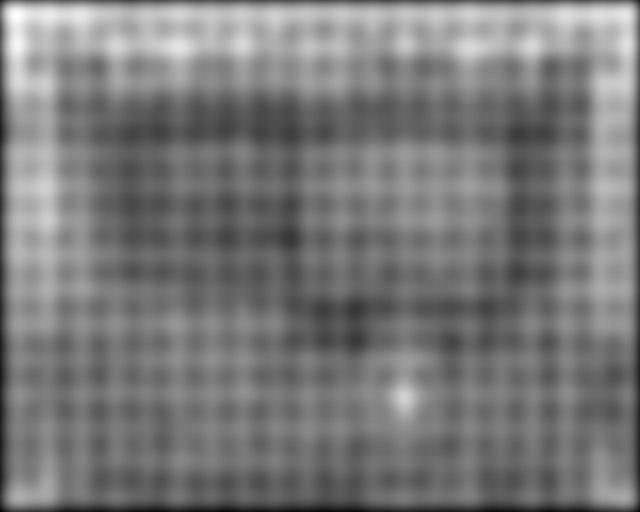}}
			\fbox{\includegraphics[width=.12\textwidth]{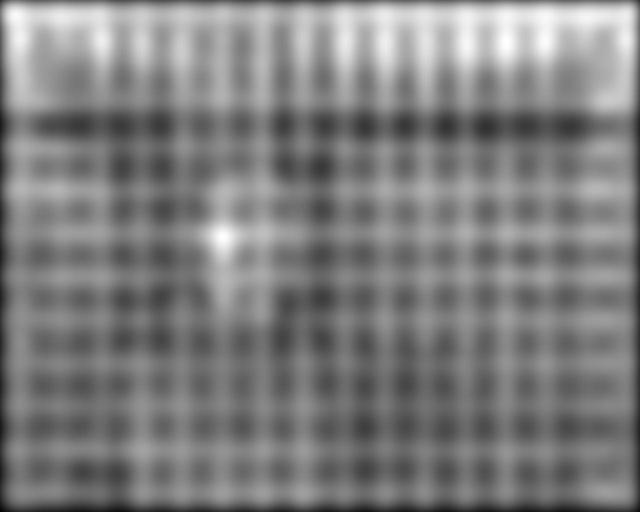}}
			\fbox{\includegraphics[width=.12\textwidth]{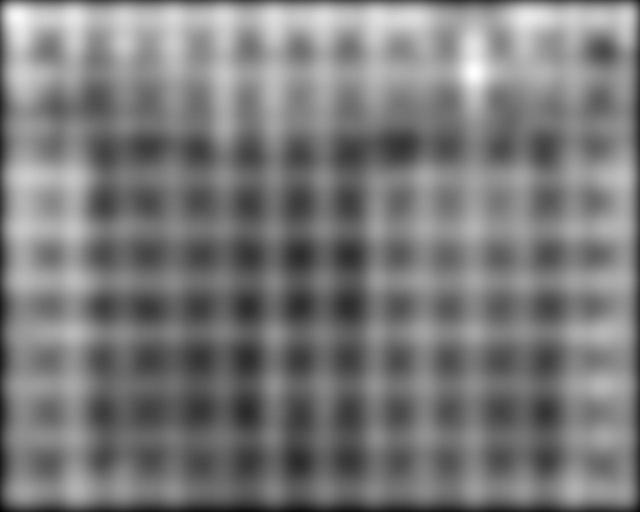}}
			\fbox{\includegraphics[width=.12\textwidth]{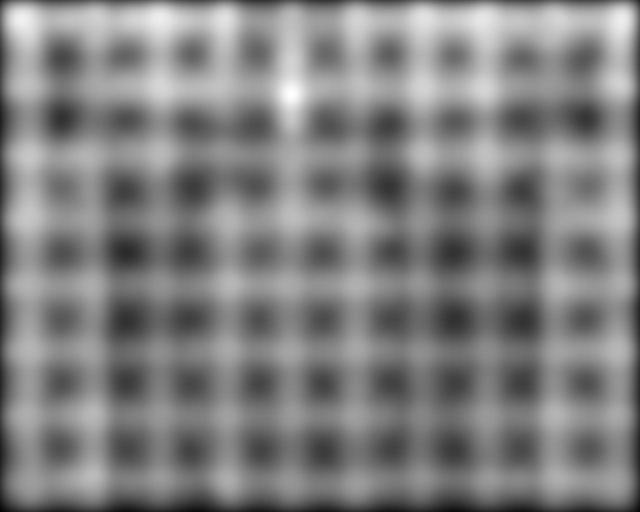}}
			\fbox{\includegraphics[width=.12\textwidth]{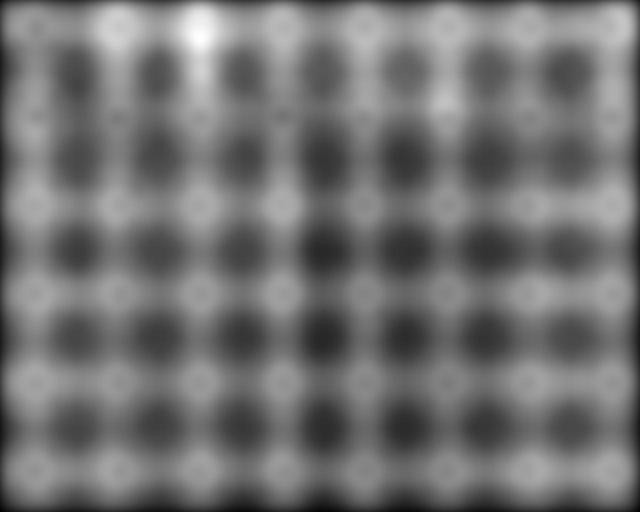}}
			\fbox{\includegraphics[width=.12\textwidth]{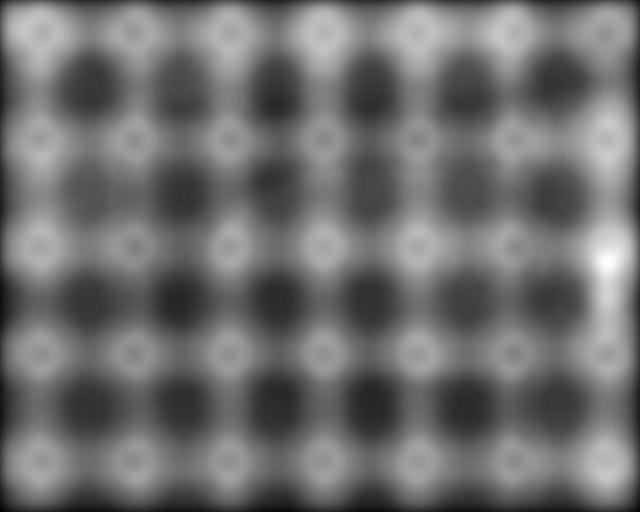}}
			\\   B
			\fbox{\includegraphics[width=.12\textwidth]{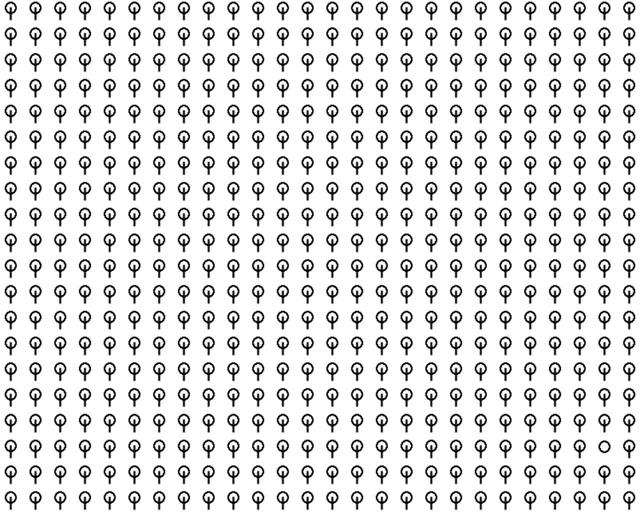}}
			\fbox{\includegraphics[width=.12\textwidth]{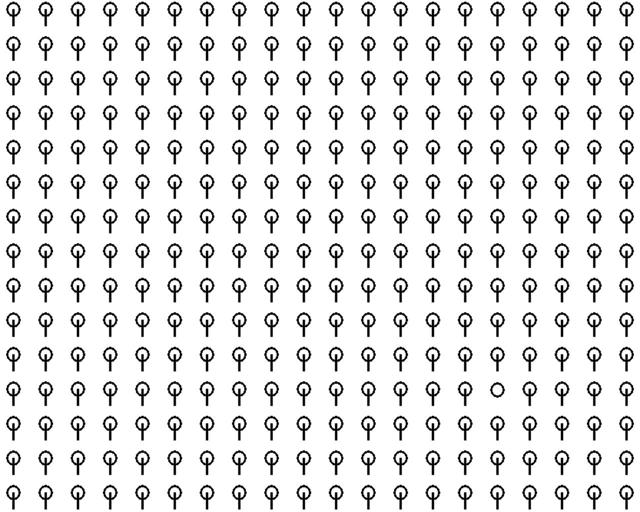}}
			\fbox{\includegraphics[width=.12\textwidth]{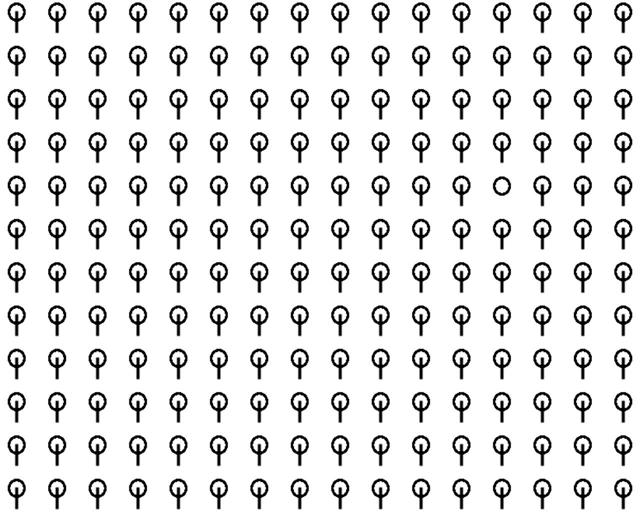}}
			\fbox{\includegraphics[width=.12\textwidth]{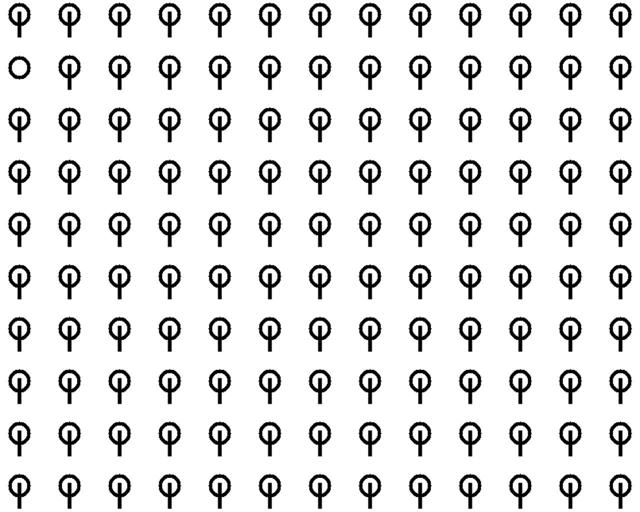}}
			\fbox{\includegraphics[width=.12\textwidth]{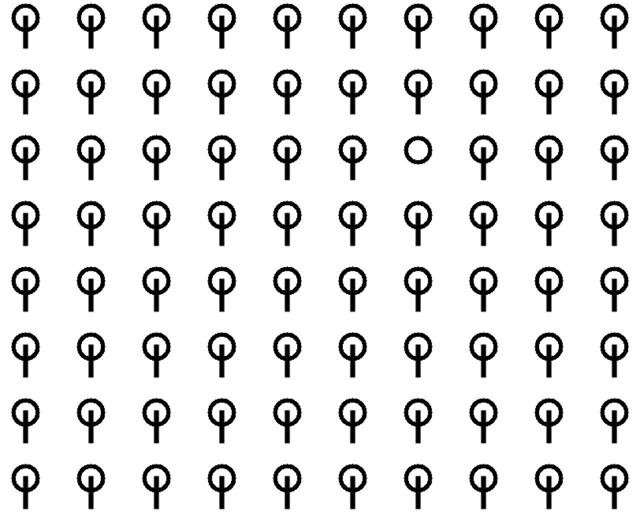}}
			\fbox{\includegraphics[width=.12\textwidth]{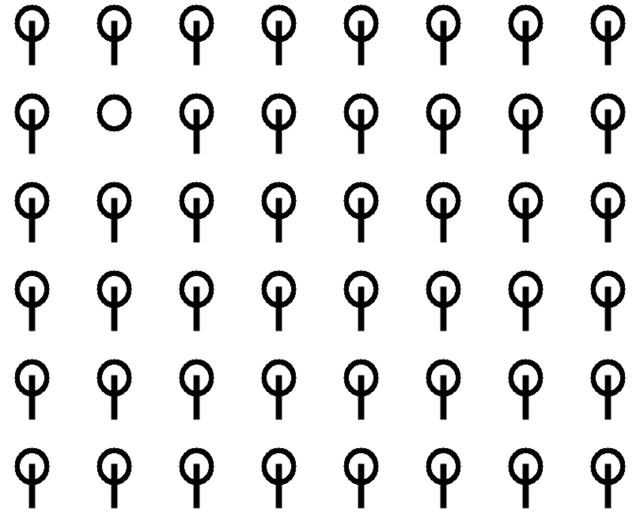}}
			\fbox{\includegraphics[width=.12\textwidth]{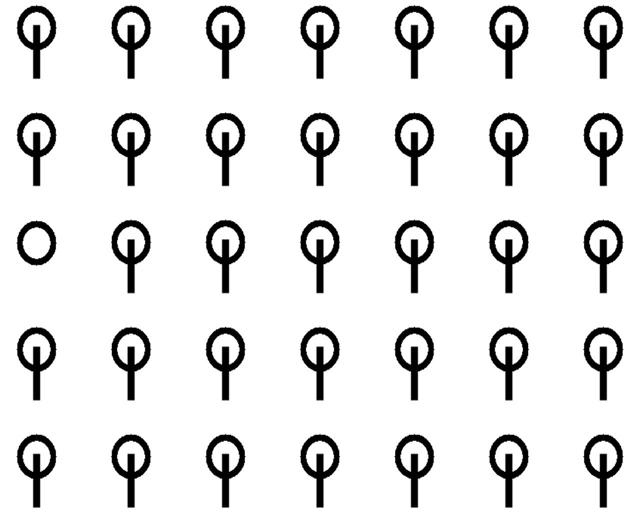}}
			\\ \hspace{2.5mm}
			\fbox{\includegraphics[width=.12\textwidth]{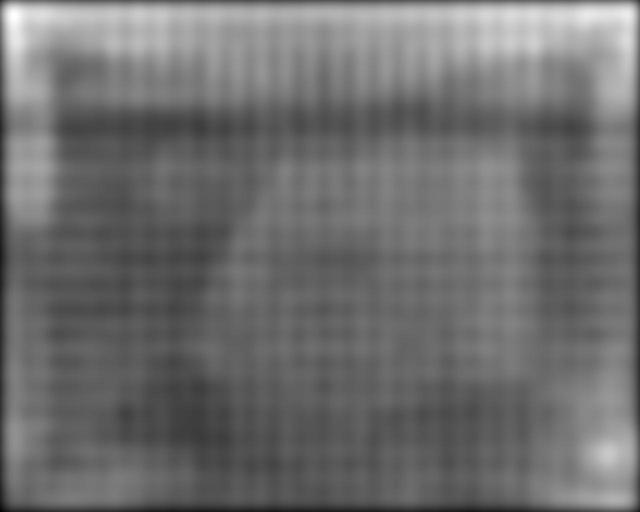}}
			\fbox{\includegraphics[width=.12\textwidth]{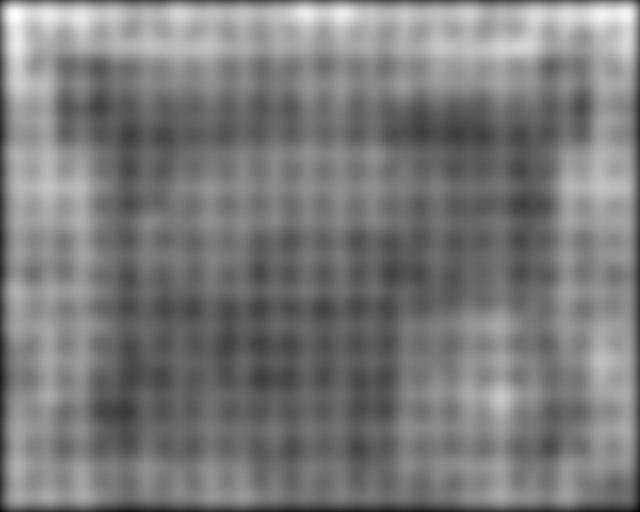}}
			\fbox{\includegraphics[width=.12\textwidth]{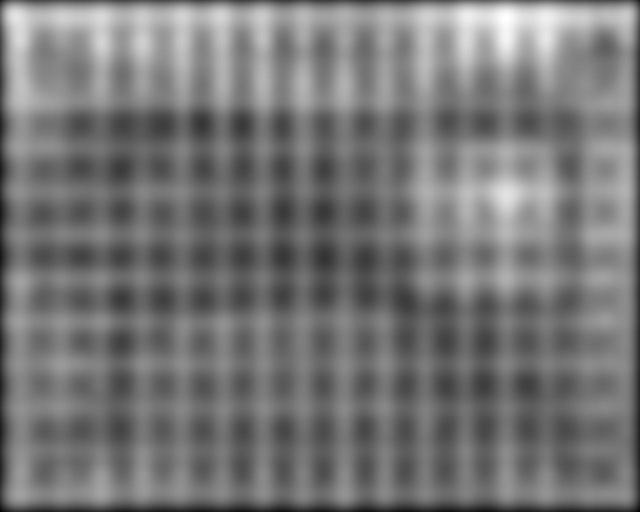}}
			\fbox{\includegraphics[width=.12\textwidth]{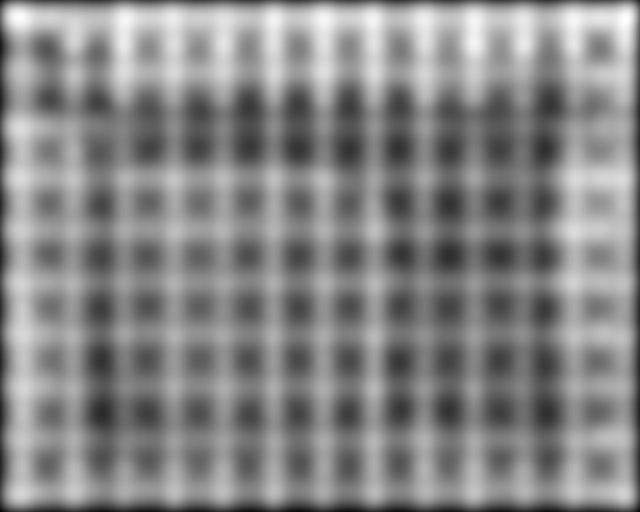}}
			\fbox{\includegraphics[width=.12\textwidth]{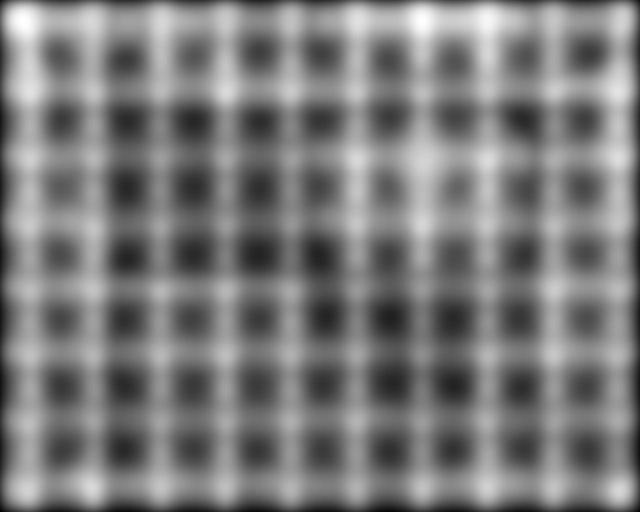}}
			\fbox{\includegraphics[width=.12\textwidth]{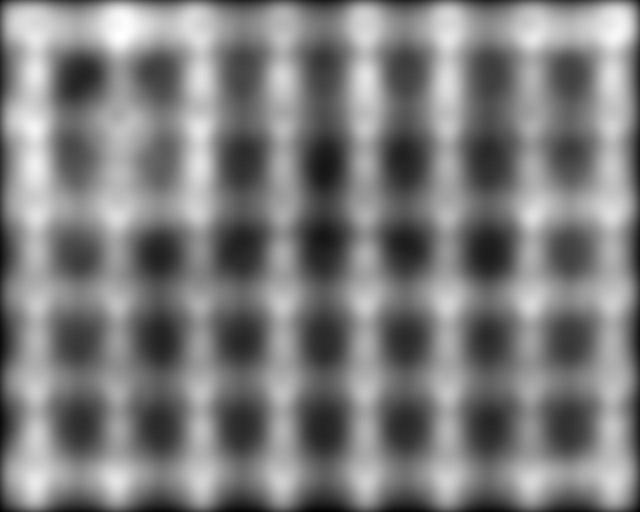}}
			\fbox{\includegraphics[width=.12\textwidth]{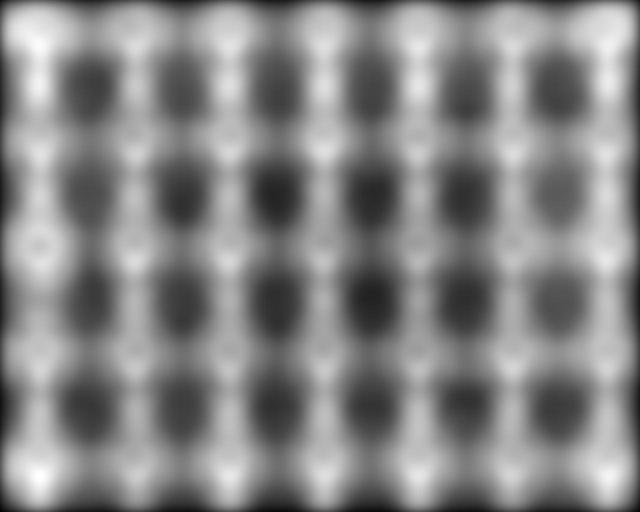}}
			\\
			\scriptsize{\hspace{.02\textwidth}$20\times 26$\hspace{.04\textwidth}$15\times 20$\hspace{.04\textwidth}$12\times 16$\hspace{.04\textwidth}$10\times 13$\hspace{.04\textwidth}$8\times 10$\hspace{.05\textwidth}$6\times 8$\hspace{.06\textwidth}$5\times 7$}
		\end{subfigure}
		\caption{Stimuli with distinct set sizes corresponding to search asymmetries present on a \textbf{(A)} salient circle crossed by a vertical bar among other circles and a \textbf{(B)} salient circle among other circles crossed by a vertical bar. Rows below \textbf{A,B} correspond to NSWAM's predicted saliency maps.}
		\label{fig:vs2}
	\end{figure}

\begin{figure}[h!]
		\centering
		\begin{subfigure}{.9\linewidth}
			\includegraphics[width=\textwidth]{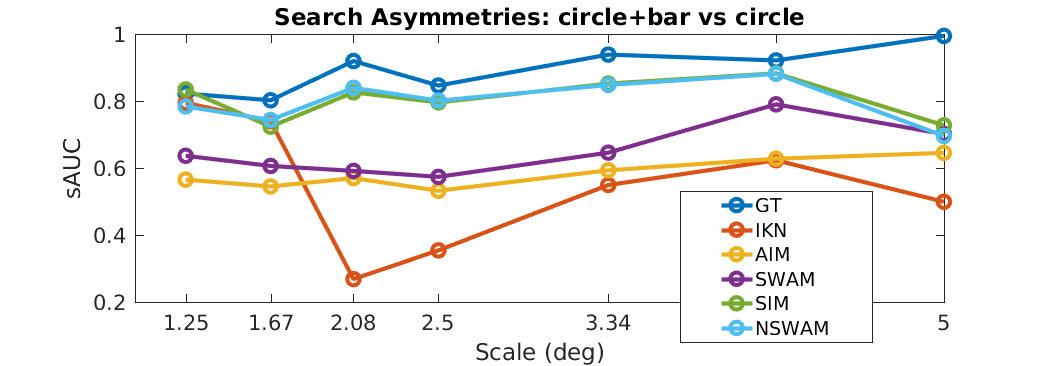}
			\caption*{\centering \textbf{A}}
		\end{subfigure}
		\begin{subfigure}{.9\linewidth}
			\includegraphics[width=\textwidth]{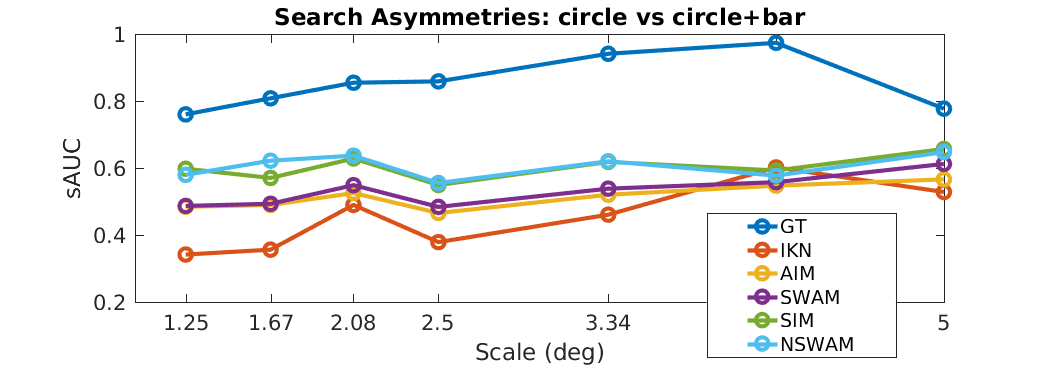}
			\caption*{\centering \textbf{B}}
		\end{subfigure}
		\caption{Results of sAUC upon varying scale and set size of \textbf{(A)} an array of circles and a salient one crossed by a vertical bar and \textbf{(B)} an array of circles crossed by a bar and a salient circle.}
		\label{fig:res_b7}
	\end{figure}

	Search asymmetries appear when searching target of type ``a" is found efficiently among distractors of type ``b", but not in the opposite case (i.e. searching for "b" among distractors of type ``a") \cite{Treisman1985}\cite{Wolfe2001}. Previous studies pointed out this concept when searching a circle crossed by a vertical bar among plain circles and searching a plain circle among circles crossed by a vertical bar. Using these two configurations, we filled a grid of distractors according to specific scales (\hyperref[fig:vs2]{Fig. \ref*{fig:vs2}}). Scale values ($s=[1.25, 1.67, 2.08, 2.5, 3.33, 4.17, 5] \deg$) change the amount of items, with arrays of $5\times 7$, $6\times 8$, $8\times 10$, $10\times 13$, $15\times 20$ and $20\times 26$ objects. In \hyperref[fig:res_b7]{Fig. \ref*{fig:res_b7}} our model is not only more efficient than other biologically-inspired models upon dissimilar sized objects but also on detecting conspicuous objects at distinct scales, accounting for lower or larger amount of distractors. sAUC for NSWAM showed to correlate for a conspicuous circle crossed by a vertical bar among circles ($\rho=.83$, $p=2.1\times 10 ^{-2}$) but not for a conspicuous circle among circles crossed by a vertical bar ($\rho=.15$, $p=.75$).
	
	\subsection{Ablation Study of Feature Integration}
	
	In this section, we include some brief results testing how distinct fusion methods can efficiently integrate information to the unique saliency map (mainly what we consider as the SC function). As mentioned in the first step of \hyperref[sec:feat3]{Section \ref*{sec:feat3}}, our original model (SWAM and NSWAM) uses the default inverse equation (\hyperref[eq:wavelets3]{Eq. \ref*{eq:wavelets3}}) which can be used for obtaining the original image (if we do not sum the conspicuity maps and normalize as the other steps).
		For this, we tested distinct mechanisms of Feature Integration, alternative to \hyperref[eq:sc1]{Eq. \ref*{eq:sc1}}:

	\begin{align}
		\begin{split}
		\hat{S}_{io}(max)=\max_{s,\theta}(\hat{S}_{iso\theta})+c_n.
		\label{eq:sc2}
		\end{split}\\
		\begin{split}
		\hat{S}_{io}(argmax)= arg\max_{s,\theta}(\hat{S}_{iso\theta})+c_n
		\end{split}
		\label{eq:sc3}
		\end{align}
	
\begin{figure}[h!]
		\centering
		\setlength{\fboxsep}{0pt}%
		\setlength{\fboxrule}{0.5pt}%
		\begin{subfigure}{\linewidth} 
			\makebox[4em]{"Image"} \includegraphics[width=0.25\textwidth]{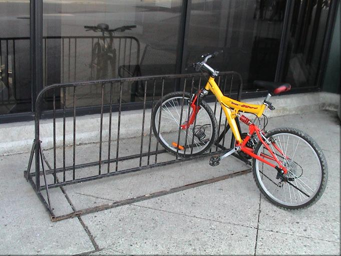}
			\includegraphics[width=0.25\textwidth]{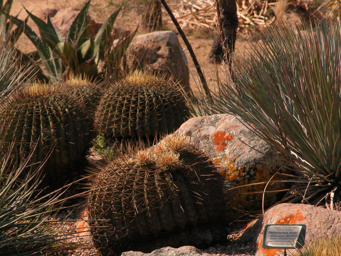}
			\includegraphics[width=0.25\textwidth]{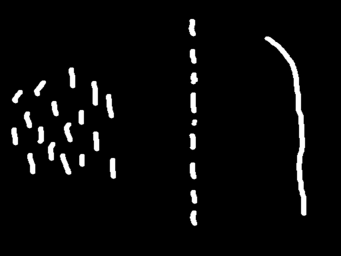}\\
			\makebox[4em]{"GT"} \includegraphics[width=0.25\textwidth]{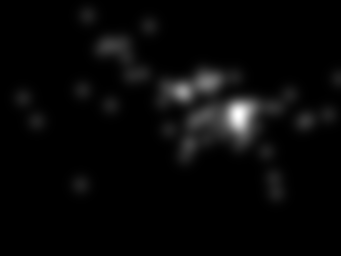}
			\includegraphics[width=0.25\textwidth]{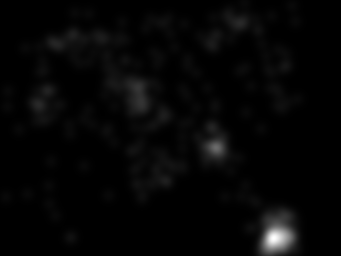}
			\includegraphics[width=0.25\textwidth]{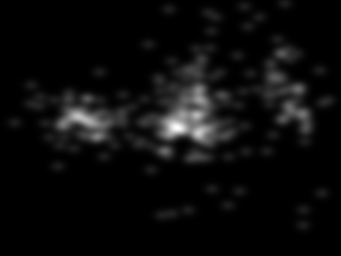}\\
			\makebox[4em]{"inverse"} \includegraphics[width=0.25\textwidth]{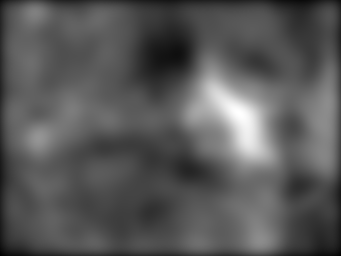}
			\includegraphics[width=0.25\textwidth]{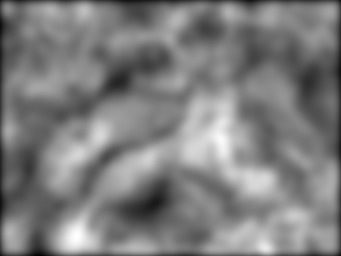}
			\includegraphics[width=0.25\textwidth]{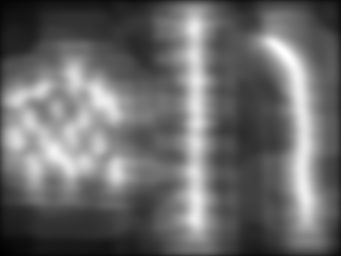}\\
			\makebox[4em]{"max"} \includegraphics[width=0.25\textwidth]{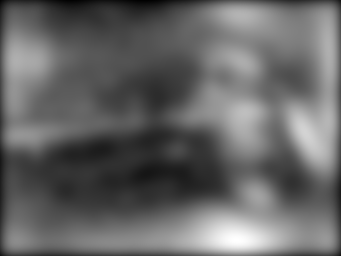}
			\includegraphics[width=0.25\textwidth]{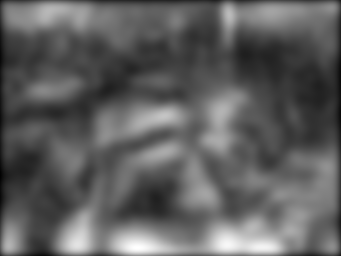}
			\includegraphics[width=0.25\textwidth]{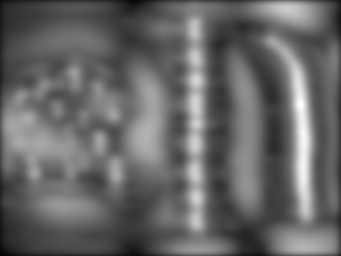}\\
			\makebox[4em]{"argmax"} \includegraphics[width=0.25\textwidth]{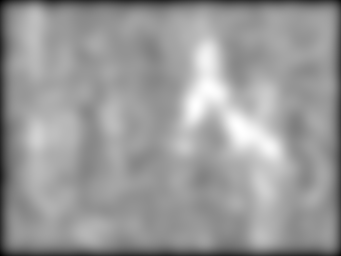}
			\includegraphics[width=0.25\textwidth]{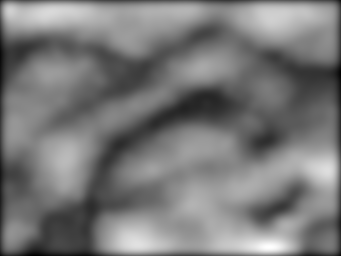}
			\includegraphics[width=0.25\textwidth]{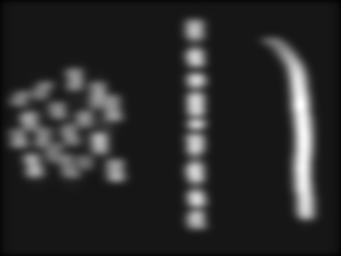}
		\end{subfigure}
		\caption{Qualitative examples for distinct Feature Integration techniques (rows 3-5), with real, nature and synthetic images (columns 1-3).}
		\label{fig:sc_ablation_qualitative}
	\end{figure}

	Where $\hat{S}_{io}(max)$ calculates the pointwise maximum of the retinotopic positions "i" in each scale "s" and orientation "$\theta$", separately for each channel "o" ($rg$, $by$, $L$). The case of $\hat{S}_{io}(argmax)$ considers the "winner" as the whole channel map $\hat{S}_{io}$ that contains the neuron with highest activity for all multiscale dimensions ($s,\theta$). We have computed the eye fixation prediction results for the all datasets in \hyperref[tab:sc_ablation]{Figure \ref*{tab:sc_ablation}}, and results show that performing the inverse transform (sum) of all maps we get best scores. In addition, we added qualitative results for \hyperref[fig:sc_ablation_qualitative]{Fig. \ref*{fig:sc_ablation_qualitative}}, with 3 examples of real, nature and synthetic images (being the inverse more similar overall to the GT).

	\begin{table} [h]
		\centering
		\includegraphics[clip,trim=1.9cm 19.5cm 2cm 2.6cm, width=\linewidth,height=5cm,keepaspectratio]{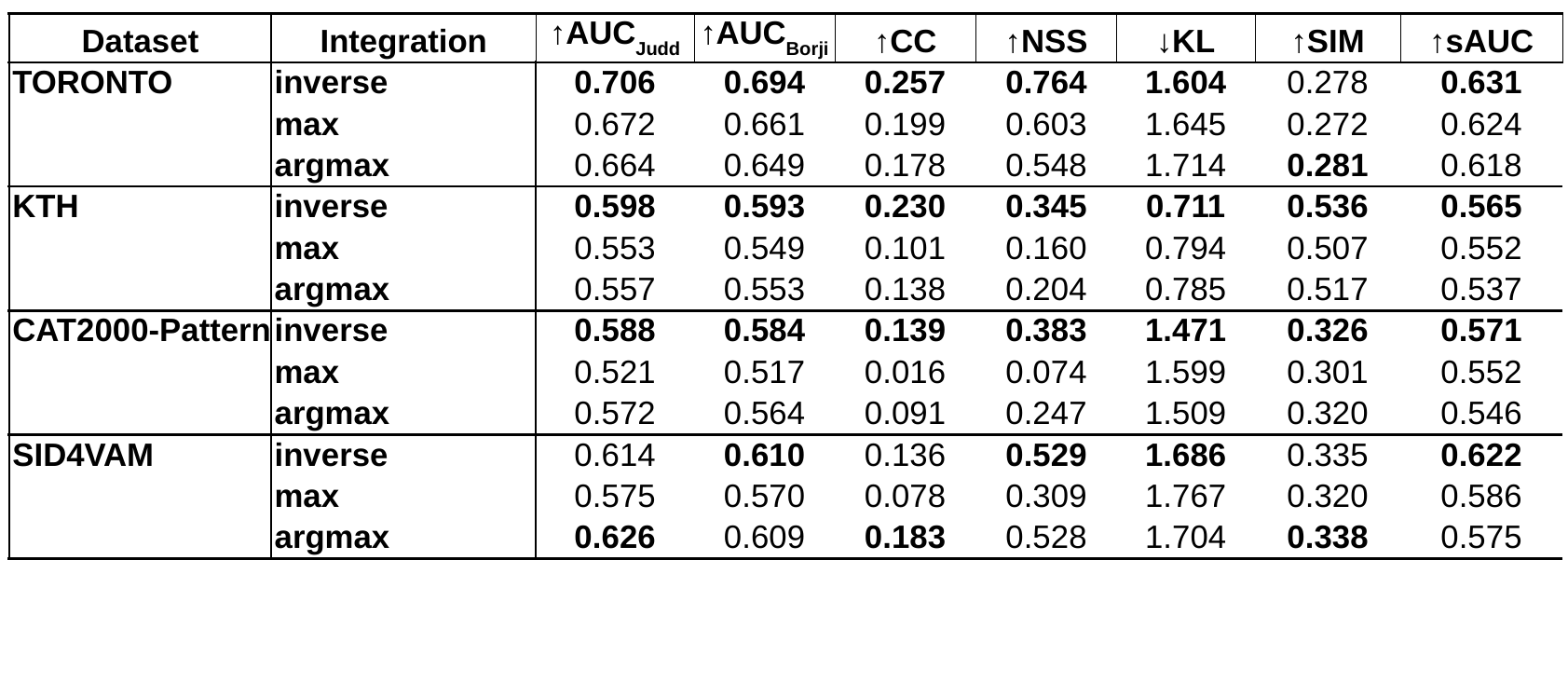}
		\caption{Results for prediction metrics for testing distinct baselines (inverse/sum, max and argmax), corresponding to mechanisms explained in Eqs. \hyperref[eq:sc1]{\ref*{eq:sc1}}, \hyperref[eq:sc2]{\ref*{eq:sc2}} and \hyperref[eq:sc3]{\ref*{eq:sc3}}.}
		\label{tab:sc_ablation}
	\end{table}

	\section{Conclusions}
	
	In this work, we hypothesize that low-level saliency is likely to be associated by the computations of V1. Concretely, we hypothesized that a neurodynamic model of V1's lateral interactions, processing each channel separately and acquiring firing rate dynamics from real image simulations, is able to simultaneously reproduce several visual processes, including low-level visual saliency. Here we have to pinpoint three statements in agreement with our findings: 
	
	\begin{itemize}
		\item First, our model of the lateral interactions in V1 show a performance similar to other state-of-the-art models on human eye fixations. In that sense, our model acquires similar results in comparison with saliency prediction baselines, specifically in metrics that penalize for center biases (sAUC and InfoGain). Additionally, our model outperforms other biologically-inspired saliency models in natural and synthetic images.
		\item Second, our model is consistent with human psychophysical measurements (tested for Visual Asymmetries, Brightness, Color, Size and Orientation contrast). Adding up to the stated hypothesis, our model presents highest performance at highest contrast from feature singleton stimuli (where salient objects pop-out easily). 
		\item Three, we remark the model plausibility by mimicking HVS physiology on its processing steps and being able to reproduce other effects such as Brightness Induction \cite{Penacchio2013}, Color Induction \cite{Cerda2016} and Visual Discomfort \cite{Penacchio2016}, efficiently working without applying any type of training or optimization and keeping the same parametrization.
	\end{itemize} 
	
	Other biologically plausible alternatives that predict attention using neurodynamic modeling \cite{Li1998}\cite{Deco2004}\cite{Chevallier2010} do not provide a unified model of the visual cortex able to reproduce these distinct tasks simultaneously, and specifically, using real static or dynamic images as input. We suggest that V1 computations work as a common substrate for several tasks, simultaneously. 
	
	Future work of interest would consist on predicting scan-paths for real scenes in order to provide gaze-wise temporal detail for saliency prediction and saccade programming. To do so, a foveation mechanism (such as a retinal \cite{Watson2014} or a cortical magnification transformation towards V1 retinotopy \cite{Schwartz1980}) would be needed in order to process each view of the scene distinctively. Other applications of the same model would be to generate saliency maps with dynamic scenes or videos (mainly used for visual tracking and salient object detection in several real world applications), integrating other features such as flicker or motion. In order to provide top-down computations for representing feature relevance apart from saliency, we could feed our model with a selective mechanism \cite{Tsotsos1995}\cite{Huang2007} for specific low-level feature maps, enabling the possibility to perform visual search tasks. As shown in Section \hyperref[sec:contrast_orientation]{\ref*{sec:contrast_orientation}}, saliency computations could be more accurately represented with a higher number of 2D Gabor/Log-Gabor filters \cite{TaiSingLee1996}\cite{Fischer2007}\cite{GarciaDiaz2012}. Considering the dependence of saliency to stimulus contrast, the usage of contrast-adaptive mechanisms \cite{Poirier2006} in the Feature Integration step (\hyperref[sec:feat3]{\ref*{sec:feat3}}) could dramatically improve results, specially for psychophysical pattern images. 
	Further modeling would include intra and inter-cortical interactions between simple and complex cells in a multilayer implementation of V1. Such implementation could adequate more detailed and efficient computations of V1, projecting the excitatory recurrent dynamics from V1 (specifically from Layer 5 complex cells, also named ``Meynert" cells) to SC \cite{Lund1975}\cite{Nhan_Callaway_2011}. Although latest hypotheses about the SC have suggested that saliency is processed in the SC and not by the visual cortex, corresponding to a distinct, feature-agnostic saliency map \cite{Veale2017}\cite{White2017b}, we claim the importance of the mechanisms of V1 to be responsible for computing distinctiveness between the stated low-level features, which might conjunctively contribute to the generation of saliency \cite{Li1999}\cite{Li2002}\cite{Yan2018}. However, modeling the computations of the pathways from the RGC to the SC would be of interest for a more integrated and complete model of eye-movement prediction, seeing the roles of the distinct projections to the SC and their computations, alternatively involved in the control of eye movements.

	\section{Compliance with Ethical Standards} 
	
	\paragraph*{\textbf{Funding: } } This work was funded by the Spanish Ministry of Economy and Competitivity
	(DPI2017-89867-C2-1-R), Agencia de Gesti\'o d'Ajuts Universitaris i de Recerca (AGAUR) (2017-SGR-649), and CERCA Programme / Generalitat de Catalunya.
	
	\paragraph*{\textbf{Conflict of Interests:  }} The authors declare that they have no conflict of interest. 
	
	\paragraph*{\textbf{Informed Consent: }} Informed consent was not required as no human or animals were involved. 
	\paragraph*{\textbf{Human and Animal Rights: }} This article does not contain any studies with human or animal subjects performed by any of the authors.
	
	
	\bibliography{main}
	
	\bibliographystyle{abbrv}

\end{document}